\documentclass[
11pt, %
english, %
singlespacing, %
liststotoc, %
headsepline, %
]{MastersDoctoralThesis} %

\thesistitle{Latent Communication \\in Artificial Neural Networks} 
\supervisor{Prof. Emanuele \textsc{Rodolà}} 
\examiner{} 
\degree{Doctor of Philosophy} 
\author{Luca \textsc{Moschella}} 
\addresses{} 

\subject{Computer Science} 
\university{\href{https://www.uniroma1.it/}{Sapienza University of Rome}} 
\department{\href{https://phd.uniroma1.it/web/COMPUTER-SCIENCE_nD3507_EN.aspx}{Department of Computer Science}} 
\group{\href{https://gladia.di.uniroma1.it/}{GLADIA research lab}} 
\faculty{\href{https://web.uniroma1.it/i3s/en}{Faculty of Information Engineering, Informatics, and Statistics}} 

\AtBeginDocument{
    \hypersetup{pdftitle=\ttitle} 
    \hypersetup{pdfauthor=\authorname} 
    \hypersetup{pdfkeywords=\keywordnames} 
}

\usepackage{amsmath,amsfonts,bm}

\def\eqref#1{equation~\ref{#1}}

\def\1{\bm{1}}

\def\vu{{\bm{u}}}
\def\vv{{\bm{v}}}

\DeclareMathAlphabet{\mathsfit}{\encodingdefault}{\sfdefault}{m}{sl}
\SetMathAlphabet{\mathsfit}{bold}{\encodingdefault}{\sfdefault}{bx}{n}

\def\sX{{\mathbb{X}}}

\DeclareMathOperator*{\argmin}{arg\,min}

\newcommand{\STAB}[1]{\begin{tabular}{@{}c@{}}#1\end{tabular}}

\algrenewcommand{\Return}{\State\algorithmicreturn~}

\newtheorem*{definition}{Definition}

\begin{document}

\frontmatter %

\pagestyle{plain} %

\begin{titlepage}
	\begin{center}

		\vspace*{.06\textheight}
		{\scshape\LARGE \univname\par}\vspace{1.5cm} %
		\textsc{\Large Doctoral Thesis}\\[0.5cm] %

		\HRule \\[0.4cm] %
		{\huge \bfseries \ttitle\par}\vspace{0.4cm} %
		\HRule \\[1.5cm] %

		\begin{minipage}[t]{0.4\textwidth}
			\begin{flushleft} \large
				\emph{Author:}\\
				\href{https://luca.moschella.dev/}{\authorname} %
			\end{flushleft}
		\end{minipage}
		\begin{minipage}[t]{0.4\textwidth}
			\begin{flushright} \large
				\emph{Supervisor:} \\
				\href{https://gladia.di.uniroma1.it/authors/rodola/}{\supname} \\
				\href{https://www.francescolocatello.com/}{Prof. Francesco \textsc{Locatello}} \\
			\end{flushright}
		\end{minipage}\\[3cm]

		\vfill

		\large \textit{A thesis submitted in fulfillment of the requirements\\ for the degree of \degreename}\\[0.3cm] %
		\textit{in the}\\[0.4cm]
		\groupname\\\deptname\\\facname\\[2cm] %

		\vfill

		{\large \today}\\[4cm] %

		\vfill
	\end{center}
\end{titlepage}

\vspace*{0.2\textheight}

\noindent\enquote{\itshape Be less curious about people and more curious about ideas. }\bigbreak

\hfill Marie Curie

\begin{abstract}
	\addchaptertocentry{\abstractname} %
	\looseness=-1
	As \glspl{nn} permeate various scientific and industrial domains, understanding the universality and reusability of their representations becomes crucial. At their core, these networks create intermediate neural representations, indicated as latent spaces, of the input data and subsequently leverage them to perform specific downstream tasks.
	This dissertation focuses on the \emph{universality and reusability of neural representations}.
	Do the latent representations crafted by a \gls{nn} remain exclusive to a particular trained instance, or can they generalize across models, adapting to factors such as randomness during training, model architecture, or even data domain? This adaptive quality introduces the notion of \textit{Latent Communication} -- a phenomenon that describes when representations can be unified or reused across neural spaces.

	\looseness=-1
	A salient observation from our research is the emergence of similarities in latent representations, even when these originate from distinct or seemingly unrelated \glspl{nn}. By exploiting a partial correspondence between the two data distributions that establishes a semantic link, we found that these representations can either be projected into a universal representation \citep{moschella2023}, coined as \glsxtrlong{rr}, or be directly translated from one space to another \citep{maiorca2023}.
	Intriguingly, this holds even when the transformation relating the spaces is unknown \citep{cannistraci2023} and when the semantic bridge between them is minimal \citep{cannistraci2023a}.
	Latent Communication allows for a bridge between independently trained \gls{nn}, irrespective of their training regimen, architecture, or the data modality they were trained on -- as long as the data semantic content stays the same (e.g., images and their captions).
	This holds true for both generation, classification and retrieval downstream tasks;  in supervised, weakly supervised, and unsupervised settings; and spans various data modalities including images, text, audio, and graphs -- showcasing the universality of the Latent Communication phenomenon.
	From a practical standpoint, our research offers the potential to repurpose and reuse models, circumventing the need for resource-intensive retraining; enables the transfer of knowledge across them; and allows for downstream performance evaluation directly in the latent space.

	\looseness=-1
	Indeed, several works leveraged the insights from our Latent Communication research \citep{wu2023,norelli2022,wang2023a,kiefer2023a,jian2023}. For example, relative representations have been instrumental in attaining state-of-the-art results in \glsxtrlong{wvlp} \citep{chen2023}.
	Reflecting its significance, \citep{moschella2023} has been presented orally at ICLR 2023 and Latent Communication has been a central theme in the UniReps: Unifying Representations in Neural Models Workshop at NeurIPS 2023, co-organized by our team.
\end{abstract}

\begin{acknowledgements}
	\addchaptertocentry{\acknowledgementname}
	First and foremost, I would like to express my gratitude to all my colleagues who worked in this line of research: Cannistraci Irene, Crisostomi Donato, Fumero Marco, Maiorca Valentino, Norelli Antonio, and Ricciardi Antonio.
	In particular, I am deeply thankful to Irene Cannistraci for her unwavering support, insightful discussions, joint work, and meticulous proofreading of this manuscript. Special thanks also go to Marco Fumero and Donato Crisostomi for their immense effort in organizing the first edition of the UniReps Workshop at NeurIPS 2023, which was a resounding success due to their dedication. A special mention goes to Valentino Maiorca, with whom I embarked on this journey. Working alongside Valentino has been an exceptional experience, and this dissertation owes much to our collaborative efforts.

	I am immensely grateful to Prof. Emanuele Rodolà for fostering a free and creative environment. Initiating an entire research direction from playful observations at the laboratory has been truly inspiring. I am equally thankful to Prof. Francesco Locatello for his insightful discussions, supervision, and the opportunity to visit and collaborate during my fantastic period at ISTA. My heartfelt thanks also go to Dr. Maria Shugrina and Dr. Vojtěch Micka for the fantastic experiences and valuable time spent during my internships at NVIDIA and NNAISENSE. I extend my sincere gratitude to the external reviewers,  Prof. Nina Miolane and Prof. Marco Baroni, whose insightful remarks and suggestions have greatly improved this work.

	My sincere thanks to all the other people with whom I had the honor to collaborate. The work we have accomplished together and the moments we have shared have been incredibly rewarding:
	Andrea Santilli, Cosimo Fiorini, Emanuele Frascaroli, Filippo Maggioli, Giambattista Parascandolo, Giorgio Mariani, Giovanni Trappolini, Leonidas Guibas, Luca Cosmo, Maks Ovsjanikov, Marco Ciccone, Matteo Boschini, Michele Bevilaqua, Nishkrit Desai, Or Litany, Or Perel, Pietro Barbiero, Pietro Liò, Riccardo Benaglia, Riccardo Marin, Roberto Dessì, Simone Antonelli, Simone Calderara, Simone Melzi, and Steve Azzolin.
	Moreover, I am thankful for the wonderful time and the enriching experiences shared with all the people at Sapienza, ISTA, NNAISENSE, and NVIDIA.

	I am grateful to all the other colleagues from the GLADIA group at Sapienza and the Causal Learning and Artificial Intelligence group at ISTA for the stimulating discussions and the time spent together: Adrian R. Minut, Arianna Rampini, Berker Demirel, Daniele Baieri, Dingling Yao, Emilian Postolache, Irene Tallini, Lorenzo Basile, Marco Pegoraro, Michele Mancusi, Riccardo Cadei, and Silvio Severino.

	My appreciation also extends to all the other people with whom I had stimulating discussions that have enriched and helped me along my journey: Alessandro Raganato, Alex Bronstein, Andrea Dittadi, Ari Morcos, Bogdan Gaza, Christos Tsirigotis, Clémentine Dominé, Emanuele Marconato, Emanuele Rossi, Even Oldridge, Fabrizio Frasca, Federico Scozzafava, Filip Szatkowski, Francesco Visin, Giovanni Zappella, Jonathan Masci, Luigi Gresele, Mateusz Pyla, Matthew Leavitt, Michael Bronstein, Patrik Reizinger, Pau Rodríguez López, Simone Azeglio, Simone Scardapane, Stefan Bejgu, Valentina Zantedeschi, Xavier Suau, and Zorah Lähner.

	Finally, I want to thank everyone who shared their insights and encouragement with me, whether we worked together directly or simply had inspiring conversations. Your contributions have been truly valuable.

\end{acknowledgements}

\tableofcontents %

\listoffigures %

\listoftables %

\let\oldtexttt\texttt %
\renewcommand{\texttt}[1]{#1} %

\printglossary[type=\glsxtrabbrvtype,title={Glossary}]

\let\texttt\oldtexttt

\printglossary[type=symbols,title={List of Symbols}]

\dedicatory{Dedicated to my mother\ldots}

\mainmatter %

\pagestyle{thesis} %

\nocite{%
	cannistraci2023,%
	crisostomi2023from,%
	cannistraci2023a,%
	maiorca2023,%
	moschella2023,%
	norelli2022,%
	ricciardi2023,%
	DBLP:journals/corr/abs-2201-10222,%
	trappolini2021shape,%
	moschellaSpectralUnionsPartial2021,%
	DBLP:journals/corr/abs-2301-03345,%
	crisostomi2022metric,%
	srivastava2023beyond,%
}

{
	\begingroup
	\emergencystretch=1em

	\printbibheading[heading=bibintoc,title={Authored Publications}]

	\printbibliography[heading=subbibintoc,keyword=author,keyword=latent,title={Latent Communication}]
	\begin{NoHyper}
		\def\thefootnote{*}\footnotetext{Equal contribution.}\def\thefootnote{\arabic{footnote}}
	\end{NoHyper}

	\printbibliography[heading=subbibintoc,keyword=author,keyword=other,title={Other Research Directions}]

	\endgroup
}

\part{Introduction}

\chapter{Introduction to Latent Communication}
\label{chap:introduction}

The intrinsic meaning of data often resides in complex, lower-dimensional structures, we define as “abstract manifolds”.
These manifolds, grounded in the Manifold Hypothesis \citep{fefferman2016testing}, represent the compact, underlying essence of high-dimensional data. Indeed, both human and machine capabilities  are limited to observing representations of meaning that manifest within high-dimensional spaces. These representations serve as proxies for deeper, underlying conceptual entities, which are not directly observable. To illustrate this concept, consider the entity “cat”. It does not reside within our immediate perceptual field but within an abstract manifold of meaning. The “cat” discerned and interpreted is not the conceptual entity per se, but its representation within a higher-dimensional space, such as images or textual descriptions of cats. Furthermore, it is crucial to recognize that different spaces may exhibit varying degrees of expressive power, potentially leading to differences in the underlying abstract manifold. Nonetheless, abstract manifolds that denote the same conceptual entities bear semantic connections and are intrinsically similar.
When we examine these related manifolds embedded into high-dimensional spaces, we directly observe a correspondence between these representations -- for instance, between the textual descriptions of cats and their visual images.
This alignment, in essence, establishes a connection between the manifold meanings, bridging the abstract conception of “cat” as described in textual captions with its visual representation in images.

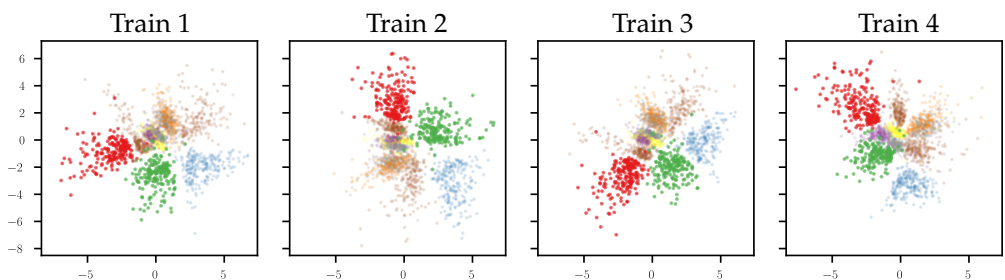
\begin{figure}[h]
    \centering
    \begin{overpic}[trim=0cm .3cm 0cm -.25cm,clip,width=.95\linewidth]{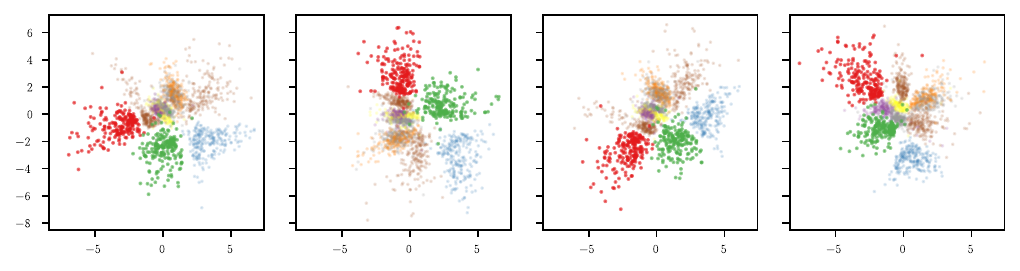}
        \put(11.5, 24.25){\small Train 1}
        \put(36.5, 24.25){\small Train 2}
        \put(60.5, 24.25){\small Train 3}
        \put(84.5, 24.25){\small Train 4}
    \end{overpic}
    \caption[Latent spaces learned by distinct trainings of the same AE]{Latent spaces learned by distinct trainings of the same AE on \gls{mnist}.
        The bottleneck has size $2$; thus, there is no dimensionality reduction in the latent space visualizations.
        The stochasticity in the training phase induces \emph{intrinsically similar representations}.
    }
    \label{introduction:fig:latent-rotation}
\end{figure}

\glsxtrshortpl{nn} play a central role in this context, learning to transform high-dimensional data (e.g., images) into meaningful representations that are helpful for solving downstream tasks. Typically, these representations are seen as elements of a vector space, denoted as latent space, which corresponds to the constrained output (explicitly or implicitly) of a key component of the \gls{nn},  e.g., the bottleneck in an \gls{ae}, or the word embedding space in \gls{nlp} tasks.
The foundational assumption is that learned latent spaces should encode the essential data characteristics and its abstract meaning required for task resolution.
Thus, they should be an optimal encoding given the data distribution, the downstream task, and the network constraints.

In practice, however, learned latent spaces may vary, even when the initial assumptions are held constant.
This phenomenon is illustrated in \Cref{introduction:fig:latent-rotation}, demonstrating the variation in latent spaces generated by training multiple times, from scratch, an \gls{ae} with a two-dimensional bottleneck on the \gls{mnist} dataset.
Perhaps unsurprisingly, these spaces differ from one another, breaking the fundamental assumptions made above. Indeed, the distribution of the latent embeddings is affected by various factors, such as the random initialization of the network weights, the data shuffling, hyperparameters, and other stochastic processes in the training phase. Although the resulting models may perform equally well on the task, this situation introduces several practical challenges. For example, it is notoriously challenging to compare latent spaces across different trainings or across different \glsxtrshortpl{nn}; perhaps more importantly, re-using neural components trained on different embeddings of the same data becomes impossible, since they are incompatible.

Interestingly, although different, the learned representations in \Cref{introduction:fig:latent-rotation} exhibit \emph{intrinsic  similarities}: the distances between the embedded representations are approximately the same across all spaces, even if their absolute coordinates differ. Indeed, in this case, the learned latent spaces are the same up to a nearly isometric transformation.\footnote{To the best of our knowledge, the first to acknowledge this behavior was \citep{Colah2015} in a blog post.}
This symmetry arises from two main causes: {(i)} the existence of a semantic correspondence between their associated abstract manifolds, which are exactly the same in this instance since the same data and task is being considered; and {(ii)} the implicit biases underlying the optimization process, as noted by \citep{Soudry2017Oct}, which compel the model to generalize. Consequently, this generalization ensures that similar samples -- with respect to the task -- are represented similarly.
Discovering such symmetries and conserved quantities is a core step for extracting meaningful representations from raw data in biological and artificial systems \citep{higgins2022symmetrybased,benton2020learning,lyle2020benefits}. Moreover, note that these emerging transformations, which relate intrinsically similar spaces, are limited to the embedded manifolds; thus, we cannot expect the same relationship between the entire ambient spaces, e.g., when considering out of distribution samples. This is a crucial point, as it implies that only a subset of the latent spaces is similar and \emph{easily alignable}.

Leveraging this key observation, this dissertation investigates the \textit{universality and reusability of these representations}. Do the latent representations crafted by a \gls{nn} remain exclusive to a particular trained instance, or can they generalize across models, adapting to factors such as randomness during training, model architecture, or even data domain?
This adaptive quality forms the basis of our exploration into \emph{Latent Communication} -- a novel paradigm that enables the unification or reuse of representations across disparate neural spaces, formally defined in \Cref{chap:problemformalization}.
We investigate two principal strategies to harness this phenomenon, exploiting the partial semantic correspondence between the two spaces:
\begin{description}
    \item[Universal Representation] projects the latent spaces into a universal representation where the embedded manifolds are extrinsically equal. In practice, this universal space must be independent of the specific training regimen, architecture, data modality, and other stochastic factors; encoding only the intrinsic information underlying the data. We show an example of universal representation in \Cref{chap:relative}, denoted as \emph{\gls{rr}} adopting a local coordinate system defined by the data itself.
    \item[Direct Translation] translates directly between two specific latent spaces, explicitly approximating an ambient space transformation that induces an alignment between the manifolds embedded within them. We show an example of direct translation in \Cref{chap:translation}, assuming the transformation relating the spaces is at most affine.
\end{description}
Through these strategies, we facilitate effective communication between latent spaces of different \glsxtrshortpl{nn}, bridging the divide between various domains, models, architectures, and modalities.

From a practical standpoint, our research paves the way for model repurposing and \emph{reuse}, eliminating the need for resource-intensive retraining and fostering a more sustainable AI development cycle; enables the \emph{transfer of knowledge} across them, and allows for performance \emph{evaluation directly in the latent space}. This holds true for generation, classification, and retrieval downstream tasks;  in supervised, weakly supervised, and unsupervised settings; and spans various data modalities including images, text, audio, and graphs -- showcasing the universality of the Latent Communication phenomenon. For example, it becomes feasible to classify images with a text classifier, or vice versa (\Cref{translation:sec:cross-modality}).

This research makes multiple contributions to the field, summarized as follows:
\begin{itemize}
    \item We empirically demonstrate that while representations learned by \glsxtrshortpl{nn} can change due to various influencing factors, often the transformation that relates them is simple (e.g., the angles between latent embeddings often remain consistent).
    \item We introduce the novel concept of \emph{Latent Communication} and formalize the \glsfirst{lcp} (\Cref{chap:problemformalization}), providing a paradigm for understanding and leveraging the inherent connections between independently trained \glsxtrshortpl{nn}; irrespective of their training regimen, architecture, or the data modality they were trained on -- as long as the data semantic content stays the same (e.g., images and their captions).
    \item For the first time, we successfully showcase \emph{Zero-Shot Stitching} (\Cref{sec:stitchingdefinition}) of neural components produced by distinct training regimens, e.g., due to different seeds, neural architectures or data domains;
    \item Provide a \textit{quantitative} latent measure of performance while training neural models, which is differentiable, does not need any labeled data, and is correlated with standard downstream performance measures such as accuracy.
    \item With an extensive set of experiments, we validate the performance of the proposed methods in multiple settings, tasks (classification, generation, retrieval), architectures (e.g., Transformers, \glspl{cnn}), and modalities (e.g., images, text, graphs, audio); showing that is possible to achieve Latent Communication across different architectural and modality changes.
\end{itemize}

Indeed, several works leveraged the insights from our Latent Communication research, and in particular the concept of \glsxtrlongpl{rr}. For example, they have proven fundamental in enabling continuous prompt transfers in \glspl{llm} \citep{wu2023}; zero-shot image captioning without requiring any multimodal model training \citep{norelli2022}; understanding shared speech-text representations \citep{wang2023a};  analyzing cognitive graphs \citep{kiefer2023a} and in the stitching of reinforcement learning agents in novel environments \citep{jian2023,ricciardi2023}; thus achieving zero-shot policy reuse. Our approach has also been instrumental in attaining state-of-the-art results in \glsxtrlong{wvlp} \citep{chen2023}. Reflecting its significance, \citep{moschella2023} has been presented orally at ICLR 2023 and Latent Communication has been a central theme in the UniReps: Unifying Representations in Neural Models Workshop at NeurIPS 2023, co-organized by our team.

\section{Structure of the Thesis}
In this dissertation, we present a novel unified perspective on the \glsfirst{lcp} research, reinterpreting several of our recent works  \citep{moschella2023, maiorca2023, cannistraci2023, cannistraci2023a, norelli2022, crisostomi2023from, ricciardi2023}. For the first time, we provide a formalization of the \gls{lcp} in \Cref{chap:problemformalization}.

Subsequently, in \Cref{chap:relative,chap:translation}, we present two distinct methodologies to solve the \gls{lcp}.
The first method, discussed in \Cref{chap:relative}, revolves around the concept of \gls{rr}, which seeks to unify latent spaces into a universal space. The second approach, outlined in \Cref{chap:translation}, focuses on establishing direct mappings between source and target spaces.

In \Cref{chap:limitations} we discuss the limitations of the current approaches to solve the \gls{lcp}, and in \Cref{chap:bridge,chap:bootstrapping} we present two methods to overcome these limitations. In \Cref{chap:bridge}, we introduce a novel approach to tackle the \gls{lcp} without any specific assumption on the transformation class relating the latent spaces. Meanwhile, in \Cref{chap:bootstrapping} we delineate a methodology to expand a small semantic correspondence between two latent spaces into a larger one, enabling the solution of the \gls{lcp} even when it was not possible before.

To further illustrate the practical implications of \gls{lcp}, we present three case studies in \Cref{chap:asif,chap:chart,chap:rl}. In \Cref{chap:asif} we show how to perform zero-shot captioning employing only unimodal models; in \Cref{chap:chart} we show how to merge distinct latent spaces; and in \Cref{chap:rl} we show how solving \gls{lcp} enables zero-shot policy reuse in reinforcement learning.

The dissertation concludes in \Cref{chap:conclusions} with a summary of the key findings. In \Cref{chap:contributions}, we outline our main contributions to the field and discuss how other researchers have utilized our work. Finally, \Cref{chap:future} presents potential avenues for future research.

\chapter{Related Work}

\section{Representation similarity}
\label{sec:representation-similarity}
Recent years have witnessed a growing consensus among researchers in the deep learning community that ``good'' \glsxtrshortpl{nn} tend to learn similar representations for semantically similar data, regardless of the architecture, training procedure, or domain in which they are applied.

This idea is supported by a plethora of empirical studies.
For example,
\cite{Morcos2018-ra} demonstrates that networks that generalize converge to more similar representations than networks that memorize;
\cite{Li2015-jo} shows that some features are learned reliably in multiple networks;
\cite{Kornblith2019-sz} verifies that wider networks learn more similar representations;
\cite{Bonheme2022-tk} shows that the \gls{vae} encoders representations in all but the mean and variance layers are similar across hyperparameters and learning objectives;
\cite{Tsitsulin2019-gg} develops an intrinsic method to characterize unaligned data manifolds of different dimensionality;
\cite{Lenc2014-gy} shows the shallow representations in the first layers of \glspl{cnn} are interchangeable across different networks;
\cite{Conneau2017-vv} shows it is possible to build a bilingual dictionary by aligning monolingual word embeddings spaces in an unsupervised way;
\cite{rakotonirina2023can} shows that automatically generated prompts can be learned on a language model and used to retrieve information from another;
and many others \citep{Barannikov2021-eb,DBLP:journals/corr/MikolovLS13,NEURIPS2021_46407417,vulic-etal-2020-good,bengio2014representation,Movshovitz-Attias2017-rn,Chang2022-ad}, recognizing that the phenomenon is particularly pronounced for large and wide models \citep{Somepalli2022-kw,Mehta2022-mc}.
Furthermore, similar observations have been made in the context of biological models \citep{Laakso2000ContentAC,Kriegeskorte:2008,raizada2012makes,acosta2023evaluation} and between artificial and biological representations \citep{sucholutsky2023a,sucholutsky2023b}, suggesting foundational principles on information representation.

Although this is still not unanimously recognized \citep{Wang2018-ho} and missing strong theoretical justifications (\Cref{related:theoretical}),
our framework is supported by the empirical evidence widely reported in these works. The \gls{lcp} assumes that well-performing \glsxtrshortpl{nn} trained on similar tasks and data produce intrinsically similar latent spaces, as formalized in \Cref{chap:problemformalization}, which allows us to unify them.

\section{Representation similarity measures}
Several metrics have been proposed to compare latent spaces generated by independent \glsxtrshortpl{nn} \citep{klabunde2023similarity}, capturing their inherent similarity up to transformations that correlate the spaces.
A classical statistical method is \gls{cca} \citep{hotelling1992relations}, which is invariant to linear transformations. While variations of \gls{cca} seek to improve robustness through techniques like \gls{svd} and \gls{svcca} \citep{raghu2017svcca} or to reduce sensitivity to perturbations using methods such as \gls{pwcca} \citep{Morcos2018-ra}.
Closely related to these metrics, the \gls{cka} metric \citep{Kornblith2019-sz} measures the similarity between latent spaces while disregarding orthogonal transformations. However, recent research \citep{davari2022reliability} demonstrates its sensitivity to transformations that shift a subset of data points in the representation space.
Furthermore, highly relevant in the biological domain, \gls{rsa} is a method \citep{Kriegeskorte2008RepresentationalSA,Nili2014ATF} used to compare and analyze the similarity of neural representations across different conditions, stimuli, or brain regions by correlating their respective similarity matrices.

\section{Manifold alignment}
\label{related:align}
Procrustes analysis has been instrumental in the alignment of latent spaces in deep \glsxtrshortpl{nn} \citep{wang2008, wang2009manifold}, particularly in \gls{nlp}, where it is well-known that latent spaces of different languages are isomorphic \citep{vulic-etal-2020-good} and can be effectively aligned
\citep{DBLP:journals/corr/MikolovLS13, xing-etal-2015-normalized}.
Rooted in shape analysis, this method efficiently uncovers correspondences between latent spaces of different models through the estimation of an optimal orthogonal transformation \citep{gower1975a}.
Previous works largely exploit Procrustes analysis to align latent spaces produced by the same architecture in different contexts \citep{Csiszarik2021-yi}, such as multilingual \gls{fasttext} embeddings \citep{bojanowski2016enriching,smith2017}. Procrustes analysis is termed ``manifold alignment'' because it aligns sets of keypoints that represent samples from lower-dimensional manifolds in a higher-dimensional space, thus aligning the intrinsic geometric structures of these manifolds.

Our research shares the common objective of aligning latent spaces, leveraging this alignment to enable or improve performance on various downstream tasks. For example, in \Cref{chap:translation} we broaden the scope of Procrustes analysis, applying it to Zero-Shot Stitching across disparate latent space dimensionalities, architectures, and data modalities.

\section{Model stitching}
Building on the observation of emergent intrinsic similarities between latent spaces, model stitching -- combining different \glsxtrshortpl{nn} to create a new model -- has become an active research topic in the field of representation learning.
For example,
\cite{Lenc2014-gy} introduces \emph{trainable} stitching layers that allow swapping parts of different networks;
\cite{Csiszarik2021-yi} demonstrates that the inner representations emerging in deep convolutional \glsxtrshortpl{nn} with the same architecture, but different initializations can be matched with a surprisingly high degree of accuracy even with a single, affine \emph{trainable} stitching layer;
while \citep{Bansal2021-oj,Csiszarik2021-yi} employ stitching to quantitatively verify statements such as ``good networks learn similar representations'' and ``more data, width, or time is better''.
Other works, such as \citep{Gygli2021-qb,DBLP:journals/corr/abs-2111-07632,https://doi.org/10.48550/arxiv.2208.03861,DBLP:journals/corr/abs-2007-14906}, try to directly produce compatible and reusable network components without stitching layers.
In general, stitching has been mostly adopted in the literature to analyze \glsxtrshortpl{nn} and verify statements regarding latent space similarity.
An exception is \citep{lahner2023on} that, concurrently to the work presented in \Cref{chap:translation}, targets the direct alignment of representational spaces, focusing on the compatibility of models trained end-to-end.

\looseness=-1In our framework, we (i) sidestep the need for trainable stitching layers and propose for the first time \emph{Zero-Shot} Model Stitching (\Cref{sec:stitchingdefinition}); and (ii) propose to employ stitching to effectively reuse neural components, enabling many practical applications, some of which presented in \Cref{chap:relative,chap:translation,chap:bootstrapping,chap:bridge,chap:casestudies,chap:asif,chap:chart,chap:rl}.

\section{Relative information}
Recognizing the importance of the relationships between data points, several methods have been proposed to exploit the relative information in the data.
For example,
the attention mechanism \citep{Vaswani2017-qw} and its variants \citep{Kossen2021-mj} exploit the relationship between features to extract meaningful representations;
\citep{Snell2017-fe} learn a metric space where the classification can be performed by measuring the distances with respect to prototype representations;
\cite{You2019-xl} introduces Position-aware Graph Neural Networks (P-GNNs) to exploit position-aware node embeddings,
\cite{Shalam2022-hj} suggested the Self Optimal Transport feature transform to enrich the sample representations with higher order relations between the instance features, while \cite{Alvarez-Melis2018-kr} suggested a general formulation of the optimal transport that accounts for global invariances in the underlying feature spaces.

Mathematically, the method presented in \Cref{chap:relative} bears resemblance to a kernel method~\citep{hofmann2008kernel}; employing similarities of embedded features as a core ingredient. However, differently from kernel methods, we do not introduce learnable parameters and, crucially, we compute the representations explicitly without resorting to a kernel trick.

\section{Invariance and Equivariance in Representations}
Invariances in \gls{nn} models can be enforced through various techniques operating at different levels, including adjustments to model architecture, training constraints, or input manipulation \citep{lyle2020benefits}.
For example,
\citep{benton2020learning} proposes a method to learn invariances and equivariances introducing augmentations in the training process; \citep{immer2022invariance} introduces a gradient-based approach that effectively captures inherent invariances in the data. Meanwhile, \citep{van2022learning} enables training of \glsxtrshortpl{nn} with invariance to specific transformations by learning weight-space equivalents instead of modifying the input data.
Other works directly incorporate invariances into the model through specific constraints, e.g., \citep{rath2023deep} enforces a multi-stream architecture to exhibit invariance to various symmetry transformations without relying on data-driven learning; \citep{kandi2019incorporating} suggests an improved \gls{cnn} architecture for better rotation invariance; and \citep{gandikota2021training} introduces a method for designing network architectures that are invariant or equivariant to structured transformations.

In contrast, the methodology presented in \Cref{chap:relative} proposes an alternative representation of the latent space that guarantees invariance to angle preserving transformation \emph{of the latent space itself}, without requiring additional training but only a subset of the data. Building on this, \Cref{chap:bridge} presents a method that directly \emph{incorporates a set of invariances} into the learned latent space, creating a product space of invariant components which, combined, can capture complex transformations between the latent spaces.

\section{Theoretical Understanding}\label{related:theoretical}
While the empirical findings discussed throughout this manuscript provide substantial evidence for the similarity of representations in \glsxtrshortpl{nn}, an exhaustive theoretical foundation is essential for fully understanding and leveraging these phenomena. Recent theoretical advancements have begun to shed light on the mechanisms behind the emerging representation similarity, offering a more solid ground for the empirical observations and methodologies employed in representation learning.

One direction of theoretical progress is the study of harmonics in \glspl{nn} weights \citep{Marchetti2023harmonics}, providing a mathematical framework for understanding the universality of neural representations. Furthermore, the intrinsic similarity of latent spaces, a core assumption of our framework, finds theoretical support in the field of linear identifiability within deep neural models, particularly in the context of nonlinear \gls{ica} \citep{roeder2021,hyvarinen2019,khemakhem2020,hyvarinen2016} and \gls{ima} \citep{sliwa2022probing,gresele2021independent,ghosh2023independent}. This body of work suggests that, despite the complexity and non-linearity of deep learning models, their learned representations may converge towards similar structures when they capture the same underlying generative factors of data.

\part{Latent Communication}

\chapter{Problem Formalization}\label{chap:problemformalization}

\begin{figure}[t]
    \centering
    \begin{overpic}[trim=0cm 0cm 0cm 0cm, clip,width=1\linewidth]{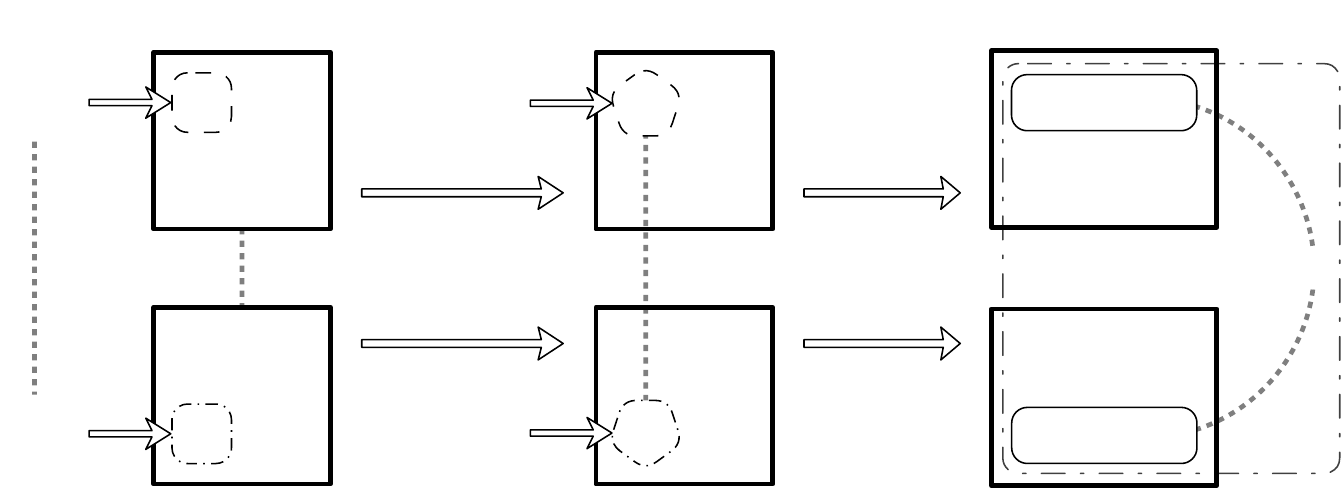}
        \put(11,35){\small Input Spaces}
        \put(43.5,35){\small Latent Spaces}
        \put(73.25,35){\small Universal Space}
        \put(16.5, 22){\large\gls{X}}
        \put(16.5, 9.5){\large\gls{Y}}
        \put(49.75, 22){\large\gls{ZX}}
        \put(49.75, 9.5){\large\gls{ZY}}
        \put(80.5, 22){\large\gls{UX}}
        \put(80.5, 9.5){\large\gls{UY}}
        \put(49, 15.5){\gls{transformation}}
        \put(1, 28.5){\small\gls{Mx}}
        \put(7.25, 30.25){\small\gls{phiX}}
        \put(1, 3.5){\small\gls{My}}
        \put(7.25, 5.75){\small\gls{phiY}}
        \put(4, 16){\small\gls{C}}
        \put(19, 16){\small\gls{pi}}
        \put(34, 28.5){\small\gls{ZMx}}
        \put(40.25, 30.5){\small\gls{phiZX}}
        \put(34, 3.5){\small\gls{ZMy}}
        \put(40.25, 6){\small\gls{phiZY}}
        \put(64, 24){\small\gls{Tx}}
        \put(64, 7.75){\small\gls{Ty}}
        \put(32, 24){\small$\gls{enc}_{\gls{X}}$}
        \put(32, 7.75){\small$\gls{enc}_{\gls{Y}}$}
        \put(96.5, 16){$=$}
        \put(76.2, 28.3){\scriptsize$\gls{Tx}(\gls{phiZX}(\gls{ZMx}))$}
        \put(76.4, 3.6){\scriptsize$\gls{Ty}(\gls{phiZY}(\gls{ZMy}))$}
        \put(96, 28.5){\gls{U}}

    \end{overpic}
    \caption[The \glsxtrlong{lcp}.]{The \glsfirst{lcp}.
    The unobservable manifolds \gls{Mx} and \gls{My} are embedded into the input spaces \gls{X} and \gls{Y} through \gls{phiX} and \gls{phiY}.
    We can observe the semantic relationship between these manifolds, denoted as \gls{C}, through a partial correspondence \gls{pi} defined between the input spaces.
    The encoding functions $\gls{enc}_{\gls{X}}$ and $\gls{enc}_{\gls{Y}}$ map the input spaces to the respective latent spaces \gls{ZX} and \gls{ZY},
    modifying the embedded manifolds and inducing a correlation between them through some transformation $\gls{transformation} \in \gls{transformationclass}$.
    The objective is to discover two specific transformations, $\gls{Tx}$ and $\gls{Ty}$, that allow the latent spaces \gls{ZX} and \gls{ZY} to be mapped into universal spaces $\gls{UX}$ and $\gls{UY}$. In the universal space \gls{U}, the latent manifolds embeddings must coincide: $\gls{Tx}(\gls{phiZX}(\gls{ZMx})) = \gls{Ty}(\gls{phiZY}(\gls{ZMy})) \subseteq \gls{U}$.
    }
    \label{fig:formalization}
\end{figure}

As discussed in \Cref{chap:introduction}, in machine learning and specifically within the context of \glsxtrshortpl{nn}, our observational capabilities are confined to the high-dimensional representations of underlying conceptual entities. Consider the notion of a “cat”, a conceptual entity that resides within an abstract manifold of meaning. What we perceive and process are not these abstract entities themselves, but their embedding within a higher-dimensional space -- namely, images of cats.
When we have semantic correspondences between two distinct data spaces, we are effectively observing an alignment between these high-dimensional spaces, e.g., between captions and images, and indirectly, the correspondence between the caption's meaning and the image's meaning. The crux of our exploration is anchored in the fact that two semantically related manifolds are similar (but not necessarily isomorphic, since different spaces may have different expressive power) and easily alignable, even by transformations that operate on the entire ambient spaces in which they are embedded.

In the following sections, we formalize the \glsfirst{lcp}, an illustration of which is shown in \Cref{fig:formalization}.

\section{Framework}\label{sec:framework}

\paragraph{Data notation.}
We denote input data spaces as \gls{X} and \gls{Y}, containing data points \gls{x} and \gls{y} respectively. We indicate with \gls{Mx} and \gls{My} their underlying abstract data manifolds, that contains data points denoted as \gls{mx} and \gls{my}.
Similarly, the symbols used to denote the latent spaces and the associated latent abstract manifolds are
$\gls{zx} \in \gls{ZX}$, $\gls{zy} \in \gls{ZY}$  and
$\gls{zmx} \in \gls{ZMx}$, $\gls{zmy} \in \gls{ZMy}$, respectively.
We use \gls{S} to indicate a generic space, e.g., the input space or the latent space~of~\glsxtrshortpl{nn}.

\paragraph{Manifold embedding.}
We consider data semantics to reside on unknown and unobservable low-dimensional abstract manifolds, denoted as \gls{M},
which are embedded\footnote{In the mathematical sense.} through the operation~$\gls{phi}_{\gls{S}}$ into observable high-dimensional spaces, denoted as \gls{S}:
\begin{equation}
    \gls{phi}_{\gls{S}}: \gls{M} \hookrightarrow \gls{S},
\end{equation}
where \gls{phi} maps data points from the manifold \gls{M} to the ambient space \gls{S}. The notation $\gls{phi}_{\gls{S}}(\gls{M})$ indicates the entirety of the \emph{manifold embedding} within \gls{S}.
Although multiple mappings \gls{phi} can embed \gls{M} into a generic high-dimensional space, for a specific configuration of \gls{S} where the embedding is already established (e.g., a dataset of images or a latent space), the mapping $\gls{phi}_{\gls{S}}$ that realizes \gls{M} within \gls{S} is unique.
The manifold embedding $\gls{phi}_{\gls{S}}(\gls{M})$ is precisely what the Manifold Hypothesis \citep{fefferman2016testing} refers to, suggesting that high-dimensional data observed in
\gls{S}  actually lies on or near this embedded lower-dimensional manifold, capturing the intrinsic geometry and essential characteristics of the data.

\paragraph{Semantic correspondence.}\label{sec:semantic_correspondence}
Given two data manifolds \gls{Mx} and \gls{My}, they are semantically related if there exists a partial correspondence
\begin{equation}\label{eq:manifoldcorrespondence}
    \gls{C} \subseteq \gls{Mx} \times \gls{My}
\end{equation}
between the two manifolds, such that $\forall (\gls{mx}, \gls{my}) \in \gls{C},\,$ \gls{mx} and \gls{my} are related by the same semantic relationship.
The correspondence $\gls{C}$ is an abstract relation between the manifolds, and it is not directly observable. However, we can observe a partial correspondence $\gls{pi}$ derived from \gls{C} and represented in the associated ambient spaces \gls{X} and \gls{Y}:
\begin{equation}\label{eq:inputcorrespondence}
    \gls{pi} \subseteq \{(\gls{phiX}(\gls{mx}_i), \gls{phiY}(\gls{my}_i)) \mid (\gls{mx}_i, \gls{my}_i) \in \gls{C}\}.
\end{equation}
One example of such correspondence $\gls{pi}$ involves images paired with one or more captions that describe them. Meanwhile, in \gls{C}, the corresponding elements associate their abstract meanings.

\paragraph{Latent Spaces.}\label{sec:latent_spaces}
We consider \glsxtrshortpl{nn} as parametric functions $\gls{N}^{\gls{theta}}$ compositions of \emph{encoding} and \emph{decoding} functions,  $\gls{N}^{\gls{theta}} = \gls{dec}^{\gls{theta}_2} \circ \gls{enc}^{\gls{theta}_1}$, where the encoder $\gls{enc}^{\gls{theta}_1}$ is responsible for computing a latent representation $\gls{zx} = \gls{enc}^{\gls{theta}_1}(\gls{x})$, $\gls{x} \in \gls{X}$ for some domain \gls{X}. This encoding function transforms the manifold embedded in the input space $\gls{phiX}(\gls{Mx})$ into a latent manifold embedding $\gls{phiZX}(\gls{ZMx})$, implicitly associated to some manifold \gls{ZMx}. This latent representation is then exploited to solve downstream tasks, such as classification, reconstruction or generation, optimizing over some objective function.
In the following, we will drop the dependence of parameters $\gls{theta}$ for notational convenience when not required, and indicate with $\gls{ZX}$ the latent space associated to $\gls{X}$. We will use the terms \emph{latent representation} and \emph{absolute representation} interchangeably, to refer to the output of the encoder.
Moreover, for each module \gls{enc} (equivalently for \gls{dec}), we indicate with $\gls{enc}_{\gls{X}}$ if the module \gls{enc} was trained on the domain \gls{X}.

\paragraph{Downstream task.}
We consider the decoder $\gls{dec}$ to be responsible for solving a generic downstream task $\gls{task}$ at hand (e.g., classification, generation, etc.). We indicate with $\gls{loss}^{\gls{task}}_{\gls{dec}}$
how well the decoder $\gls{dec}$ is performing on the task $\gls{task}$, i.e., the loss. Furthermore, we assume that the loss is computed on a test split, and the \gls{dec} is trained on a training split of the data.

Most importantly, we indicate with $\gls{loss}^{\gls{task}}(\gls{S})$ the lowest possible loss achievable by any decoder $\gls{dec}$ trained from scratch on \gls{S} to solve \gls{task}:
\begin{equation}
    \gls{loss}^{\gls{task}}(S) = \min_{\gls{dec} \in \gls{Dall}} \gls{loss}^{\gls{task}}_{D}(S),
\end{equation}
where $\gls{Dall}$ is the set of all possible decoders.

\section{Problem Statement}\label{sec:latentcommunicationproblem}

\subsection{Assumptions.}

\paragraph{Semantic Correspondence.}
We assume that the data manifolds \gls{Mx} and \gls{My} are related by a semantic correspondence $\gls{C}$, partially observable through $\gls{pi}$.
Moreover, we assume such correspondence is partially provided as \emph{parallel anchors} $\gls{parallelanchors} \subseteq \gls{pi}$.

\paragraph{Good Encoders.}
Throughout our work, we assume that the encoders $\gls{enc}_{\gls{X}}$ and $\gls{enc}_{\gls{Y}}$ are good. Formally, we can express this assumption as follows:
\begin{equation}
    \gls{loss}^{\gls{task}}(\gls{X}) = \gls{loss}^{\gls{task}}(\gls{ZX})
    \quad \text{and} \quad
    \gls{loss}^{\gls{task}}(\gls{Y}) = \gls{loss}^{\gls{task}}(\gls{ZY}),
\end{equation}
that is, the task $\gls{task}$ can be solved with the same performance on the input spaces \gls{X} and \gls{Y}, as well as on the respective latent spaces \gls{ZX} and \gls{ZY}.

In practice, this means that good encoders map data into the latent space without losing information useful for the task $\gls{task}$ (e.g., pre-trained universal feature extractors).

\paragraph{Emerging Similarities.}\label{sec:randomfactors}
As previously discussed, we argue that the learned latent spaces are not only a function of the data, the specific loss and the task; but in practice they are also affected by the optimization process used to train the network due to weight initialization, data shuffling, hyperparameters, data domain and other stochastic or non-semantic factors.
We denote these factors collectively by $\gls{randomfactors}$.

In particular, as shown in \Cref{introduction:fig:latent-rotation,relative:fig:latent-rotation-pca-proof} and widely observed in the literature (\Cref{sec:representation-similarity}), changing these factors induces some transformation $\gls{transformation}$ over the latent manifold embedding:
\begin{equation}
    \gls{randomfactors} \rightarrow  \gls{randomfactors}'
    \quad
    \text{ implies }
    \quad
    \gls{enc}^{\gls{theta}}(\gls{phi}(\gls{mx}))
    \rightarrow
    \gls{transformation}\gls{enc}^{\gls{theta}}(\gls{phi}(\gls{mx})),
    \quad
    \forall \gls{mx} \in \gls{M},
\end{equation}
where \gls{phi} is the embedding operation described in \Cref{sec:framework}.
We assume that these transformations fall into some unknown class of transformation $\gls{transformation} \in \gls{transformationclass}$ (e.g., orthogonal transformations),
when the variation factors are restricted to elements of~\gls{randomfactors}.

\subsection{Problem.}
We are given the input spaces \gls{X} and \gls{Y},
their associated abstract manifolds \gls{Mx} and \gls{My} in semantic correspondence $\gls{C} \subseteq \gls{Mx} \times \gls{My}$ observable through $\gls{pi}$, and
embedded in \gls{X} and \gls{Y} through the embedding functions $\gls{phiX}$ and $\gls{phiY}$;
and, two \glsxtrshortpl{nn} $\gls{N}_{\gls{X}} = \gls{dec}_{{\gls{X}}} \circ \gls{enc}_{{\gls{X}}}$,
$\gls{N}_{\gls{Y}} = \gls{dec}_{{\gls{Y}}} \circ \gls{enc}_{{\gls{Y}}}$
trained on \gls{X} and \gls{Y}, respectively, to solve the task $\gls{task}$.

Our objective is to unify the latent manifolds embeddings
$\gls{phiZX}(\gls{ZMx})$
and
$\gls{phiZY}(\gls{ZMy})$
into a universal space $\gls{U}$, by finding
$\gls{Tx}: \gls{ZX} \to \gls{UX}$
and
$\gls{Ty}: \gls{ZY} \to \gls{UY}$:
\begin{gather}
    \forall (\gls{mx}, \gls{my}) \in \gls{C},
    \quad
    \gls{Tx}(\gls{enc}_{\gls{X}}(\gls{phiX}{(\gls{mx})}))
    =
    \gls{Ty}(\gls{enc}_{\gls{Y}}(\gls{phiY}{(\gls{my})}))
    \subseteq \gls{U} \nonumber\\
    \text{such that} \\
    \gls{loss}^{\gls{task}}(\gls{ZX}) = \gls{loss}^{\gls{task}}(\gls{UX})
    \quad \text{and} \quad
    \gls{loss}^{\gls{task}}(\gls{ZY}) = \gls{loss}^{\gls{task}}(\gls{UY}). \nonumber
\end{gather}
Note that, the transformations are constrained to align only the latent manifold embeddings $\gls{phiZX}(\gls{ZMx})$ and
$\gls{phiZY}(\gls{ZMy})$, not necessarily requiring alignment of the entire spaces \gls{ZX} and \gls{ZY}.
In practice, this implies we are trying to find transformations of the latent ambient spaces that align as best as possible the manifolds embedded in them, without losing information useful for the task $\gls{task}$.

\section{Corollary problems}\label{chap:corollaryproblems}
Solving the general \glsxtrshortpl{lcp} allows us to directly address several corollary tasks, which are of extreme practical interest. In the following sections, we describe how the \gls{lcp} enables the reuse of neural components (\Cref{sec:stitchingdefinition}), a downstream performance evaluation directly in the latent space (\Cref{sec:latentcomparisondefinition}), and the development of advanced retrieval systems (\Cref{sec:retrievaldefinition}).

\subsection{Zero-Shot Stitching} \label{sec:stitchingdefinition}
Solving the \gls{lcp} defined in \Cref{sec:latentcommunicationproblem} enables zero-shot interoperability of pre-trained neural components. In previous works, such as \cite{Lenc2014-gy,Bansal2021-oj}, stitching layers are \emph{trainable} linear projections that allow comparing the representations of different networks.
Instead, our framework unlocks the possibility of \emph{Zero-Shot Stitching} different neural components, treating them as frozen black-box modules.

We define a generic \emph{stitched model} as the composition of an encoder, that embeds data, plus an independent decoder specialized in a downstream task (e.g., classification, reconstruction):
\begin{equation}
    \gls{N}_{\gls{X}\gls{Y}} = \gls{dec}_{\gls{Y}} \circ \gls{enc}_{\gls{X}}.
\end{equation}
The stitching operation is always performed without training or fine-tuning, in a zero-shot fashion.

In \Cref{chap:rl,relative:sec:model-reusability,translation:sec:latent-translation,bridge:sec:exp-downstream-task,bootstrapping:sec:stitching},
we show that the latent communication framework allows us to stitch together independent neural components, demonstrating empirically that re-using neural components is possible without the necessity for extensive retraining or fine-tuning.

\subsection{Latent Model Evaluation} \label{sec:latentcomparisondefinition}
Unifying the latent spaces into universal spaces allows us to compare the latent spaces across variations of the factors \gls{randomfactors}.
Interestingly, in \Cref{relative:rebase:sec:manifold-performance} we show that solving the \gls{lcp} implies having a quantitative latent measure of downstream performance, provided that a reliable reference model is available.
This measure does not require any labeled data and correlates with standard downstream performance measures. Consequently, we can assess the quality of a model directly, potentially during training, without the need to solve the downstream task explicitly.

\subsection{Retrieval} \label{sec:retrievaldefinition}
The solution to the \gls{lcp} facilitates the development of advanced retrieval systems that leverage independently computed representations. This allows for the retrieval of data points from one space using queries from another space, without the necessity for a shared training set, we showcase it in
\Cref{relative:sec:word-embeddings,bootstrapping:sec:retrieval}.
Finally, we demonstrate that solving the \gls{lcp} also enables zero-shot captioning, by retrieving images using text queries, and vice versa, without any multimodal model training, as we show in \Cref{chap:asif}.

\Chapter{Universal Representations}{Relative representations enable zero-shot latent space communication\footnote{\fullcite{moschella2023}}}
\label{chap:relative}

\begin{quotation}
    \noindent
    In this Chapter, we tackle the \gls{lcp} defined in \Cref{chap:problemformalization} with the additional assumption that the latent manifolds embeddings $\gls{phiZX}(\gls{ZMx}) \subseteq \gls{ZX}$ and $\gls{phiZY}(\gls{ZMy}) \subseteq \gls{ZY}$ are always approximately related by a transformation $\gls{transformation} \in \gls{transformationclass}$, where \gls{transformationclass} is the class of transformations that preserve angle norms. Referring to \Cref{fig:formalization}, we define analytically and independently $\gls{Tx}$ and \gls{Ty} as parameter-free \emph{relative projections}, that implicitly unify the latent manifold embeddings in \gls{U}.
\end{quotation}

\section{Introduction}

In \Cref{chap:introduction}, we discussed how the learned latent spaces are subject to changes even when the factors \gls{randomfactors} remain fixed.  We illustrated this phenomenon in \Cref{introduction:fig:latent-rotation} with a toy example on a bi-dimensional \gls{ae}, and formalized the problem of unifying them in~\Cref{chap:problemformalization}.

\begin{figure}[ht]
    \centering
    \begin{overpic}[trim=0cm 0cm 0cm 0cm,clip,width=\linewidth]{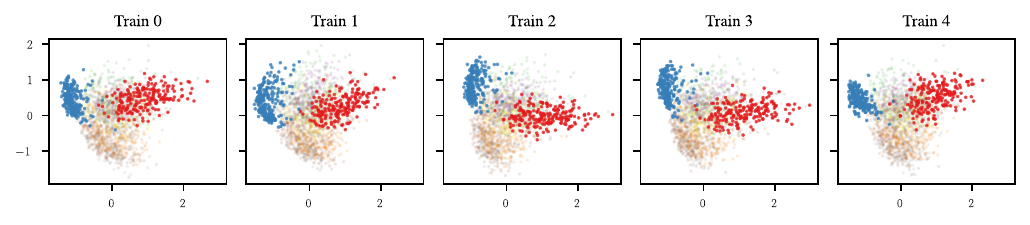}
    \end{overpic}
    \begin{overpic}[trim=0cm 0cm 0cm 0cm,clip,width=\linewidth]{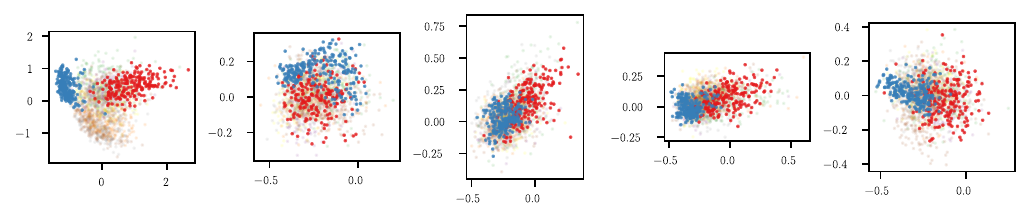}
    \end{overpic}
    \caption[Latent spaces learned by distinct trainings of an high-dimensional AE]{Latent spaces learned by distinct trainings of the same high-dimensional AE on the \gls{mnist} dataset. Each column is the latent space obtained by the AE with a different seed.
        On the first row, the dimensionality reduction is performed through PCAs fitted independently on each latent space, meanwhile, on the second row PCA is fitted on the leftmost latent space and then applied to all of them.
    }
    \label{relative:fig:latent-rotation-pca-proof}
\end{figure}

In \Cref{relative:fig:latent-rotation-pca-proof}, we further illustrate the phenomenon, exploiting the properties of \gls{pca} to demonstrate it also happens on high-dimensional latent spaces. Indeed, the \emph{second row} of the figure proves that latent spaces learned by distinct trainings of the same high-dimensional AE are extrinsically different; since \gls{pca} fitted on one latent space and applied to the others does not align them (up to rotations and reflections). This extrinsic difference is a significant challenge in addressing any of the tasks outlined in \Cref{chap:corollaryproblems}; for instance, it hinders any form of reuse or comparison between neural components trained on different embeddings of the same data, since they are incompatible. Nevertheless, the \emph{first row}  in \Cref{relative:fig:latent-rotation-pca-proof} shows that the  high-dimensional latent spaces, although extrinsically different, are \emph{intrinsically similar}, as the \gls{pca} fitted independently on each latent space produces similar results.

Motivated by these empirical observations, in this Chapter, we address the \gls{lcp} defined in \Cref{chap:problemformalization} with an additional assumption that the latent manifolds embeddings $\gls{phiZX}(\gls{ZMx}) \subseteq \gls{ZX}$ and $\gls{phiZY}(\gls{ZMy}) \subseteq \gls{ZY}$ are always approximately related by a transformation $\gls{transformation} \in \gls{transformationclass}$, where \gls{transformationclass} is the class of transformations that preserve angle norms. To tackle this simplified problem, we suggest adopting a local coordinate system defined by the data itself. Data points in the latent space becomes a set of coefficients that encode the point as a function of other data samples, instead of an independent point in $\gls{Rd}$. The proposed \emph{\glsfirst{rr}} directly encodes the intrinsic information underlying the data, and with an appropriately chosen similarity function (e.g., cosine similarity), depends solely on the angles norms between embeddings by construction; de facto infusing an invariance to angle norm preserving transformations in the latent space that unifies them.

We show how neural architectures can leverage these \glsxtrshortpl{rr} to guarantee, in practice, invariance to latent isometries and local rescalings, enabling a variety of applications from zero-shot model \Cref{sec:stitchingdefinition} stitching to latent space comparison \Cref{sec:latentcomparisondefinition} between diverse settings.
Remarkably, this enables a form of compositionality between learning models; it allows, for instance, to stitch together an encoder trained on \gls{imagenet1k} with a decoder trained on \gls{cifarh}, as we showcase in our experiments. We extensively validate the generalization capability of our approach on different datasets, spanning various modalities (images, text, graphs), tasks (e.g., classification, reconstruction) and architectures (e.g., \glsxtrshortpl{cnn}, \glsxtrshortpl{gnn}, transformers).

The main contributions can be summarized as follows:
\begin{itemize}
    \itemsep0em
    \item We show that the representations learned by \glsxtrshortpl{nn} are subject to change due to several training factors; {nonetheless, the norm of the angles between latent embeddings are often preserved}.
    \item We introduce a novel {relative representation} for latent embeddings, that is {invariant by construction to the} {transformations} induced by the factors \gls{randomfactors}.
    \item For the first time, we successfully demonstrate \emph{Zero-Shot Stitching} (\Cref{sec:stitchingdefinition}) of neural components produced by distinct training regimens, e.g., due to different seeds or different neural architectures; we validate our findings on different data modalities (e.g., images, text).
    \item Our framework also provides a \emph{quantitative} latent measure of performance (\Cref{sec:latentcomparisondefinition}) while training neural models, which is differentiable, does not need any labeled data, and is correlated with standard downstream performance measures such as accuracy.
\end{itemize}

\section{Relative Representations}\label{relative:sec:relative-representations}

\paragraph{Assumption.}
{In this Chapter, we make the core assumption that \gls{transformationclass} is the class of transformations that preserve the norm of the angles between elements of the latent space, namely $|\angle(\gls{zx}_{i},\gls{zx}_{j})| = |\angle(\gls{transformation} \gls{zx}_{i},\gls{transformation} \gls{zx}_{j})|$ for every $(\gls{x}_{i},\gls{x}_{j}) \in \gls{X}$. By ``angle norm'', we mean the absolute value of the angle between two elements, which ensures that global reflections do not change the sign of the considered similarity.
    While this assumption might seem too restrictive, in practice it arises in several real scenarios, as we show in the following sections.
    Indeed, in classification tasks, this assumption is further supported by \Cref{translation:fig:rescaled-layer-acc,translation:fig:scale-inv-mono} which show that the embeddings scale does not affect classification performance. Therefore, only the angle between embeddings is relevant. Additionally, in \Cref{chap:bridge}, we completely remove this assumption.
}

\begin{figure}[ht]
    \centering
    \begin{minipage}{0.51\linewidth}
        \centering
        \begin{tikzpicture}
            \begin{axis}[
                    axis lines=center,
                    view={60}{40}, %
                    domain=0:4,
                    y domain=0:4,
                    zmin=0, zmax=4,
                    enlargelimits=upper,
                    colormap/viridis,
                    samples=50
                ]

                \addplot3 [surf, opacity=0.5]
                {sin(deg(sqrt(x^2+y^2))) + 2};

                \addplot3 [only marks, mark=*, red] coordinates {
                        (1,1,1.2) %
                        (2,2,0.5) %
                        (3,3,2.5) %
                    };

                \addplot3 [only marks, mark=*, blue] coordinates {
                        (1.5,1.5,1.7) %
                    };

                \node[coordinate,pin=left:{\(a_1\)}] at (axis cs:1,1,1.2) {};
                \node[coordinate,pin=right:{\(a_2\)}] at (axis cs:1.8,1.8,0.4) {};
                \node[coordinate,pin=above:{\(a_3\)}] at (axis cs:3,3,2.5) {};
                \node[coordinate,pin=below:{\(x\)}] at (axis cs:1.5,1.5,1.7) {};

            \end{axis}
        \end{tikzpicture}
    \end{minipage}
    \hfill
    \begin{minipage}{0.48\linewidth}
        \centering
        \tdplotsetmaincoords{60}{110} %
        \begin{tikzpicture}[tdplot_main_coords]
            \begin{axis}[
                    axis lines=center,
                    xlabel={$d(x, a_1)$},
                    ylabel={$d(x, a_2)$},
                    zlabel={$d(x, a_3)$},
                    every axis x label/.style={
                            at={(ticklabel* cs:1.05)},
                            anchor=west,
                        },
                    every axis y label/.style={
                            at={(ticklabel* cs:1.05)},
                            anchor=west,
                        },
                    every axis z label/.style={
                            at={(ticklabel* cs:1.15)},
                            anchor=west,
                        },
                    xtick=\empty,
                    ytick=\empty,
                    ztick=\empty,
                    xmin=0, xmax=4,
                    ymin=0, ymax=4,
                    zmin=0, zmax=4
                ]

                \addplot3 [only marks, mark=*, mark options={color=blue}] coordinates {(1,2,3)};

                \addplot3 [dotted, thick] coordinates {(0,2,3) (1,2,3)};
                \addplot3 [dotted, thick] coordinates {(1,0,3) (1,2,3)};
                \addplot3 [dotted, thick] coordinates {(1,2,0) (1,2,3)};
                \addplot3 [dotted, thick] coordinates {(0,0,3) (0,2,3)};
                \addplot3 [dotted, thick] coordinates {(0,2,0) (0,2,3)};
                \addplot3 [dotted, thick] coordinates {(0,0,3) (1,0,3)};
                \addplot3 [dotted, thick] coordinates {(1,0,0) (1,0,3)};
                \addplot3 [dotted, thick] coordinates {(1,2,0) (1,0,0)};
                \addplot3 [dotted, thick] coordinates {(0,2,0) (1,2,0)};

                \addplot3 [only marks, mark=*, mark options={color=red}] coordinates {
                        (0,2,3) %
                        (1,0,3) %
                        (1,2,0) %
                    };

            \end{axis}

        \end{tikzpicture}
    \end{minipage}
    \caption[\glsxtrlong{rr}]{\glsfirst{rr}.
    {\em (left)}: a sample $x$ and three anchor samples $a_1, a_2, a_3$ are embedded in a latent space and lie on the underlying embedded data manifold.
        {\em (right)}: each dimension is treated as coefficients in a coordinate system defined by the anchors, the new representation of $x$ is given by its similarities with respect to the anchors. Anchors are orthogonal in this example only for visualization purposes.}
    \label{relative:fig:relative-distances}
\end{figure}
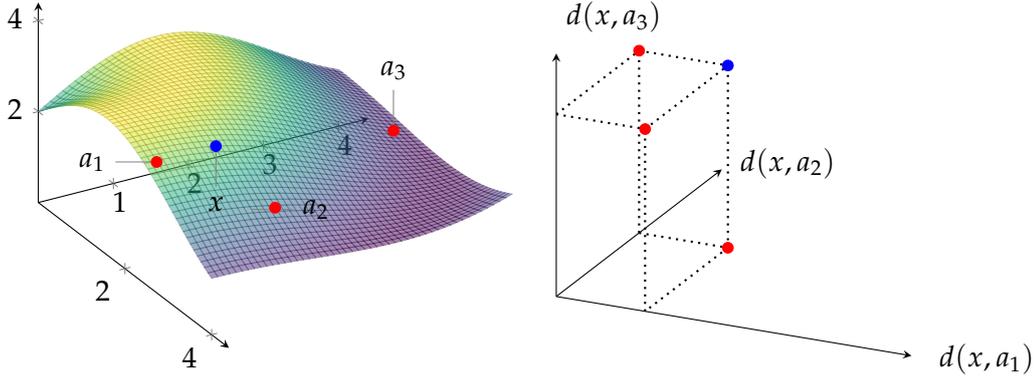

\paragraph{Method.}
To build our representation, we start by selecting a subset $\gls{anchors}$ of the training data \gls{X}, which we denote as \emph{anchors}.
Every sample in the training distribution will be represented with respect to the embedded anchors
$\gls{za}_{j} = \gls{enc}(\gls{a}_{j}) \text{ with } \gls{a}_{j} \in \gls{anchors}$.
As a measure capturing the relation between the anchors and the other samples, we consider a generic similarity function $\gls{sim}: \gls{Rd} \times  \gls{Rd} \rightarrow \gls{R}$, yielding a scalar score between two absolute representations $\gls{sim}(\gls{zx}_i, \gls{zx}_j).$
Given the anchors $\gls{anchors}$ in an arbitrary ordering $\gls{a}_1, \dots, \gls{a}_{|\gls{anchors}|}$, we define the \gls{rr} of $\gls{x}_i \in \gls{X}$ as:
\begin{align}
    \gls{r}_{\gls{x}_i} =
    (
    \gls{sim}(\gls{zx}_{i}, \gls{za}_{1}),
    \gls{sim}(\gls{zx}_{i}, \gls{za}_{2}),
    \dots,
    \gls{sim}(\gls{zx}_{i}, \gls{za}_{|\gls{anchors}|})
    )  \,,
\end{align}
for convenience, we equivalently define the relative projection function \gls{relativeprojection}:
\begin{equation}\label{relative:eq:relative-projection}
    \gls{relativeprojection}(\gls{zx}; \gls{ZA}, \gls{sim})
    =
    \bigoplus_{\gls{za}_i \in \gls{ZA}} \gls{sim}( \gls{zx}, \gls{za}_i)
\end{equation}
where $\bigoplus$ denotes row-wise concatenation and all embeddings are produced by the same encoding function $\gls{enc}$.
For notational convenience, we denote the relative projection of a set of samples
\gls{ZX} with $\gls{relativeprojection}(\gls{ZX};\gls{ZA},\gls{sim})$, which is defined as the collection of relative projections of individual samples $\gls{zx} \in \gls{ZX}$.
\Cref{relative:fig:relative-distances} illustrates the key differences between absolute and \glsxtrshortpl{rr}.

\paragraph{Choice of the anchors.}
Anchors directly affect the expressivity of the \gls{rr} space, and are related to the task at hand. For example, in a classification task, we should sample anchors from each class in the training set, in order to well represent each data sample in \gls{X}. We refer to \Cref{relative:sec:anchor-num} for an analysis of different anchor selection strategies.

\paragraph{Parallel anchors.} \label{sec:parallelanchors}
One case of interest arises when the data comes from different domains or modalities $\gls{X} \neq \gls{Y}$, and we are given a partial correspondence \gls{pi}, as defined in \Cref{sec:semantic_correspondence}.
In this case, we can obtain \emph{parallel anchors}\footnote{We use the term ``parallel'' to indicate they represent the same underlying meaning.} by sampling simultaneously from both domains \gls{X} and \gls{Y}:
\begin{equation}
    \gls{parallelanchors} \subseteq \gls{pi}
\end{equation}
We show an example of parallel anchors in \Cref{relative:sec:nlp-app}, where \gls{X} and \gls{Y} are Amazon reviews in two different languages; illustrate a strategy to automatically expand this correspondence in \Cref{chap:bootstrapping}; and further explore a multimodal application in~\Cref{chap:asif}.

\paragraph{Out of domain anchors.}\label{relative:sec:ood-anchors}
Surprisingly, the choice of the anchors is not restricted to elements in the training distribution. Given an encoder pre-trained on a fixed training distribution, we can pick elements from a set $\gls{OODanchors}$ that is out-of-domain w.r.t. \gls{X}, and build the \glsxtrshortpl{rr} on top of $\gls{OODanchors}$. We refer to these as \emph{OOD anchors} and exploit them, e.g., to solve domain adaptation tasks where we do not have access to a correspondence, and have scarce data labels. We refer to the \Cref{relative:sec:nlp-app,relative:sec:cv-app} for real-world examples.

\paragraph{Universal Representations.}
In this work, we choose the cosine similarity as the similarity function due to the properties it induces on the \gls{rr}.
The cosine similarity is the dot product of unit vectors, corresponding to the cosine of the angle $\cos\theta$ between the two.
Importantly, $\cos\theta$ does not change if we apply the same angle-norm preserving transformation \gls{transformation} to them,
i.e., the cosine similarity is invariant to rotations, reflections, and independent rescaling of each point.
While this is not true for translations, \glsxtrshortpl{nn} commonly employ normalization techniques (e.g., InstanceNorm \citep{instance-norm}) to center the latent spaces. Under this assumption, cosine similarity guarantees  \glsxtrshortpl{rr} invariant also to translations. %

    {This means we have the freedom to change the embedding function $\gls{enc}$ with any other function $\gls{enc}'$ that produces} {different representations with same angles, i.e.:
        \begin{align}
            [
                \gls{sim}(\gls{zx}_i, \gls{za}_{1}),
                \dots,
                \gls{sim}(\gls{zx}_i, \gls{za}_{|\gls{anchors}|})
            ]
            =
            [
            \gls{sim}( \gls{transformation} \gls{zx}_i,  \gls{transformation}{\gls{za}}_1),
            \dots,
            \gls{sim}( \gls{transformation} \gls{zx}_i, \gls{transformation}{\gls{za}}_{|\gls{anchors}|} )
            ]
        \end{align}
        where $\gls{sim}$ is the cosine similarity and \gls{transformation}, induced by $\gls{enc}'$, is an arbitrary {angle-norm preserving} transformation.} %

An implication of this invariance is that we can
solve the \gls{lcp} (defined in \Cref{chap:problemformalization}), simplified by the assumption that \gls{transformationclass} is the class of angle-norm preserving transformations. Indeed, we can define $\gls{Tx}$ and $\gls{Ty}$ as independent relative projections over parallel anchors \gls{parallelanchors}:
\begin{align}
    \gls{Tx}(\gls{zx}) & = \gls{relativeprojection}(\gls{zx};\, \gls{ZA}_{\gls{X}},\, cosine) &  & \forall\, \gls{zx} \in \gls{ZX}                \\
    \gls{Ty}(\gls{zy}) & = \gls{relativeprojection}(\gls{zy};\, \gls{ZA}_{\gls{Y}},\, cosine) &  & \forall\, \gls{zy} \in \gls{ZY}  \,, \nonumber
\end{align}
these transformations are enough to unify the latent manifold embeddings:
\begin{equation}
    \gls{Tx}(\gls{phiZX}(\gls{ZMx})) = \gls{Ty}(\gls{phiZY}(\gls{ZMy})) \subseteq \gls{U} \,,
\end{equation}
as we demonstrate empirically in \Cref{relative:sec:latent-communication,relative:sec:model-reusability}.
Additionally, as we demonstrate in \Cref{relative:sec:abs-vs-rel}, the original task can be solved in this universal space with a comparable performance.

We remark that other choices of similarity function can be made to enforce different invariances into the representation, refer to \Cref{chap:bridge} for an extensive exploration of this aspect.

\section{Latent Evaluation}\label{relative:sec:latent-communication}
In this Section, we demonstrate how \glsxtrshortpl{rr} can effectively be used to produce latent spaces that are stable under a variety of factors \gls{randomfactors} as described in \Cref{sec:latentcomparisondefinition}. To remark, our main question is the following: Given two different learning models that are trained independently, can we compare their latent embeddings?
We answer in the positive, showing the gained invariance enables effective communication between different, but semantically equivalent latent spaces.

In particular, we analyze how different word embedding spaces, once projected onto \glsxtrshortpl{rr}, are intrinsically the same (\Cref{relative:sec:word-embeddings}); we then show how the similarity between the relative counterparts of two or more embedding spaces is a surprisingly good predictor of model downstream performance (\Cref{relative:rebase:sec:manifold-performance}); finally, we confirm that \glsxtrshortpl{rr} in the training phase are not detrimental to performance (\Cref{relative:sec:abs-vs-rel}).

\subsection{Word Embeddings}\label{relative:sec:word-embeddings}
\paragraph{Experimental setting.}
We select two different word embeddings on the English language, namely \gls{fasttext} and \gls{word2vec}. Both models are pre-trained on different data, but partly share a vocabulary from which we extract $\approx20$K words. Using $300$ randomly drawn parallel anchor, we convert each embedding space to a relative one.
In \Cref{relative:fig:latent-rotation-comparison} (left), we show the original and the relative embeddings.
    {
        For each word $w$, we consider its corresponding encodings $x$ and $y$ in the source and target space. We apply three different metrics to measure their similarity (in a setting similar to \cite{vulic-etal-2020-good}):
        (i) \emph{Jaccard}: the discrete Jaccard similarity between the set of word neighbors of $x$ in source and target;
        (ii) \emph{Mean Reciprocal Rank}: measures the (reciprocal) ranking of $w$ among the top-k neighbors of $x$ in the target space;
        (iii) \emph{Cosine}: measures the cosine similarity between $x$ and $y$.}
Additional details in \Cref{relative:appendix:word-embeddings}.

\begin{table}[ht]
    \caption[Similarity across word embeddings in absolute and relative spaces]{Qualitative \emph{(left)} and quantitative \emph{(right)} comparisons of English word embeddings using absolute and \glsxtrshortpl{rr}. {PCA is applied only for visualization}. All metrics are calculated with $K=10$  {averaged over 20k words and across 10 different random seeds. See \Cref{relative:fig:word-embedding-proj} for other dimensionality reductions, refer to \Cref{relative:tab:quantitative-analysis-cv-all} and \Cref{relative:fig:qualitative-retrieval-cv} for the same experiment on \gls{cifart}, showcasing this result also holds on different data modalities.}}
    \label{relative:fig:latent-rotation-comparison}
    \noindent\makebox[\textwidth][c]{%
        \hspace{1cm}
        \begin{minipage}{.29\textwidth}
            \begin{overpic}[trim=-1cm 1cm -0.5cm 0cm,width=1\linewidth]{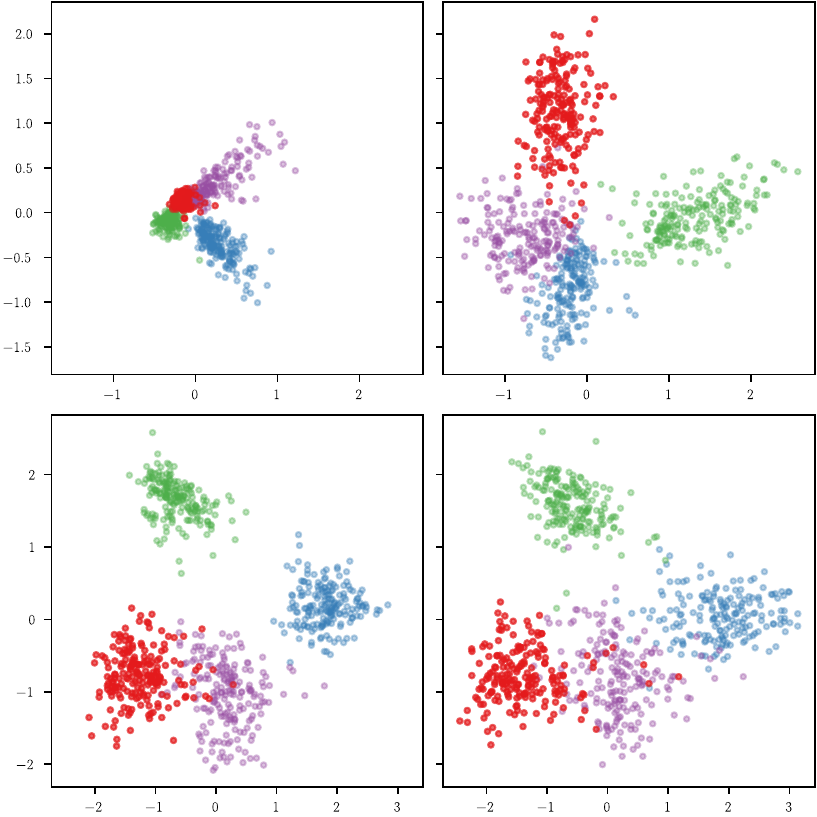}
                \put(11.5, 88){\gls{fasttext}}
                \put(55.5, 88){\gls{word2vec}}
                \put(-1, 46){\rotatebox{90}{\small Absolute}}
                \put(-1, 2){\rotatebox{90}{\small Relative}}
            \end{overpic}
        \end{minipage}
        \hspace{0.1cm}
        \begin{minipage}{.70\textwidth}
            \footnotesize
            \begin{tabular}{llllll}
                \toprule
                                                                             & Source                          & Target         & \multicolumn{1}{c}{Jaccard ↑} & \multicolumn{1}{c}{MRR ↑} & \multicolumn{1}{c}{Cosine ↑} \\
                \midrule
                \multirow{4}{*}{{\STAB{\rotatebox[origin=c]{90}{Absolute}}}} & \multirow{2}{*}{\texttt{{FT}}}  & \texttt{{FT}}  & $1.00 \pm 0.00$               & $1.00 \pm 0.00$           & $1.00 \pm 0.00$              \\
                                                                             &                                 & \texttt{{W2V}} & $0.00 \pm 0.00$               & $0.00 \pm 0.00$           & $0.01 \pm 0.00$              \\[0.5ex]
                                                                             & \multirow{2}{*}{\texttt{{W2V}}} & \texttt{{FT}}  & $0.00 \pm 0.00$               & $0.00 \pm 0.00$           & $0.01 \pm 0.00$              \\
                                                                             &                                 & \texttt{{W2V}} & $1.00 \pm 0.00$               & $1.00 \pm 0.00$           & $1.00 \pm 0.00$              \\
                \cmidrule(lr){2-6}
                \multirow{4}{*}{\STAB{\rotatebox[origin=c]{90}{Relative}}}   & \multirow{2}{*}{\texttt{{FT}}}  & \texttt{{FT}}  & $1.00 \pm 0.00$               & $1.00 \pm 0.00$           & $1.00 \pm 0.00$              \\
                                                                             &                                 & \texttt{{W2V}} & $0.34 \pm 0.01$               & $0.94 \pm 0.00$           & $0.86 \pm 0.00$              \\[0.5ex]
                                                                             & \multirow{2}{*}{\texttt{{W2V}}} & \texttt{{FT}}  & $0.39 \pm 0.00$               & $0.98 \pm 0.00$           & $0.86 \pm 0.00$              \\
                                                                             &                                 & \texttt{{W2V}} & $1.00 \pm 0.00$               & $1.00 \pm 0.00$           & $1.00 \pm 0.00$              \\
                \bottomrule
            \end{tabular}
        \end{minipage}
    }
\end{table}

\paragraph{Result analysis.}
\Cref{relative:fig:latent-rotation-comparison} \emph{(left)} highlights clusters of semantically similar words and shows that the absolute representations are incoherent across the two latent spaces, while the relative embeddings are highly similar.
The average Jaccard distance reported in \Cref{relative:fig:latent-rotation-comparison} \emph{(right)}, says that the {word} neighborhoods of the \glsxtrshortpl{rr} are matched exactly 34\% of the time in one direction, and 39\% of the time in the other one (the missing 61\% is due to semantic differences, that are not taken into account by the discrete nature of the Jaccard metric). By contrast, the absolute embeddings are never matched exactly (Jaccard score equal to zero); for a match to happen, it would mean that the \gls{fasttext} and \gls{word2vec} embeddings of a given English word are almost the same, which is highly unlikely.
    {MRR, close to a perfect score for the \glsxtrshortpl{rr}, shows that the most-similar word to a given one is usually itself, even if their cosine similarity doesn't reach 1.}

Overall, these results show that \glsxtrshortpl{rr} are preserved across different word embedding models, {validating our assumptions.}

\subsection{Latent distance as a performance proxy}\label{relative:rebase:sec:manifold-performance}

\paragraph{Experimental setting.}
In this experiment, we consider a node classification task on the \gls{cora} graph dataset. We first train a \emph{reference} model that achieves good accuracy on a validation set. Then, we train $\approx2000$ models with various combinations of seed, number of epochs, number of layers, dropout probability, activation functions, optimizer type, learning rate or type of graph embedder  (refer to \Cref{relative:tab:manifold-performance-parameters} for further details).
All the models are trained using absolute representations, which are converted to relative post-training
by projecting the embeddings onto $300$ randomly drawn but fixed anchors. For each model, we measure its classification accuracy and compute the similarity of its space with the reference one. This similarity is computed as the average cosine similarity between the node embeddings produced by a given model and the corresponding embeddings in the reference one. %

\begin{figure}[ht]
    \centering
    \begin{overpic}[trim=-.5cm -.4cm -.5cm 0cm,clip,width=.73\linewidth]{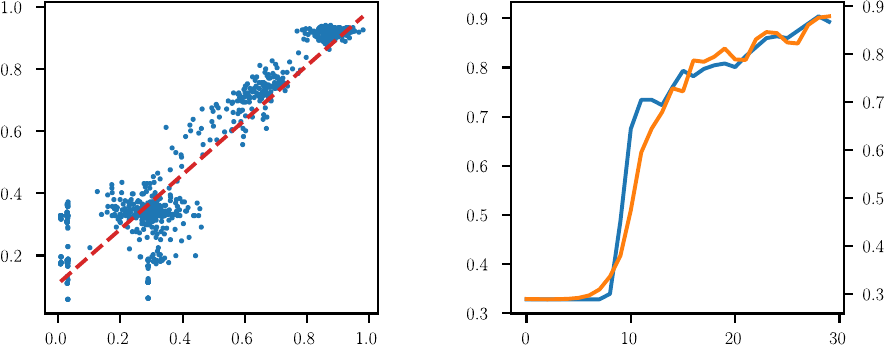}
        \put(-1, 13){\rotatebox{90}{Performance}}
        \put(18, -1){Similarity}

        \put(48, 13){\rotatebox{90}{\color{pltblue}Performance}}
        \put(70, -1){Epochs}
        \put(97.5, 16){\rotatebox{90}{\color{pltorange}Similarity}}
    \end{overpic}
    \caption[Graph node classification task on \gls{cora}]{Graph node classification task on \gls{cora}. \emph{Left:} Correlation between the performance of $\approx2000$ models and the similarity of their latent spaces with respect to a well-performing reference model.
        \emph{Right:} The same correlation plotted over time. The mean Pearson correlation over all models is $0.955$, after filtering out the models having the best validation accuracy below $0.5$.}
    \label{relative:fig:data-manifold}
\end{figure}
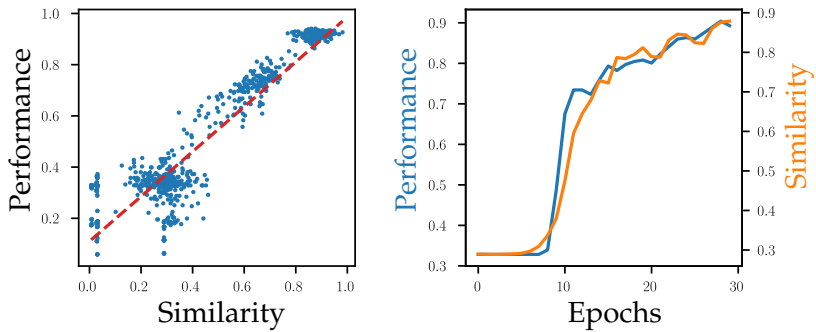

\paragraph{Result analysis.}
The scatter plot in \Cref{relative:fig:data-manifold} \emph{(left)} shows that better-performing models tend to be the ones with the latent spaces most similar to the reference model. The performance-similarity correlation also holds over time, as shown in \Cref{relative:fig:data-manifold} \emph{(right)}. %
Additional correlation examples are in \Cref{relative:fig:correlation-grid}.
Interestingly, this metric is differentiable, enabling an explicit supervision signal on the latent space, which does not require labeled data and could be readily exploited in a teacher-student framework.

Overall, these results suggest that the similarity between the \glsxtrshortpl{rr} of latent spaces is a remarkably good proxy to evaluate model performance.

\subsection{Training with Absolute vs. Relative representations}\label{relative:sec:abs-vs-rel}
\paragraph{Experimental setting.}
Finally, we compare architectures that do or do not employ the \gls{rr} while training. In these experiments, the models vary slightly according to the dataset; however, the relative and absolute versions are always comparable in terms of architecture, number of learnable parameters and hyperparameters. We refer to \Cref{relative:sec:implementatin-details} and the open-source code for further details on their implementation.
In this Section, we consider classification tasks on several datasets, spanning the image domain   \citep{mnist,Xiao2017-er,Krizhevsky2009-hv} and the graph domain  \citep{Planetoid}.

\begin{table}[ht]
    \caption[Performance comparison between relative and absolute representations]{Performance comparison between relative and absolute representations on various image and graph datasets. The metric is the classification weighted F1 score ($\pm$ std), over 6 seeds.}
    \label{relative:tab:abs-rel-performance-comparison}
    \centering
    \resizebox{\textwidth}{!}{%
        \begin{tabular}{ccccccccccccc}
            \toprule
                              & \multicolumn{4}{c}{\bf Image Classification} & \multicolumn{3}{c}{\bf Graph Node Classification}                                                                                             \\
            \cmidrule(lr){2-5}\cmidrule(lr){6-8}
            \multicolumn{1}{c}{}
                              & \multicolumn{1}{c}{\bf \gls{mnist}}
                              & \multicolumn{1}{c}{\bf \glsxtrshort{fmnist}}
                              & \multicolumn{1}{c}{\bf \gls{cifart}}
                              & \multicolumn{1}{c}{\bf \gls{cifarh}}
                              & \multicolumn{1}{c}{\bf \gls{cora}}
                              & \multicolumn{1}{c}{\bf \gls{citeseer}}
                              & \multicolumn{1}{c}{\bf \gls{pubmed}}
            \\ \midrule
            \textbf{Relative} & $97.91 \pm 0.07$                             & $90.19 \pm 0.27$                                  & $87.70 \pm 0.09$ & $66.72 \pm 0.35$ & $0.89 \pm 0.02$ & $0.77 \pm 0.03$ & $0.91 \pm 0.01$ \\
            \textbf{Absolute} & $97.95 \pm 0.10$                             & $90.32 \pm 0.21$                                  & $87.85 \pm 0.06$ & $68.88 \pm 0.14$ & $0.90 \pm 0.01$ & $0.78 \pm 0.03$ & $0.91 \pm 0.01$ \\
            \bottomrule
        \end{tabular}
    }
\end{table}

\paragraph{Result analysis.}
The results, reported in \Cref{relative:tab:abs-rel-performance-comparison}, show that \glsxtrshortpl{rr}, when used at training time, are not detrimental to performance in general. This is further shown in
\Cref{%
    relative:tab:cross-train-performance-comparison,%
    relative:tab:cross-model-monolingual,%
    relative:tab:cifar100-imagenet,%
    relative:tab:multilingual-full-coarse-grained,%
    relative:tab:multilingual-en-coarse-grained,%
    relative:tab:multilingual-full-fine-grained,%
    relative:tab:xmlr-multilingual-full-fine-grained,%
    relative:tab:cifar100-fine},
where a subset of the results compares the absolute and \glsxtrshortpl{rr} on a variety of domains, datasets, and tasks.
While the information relevant to the machine learning task seems to be preserved, an intriguing future research question is to determine what specific information is lost when infusing specific invariances to unify the representations.%

Overall, these results show that \glsxtrshortpl{rr} are effective when involved in end-to-end training, without significant performance drops in the downstream task.

\section{Zero-Shot Model Stitching}\label{relative:sec:model-reusability}

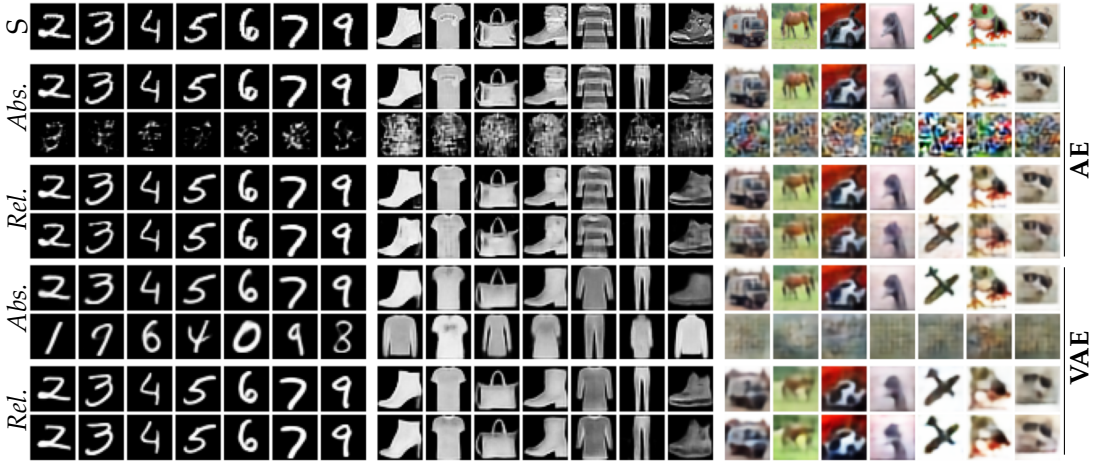
\begin{figure}[ht]
    \centering
    \begin{overpic}[trim=-0.27cm -0.15cm -.3cm 0cm,clip,width=1\linewidth]{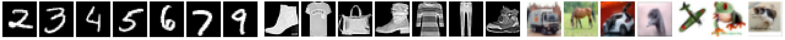}
        \put(0, 2.75){\rotatebox{90}{S}}
    \end{overpic}
    \begin{overpic}[trim=-0.27cm 0cm -.3cm 0cm,clip,width=1\linewidth]{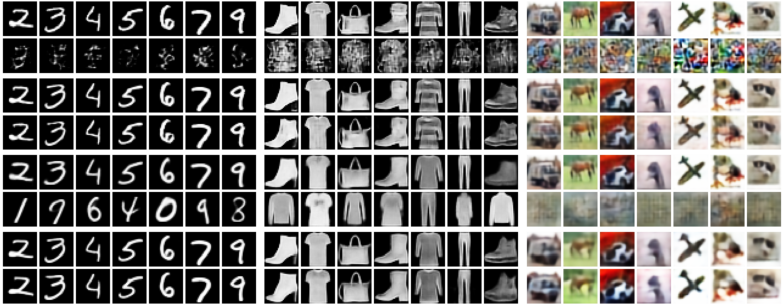}
        \put(0, 31){\rotatebox{90}{\small \textit{Abs.}}}
        \put(0, 21.75){\rotatebox{90}{\small \textit{Rel.}}}
        \put(0, 12.25){\rotatebox{90}{\small \textit{Abs.}}}
        \put(0, 3){\rotatebox{90}{\small \textit{Rel.}}}
        \put(98.5, 26.5){\rotatebox{90}{\small \textbf{AE}}}
        \put(98.5, 7.25){\rotatebox{90}{\small \textbf{VAE}}}
        \put(98, 19.25){\color{black}\line(0,1){17.75}}
        \put(98, .75){\color{black}\line(0,1){17.5}}
    \end{overpic}
    \caption[Zero-Shot Stitching Reconstruction examples]{Zero-Shot Stitching Reconstruction examples. Each column is a different image, row pairs are different architectures. In each pair, we first report the non-stitched reconstructions, then the stitched ones.}
    \label{relative:fig:encoder-decoder-swap}
\end{figure}

Hereafter, we showcase the Zero-Shot Stitching, as defined in \Cref{sec:stitchingdefinition}, capabilities of \gls{rr} across combinations of
different stochasticity sources (\Cref{relative:fig:encoder-decoder-swap,relative:tab:cross-train-performance-comparison}),
neural architectures (\Cref{relative:tab:cross-model-monolingual,relative:tab:multilingual-en-coarse-grained})
or datasets (\Cref{relative:tab:cifar100-imagenet}).
Finally, we present strong real-world applications in  NLP (\Cref{relative:sec:nlp-app}) and CV (\Cref{relative:sec:cv-app}), e.g., zero-shot predictions on novel languages.
Refer to \Cref{relative:sec:implementatin-details} for additional implementation details.

\subsection{Image Reconstruction}\label{relative:sec:model-reusability-ae}
\paragraph{Experimental setting.}
We perform Zero-Shot Stitching with \glsxtrshortpl{ae} and \glsxtrshortpl{vae} \emph{trained {on \glsxtrshortpl{rr}}} end-to-end on several datasets.
For each combination of model and dataset, we perform 5 trainings with different seeds, and zero-shot stitch together the resulting encoders and decoders.

\paragraph{Result analysis.}
In \Cref{relative:fig:encoder-decoder-swap}, the stitched models that employ absolute representations (\emph{Abs.}) produce erroneous predictions, since the latent spaces obtained from distinct trainings are incompatible.
Interestingly, although the absolute \gls{vae} does not produce compatible latent spaces, it is regularized.
As a result, the embeddings produced by the encoders correspond to incorrect but semantically meaningful reconstructions.
Instead, \gls{vae} based on \glspl{rr} (\emph{Rel.}) exhibit almost indistinguishable reconstructions between the models trained end-to-end and the stitched ones. Quantitative results are in \Cref{relative:tab:cross-train-performance-comparison}.

These results {support our claim that \glsxtrshortpl{rr} are empirically invariant} to the variation factors \gls{randomfactors}.

\begin{table}[ht]
    \caption[Zero-Shot Stitching performance]{Zero-Shot Stitching performance.
        The MSE ($\pm$ std) between the ground truth $\sX$ and the reconstructions is computed over 5 different seeds. Stitching with our \glsxtrshortpl{rr} yields an error up to two orders of magnitude less than the absolute counterpart.}
    \label{relative:tab:cross-train-performance-comparison}
    \centering
    \resizebox{\textwidth}{!}{%
        \begin{tabular}{cccrrrr|r}
            \toprule
                                                                           & \multicolumn{1}{c}{\bf }
                                                                           & \multicolumn{1}{c}{\bf }
                                                                           & \multicolumn{1}{c}{\bf \gls{mnist}}
                                                                           & \multicolumn{1}{c}{\bf \glsxtrshort{fmnist}}
                                                                           & \multicolumn{1}{c}{\bf \gls{cifart}}
                                                                           & \multicolumn{1}{c}{\bf \gls{cifarh}}
                                                                           & \multicolumn{1}{c}{\textbf{MSE ↓}}                                                                                         \\
            \midrule
            \multirow{4}{*}{\STAB{\rotatebox[origin=c]{90}{\textbf{AE}}}}  & \multirow{2}{*}{\STAB{\rotatebox[origin=c]{90}{\textit{Abs.}}}} & Non-Stitch.      &
            $0.66 \pm 0.02$                                                & $1.57 \pm 0.03$                                                 & $1.94 \pm 0.08$  & $2.13 \pm 0.08$   & $1.58 \pm 0.05$   \\
                                                                           &                                                                 & Stitch.          &
            $97.79 \pm 2.48$                                               & $120.54 \pm 6.81$                                               & $86.74 \pm 4.37$ & $97.17 \pm 3.50$  & $100.56 \pm 4.29$ \\[1ex]
                                                                           & \multirow{2}{*}{\STAB{\rotatebox[origin=c]{90}{\textit{Rel.}}}} & Non-Stitch.      &
            $1.18 \pm 0.02$                                                & $3.59 \pm 0.04$                                                 & $2.83 \pm 0.13$  & $3.50 \pm 0.08$   & $2.78 \pm 0.07$   \\
                                                                           &                                                                 & Stitch.          &
            $2.83 \pm 0.20$                                                & $6.37 \pm 0.29$                                                 & $5.39 \pm 1.18$  & $18.03 \pm 12.46$ & $8.16 \pm 3.53$   \\
            \midrule
            \multirow{4}{*}{\STAB{\rotatebox[origin=c]{90}{\textbf{VAE}}}} & \multirow{2}{*}{\STAB{\rotatebox[origin=c]{90}{\textit{Abs.}}}} & Non-Stitch.      &
            $1.31 \pm 0.04$                                                & $4.38 \pm 0.03$                                                 & $2.68 \pm 0.06$  & $3.00 \pm 0.03$   & $2.84 \pm 0.04$   \\
                                                                           &                                                                 & Stitch.          &
            $98.51 \pm 1.49$                                               & $118.96 \pm 2.96$                                               & $69.02 \pm 1.54$ & $78.57 \pm 1.88$  & $91.27 \pm 1.97$  \\[1ex]
                                                                           & \multirow{2}{*}{\STAB{\rotatebox[origin=c]{90}{\textit{Rel.}}}} & Non-Stitch.      &
            $2.97 \pm 0.14$                                                & $6.81 \pm 0.06$                                                 & $5.18 \pm 0.22$  & $5.93 \pm 0.14$   & $5.22 \pm 0.14$   \\
                                                                           &                                                                 & Stitch.          &
            $13.43 \pm 6.79$                                               & $24.03 \pm 13.15$                                               & $11.20 \pm 3.15$ & $11.23 \pm 2.38$  & $14.97 \pm 6.37$  \\
            \bottomrule
        \end{tabular}
    }
\end{table}

\subsection{Text Classification}\label{relative:sec:nlp-app}
In this Section, we show practical examples of the use of parallel anchors (\Cref{relative:sec:relative-representations}).

\paragraph{Experimental setting.}
We consider two different text classification settings.

\emph{Cross-lingual}: given a review, predict the associated star rating, done on multi-lingual data from the \gls{amazon} dataset.
Following the original paper, we work on a binarized version of the task, with FScore and \glsxtrshort{mae} as metrics. In \Cref{relative:tab:multilingual-full-fine-grained}, we report results on the fine-grained formulation.
We adopt four different pre-trained language-specific \glsxtrshort{robertab}  transformers and evaluate their Zero-Shot Stitching performance on languages never seen by the classifier. %
We use parallel anchors in two modalities: (i) \emph{Translated}: consider English reviews translated\footnote{We used the \texttt{=GOOGLETRANSLATE} function available in Google Sheets.} into the other languages; (ii) \emph{Wikipedia}: adopt an external corpus, WikiMatrix \citep{wikimatrix}, providing parallel sentences extracted from Wikipedia.

\emph{Cross-architecture}: assessed on three different datasets: \gls{trec} (coarse), \gls{dbpedia}, \gls{amazon} (English split). We adopt two different pre-trained BERT  transformers, \glsxtrshort{bertbc} and \glsxtrshort{bertbu}, \glsxtrshort{electrabd}  and \glsxtrshort{robertab}.

\begin{table}[ht]
    \scriptsize
    \centering
    \caption[Cross-lingual Zero-Shot Stitching performance comparison]{
        Cross-lingual Zero-Shot Stitching performance comparison. %
        The table reports the mean weighted F1 ($\pm$ std) and \glsxtrshort{mae} on \gls{amazon} coarse-grained, across 5 different seeds.}
    \label{relative:tab:multilingual-en-coarse-grained}
    \resizebox{\textwidth}{!}{%
        \begin{tabular}{clrrrrrr}
            \toprule
                                &                  & \multicolumn{2}{c}{Absolute} & \multicolumn{4}{c}{Relative}                                                                                                                                                       \\
            \cmidrule(lr){3-4}
            \cmidrule(l){5-8}
                                &                  & \multicolumn{2}{c}{}         & \multicolumn{2}{c}{Translated}        & \multicolumn{2}{c}{Wikipedia}                                                                                                              \\
            \cmidrule(lr){5-6}
            \cmidrule(l){7-8}
            \textbf{Decoder}    & \textbf{Encoder} & \multicolumn{1}{c}{FScore}   & \multicolumn{1}{c}{\glsxtrshort{mae}} & \multicolumn{1}{c}{FScore}    & \multicolumn{1}{c}{\glsxtrshort{mae}} & \multicolumn{1}{c}{FScore} & \multicolumn{1}{c}{\glsxtrshort{mae}} \\
            \midrule
            \multirow{4}{*}{en} & en               & $91.54 \pm 0.58$             & $0.08 \pm 0.01$                       & $90.06 \pm 0.60$              & $0.10 \pm 0.01$                       & $90.45 \pm 0.52$           & $0.10 \pm 0.01$                       \\
                                & es               & $43.67 \pm 1.09$             & $0.56 \pm 0.01$                       & $82.78 \pm 0.81$              & $0.17 \pm 0.01$                       & $78.53 \pm 0.30$           & $0.21 \pm 0.00$                       \\
                                & fr               & $54.41 \pm 1.61$             & $0.45 \pm 0.02$                       & $78.49 \pm 0.66$              & $0.21 \pm 0.01$                       & $70.41 \pm 0.57$           & $0.29 \pm 0.01$                       \\
                                & ja               & $48.72 \pm 0.90$             & $0.51 \pm 0.01$                       & $65.72 \pm 0.55$              & $0.34 \pm 0.01$                       & $66.31 \pm 0.80$           & $0.34 \pm 0.01$                       \\
            \bottomrule
        \end{tabular}
    }
\end{table}

\begin{table}[ht]
    \caption[Cross-architecture Zero-Shot Stitching performance comparison]{Cross-architecture Zero-Shot Stitching performance comparison. The table reports the mean weighted F1 ($\pm$ std) for each dataset, across 5 different seeds.}
    \label{relative:tab:cross-model-monolingual}
    \centering
    \footnotesize
    \begin{tabular}{llrrrr}
        \toprule
                                                                 &            & \multicolumn{1}{c}{\gls{trec}} & \multicolumn{1}{c}{\gls{dbpedia}} & \multicolumn{2}{c}{\gls{amazon}}                                        \\ \cmidrule{5-6}
                                                                 &            &                                &                                   & \multicolumn{1}{c}{\textit{Coarse}} & \multicolumn{1}{c}{\textit{Fine}} \\
        \midrule
        \multirow{2}{*}{{\STAB{\rotatebox[origin=c]{90}{Abs.}}}} & Non-Stitch & $91.70 \pm 1.39$               & $98.62 \pm 0.58$                  & $87.81 \pm 1.58$                    & $55.35 \pm 3.19$                  \\
                                                                 & Stitch     & $21.49 \pm 3.64$               & $6.96 \pm 1.46$                   & $49.58 \pm 2.95$                    & $19.01 \pm 2.04$                  \\
        \midrule
        \multirow{2}{*}{{\STAB{\rotatebox[origin=c]{90}{Rel.}}}} & Non-Stitch & $88.08 \pm 1.37$               & $97.42 \pm 2.05$                  & $85.08 \pm 1.93$                    & $48.92 \pm 3.57$                  \\
                                                                 & Stitch     & $75.89 \pm 5.38$               & $80.47 \pm 21.14$                 & $72.37 \pm 7.32$                    & $33.24 \pm 7.21$                  \\
        \bottomrule
    \end{tabular}
\end{table}

\paragraph{Result analysis.}
\Cref{relative:tab:multilingual-en-coarse-grained,relative:tab:cross-model-monolingual} show for the first time that it is possible to learn to solve a downstream task on a specific language or transformer and perform predictions on another.

Stitching with absolute representations yields performances comparable to random guessing across the board, proving that \glsxtrshortpl{rr} are a key element for the success of this kind of Zero-Shot Stitching.
Moreover, \Cref{relative:tab:multilingual-en-coarse-grained} highlights the robustness that \glsxtrshortpl{rr} have on the choice of anchors, even when they are noisy (\emph{Translated} case), or their distribution differs from one of the downstream task (\emph{Wikipedia} case), as long as their encoding can be handled correctly by the encoder. In our case, the encoder is pre-trained to represent a variety of texts in a specific language, thus, even if WikiMatrix has a completely different domain from \gls{amazon}, the transformer still computes a meaningful representation, comparable  with those of the reviews.
We report in \Cref{relative:tab:multilingual-full-coarse-grained,relative:tab:multilingual-full-fine-grained} complete results on all languages combination, and in \Cref{relative:tab:xmlr-multilingual-full-fine-grained} the performance obtained by a multi-lingual transformer; %
that, to the best of our knowledge, is the only alternative {for obtaining} compatible representations across languages.

According to these results, \glsxtrshortpl{rr} show invariance to different architectures and data distribution shifts (e.g., different train languages).

\subsection{Image Classification}\label{relative:sec:cv-app}
In this Section, we show practical examples of the use of OOD anchors (\Cref{relative:sec:relative-representations}).

\paragraph{Experimental setting.}
We consider a classification task on the datasets \gls{imagenet1k} and \gls{cifarh} with coarse labels (20), and 4 different pre-trained image encoders: %
three variants of the ViT transformer (\glsxtrshort{vitsp16224}, \glsxtrshort{vitbp16224} and \glsxtrshort{vitbr50384}) and \glsxtrshort{rexnet100}. %

\begin{table}[ht]
    \scriptsize
    \centering
    \caption[Zero-Shot Stitching performance with different encoding techniques]{
        Zero-Shot Stitching performance comparison with different encoding techniques. The table reports the mean weighted F1 ($\pm$ std) on \gls{cifarh} coarse-grained and \gls{imagenet1k}, across 5 seeds.}
    \label{relative:tab:cifar100-imagenet}
    \begin{tabular}{llrrrr}
        \toprule
                                                    &                            & \multicolumn{2}{c}{\gls{cifarh}} & \multicolumn{2}{c}{\gls{imagenet1k}}                                                               \\\cmidrule(lr){3-4}\cmidrule(lr){5-6}
        \textbf{Decoder}                            & \textbf{Encoder}           & \multicolumn{1}{c}{Absolute}     & \multicolumn{1}{c}{Relative}         & \multicolumn{1}{c}{Absolute} & \multicolumn{1}{c}{Relative} \\
        \midrule
        \multirow{4}{*}{{\glsxtrshort{rexnet100}}}  & {\glsxtrshort{rexnet100}}  & $82.06 \pm 0.15$                 & $80.22 \pm 0.28$                     & $73.78 \pm 0.29$             & $72.61 \pm 0.16$             \\
                                                    & {\glsxtrshort{vitbp16224}} & \multicolumn{1}{c}{-}            & $54.98 \pm 0.44$                     & \multicolumn{1}{c}{-}        & $37.39 \pm 0.36$             \\
                                                    & {\glsxtrshort{vitbr50384}} & \multicolumn{1}{c}{-}            & $53.33 \pm 0.37$                     & \multicolumn{1}{c}{-}        & $42.36 \pm 0.36$             \\
                                                    & {\glsxtrshort{vitsp16224}} & \multicolumn{1}{c}{-}            & $59.82 \pm 0.32$                     & \multicolumn{1}{c}{-}        & $43.75 \pm 0.27$             \\
        \cmidrule{1-6}
        \multirow{4}{*}{{\glsxtrshort{vitbp16224}}} & {\glsxtrshort{rexnet100}}  & \multicolumn{1}{c}{-}            & $76.81 \pm 0.49$                     & \multicolumn{1}{c}{-}        & $30.78 \pm 0.81$             \\
                                                    & {\glsxtrshort{vitbp16224}} & $93.15 \pm 0.05$                 & $91.94 \pm 0.10$                     & $80.91 \pm 0.29$             & $78.86 \pm 0.33$             \\
                                                    & {\glsxtrshort{vitbr50384}} & $6.21 \pm 0.33$                  & $81.42 \pm 0.38$                     & $0.07 \pm 0.05$              & $44.72 \pm 0.57$             \\
                                                    & {\glsxtrshort{vitsp16224}} & \multicolumn{1}{c}{-}            & $84.29 \pm 0.86$                     & \multicolumn{1}{c}{-}        & $48.31 \pm 0.72$             \\
        \cmidrule{1-6}
        \multirow{4}{*}{{\glsxtrshort{vitbr50384}}} & {\glsxtrshort{rexnet100}}  & \multicolumn{1}{c}{-}            & $79.79 \pm 0.43$                     & \multicolumn{1}{c}{-}        & $53.46 \pm 0.68$             \\
                                                    & {\glsxtrshort{vitbp16224}} & $4.69 \pm 0.07$                  & $84.46 \pm 0.19$                     & $0.08 \pm 0.04$              & $62.21 \pm 0.54$             \\
                                                    & {\glsxtrshort{vitbr50384}} & $91.41 \pm 0.09$                 & $90.77 \pm 0.16$                     & $82.55 \pm 0.30$             & $81.88 \pm 0.16$             \\
                                                    & {\glsxtrshort{vitsp16224}} & \multicolumn{1}{c}{-}            & $84.66 \pm 0.16$                     & \multicolumn{1}{c}{-}        & $61.32 \pm 0.36$             \\
        \cmidrule{1-6}
        \multirow{4}{*}{{\glsxtrshort{vitsp16224}}} & {\glsxtrshort{rexnet100}}  & \multicolumn{1}{c}{-}            & $75.35 \pm 0.41$                     & \multicolumn{1}{c}{-}        & $37.58 \pm 0.44$             \\
                                                    & {\glsxtrshort{vitbp16224}} & \multicolumn{1}{c}{-}            & $81.23 \pm 0.31$                     & \multicolumn{1}{c}{-}        & $50.08 \pm 0.63$             \\
                                                    & {\glsxtrshort{vitbr50384}} & \multicolumn{1}{c}{-}            & $78.35 \pm 0.69$                     & \multicolumn{1}{c}{-}        & $45.45 \pm 1.41$             \\
                                                    & {\glsxtrshort{vitsp16224}} & $90.07 \pm 0.19$                 & $88.85 \pm 0.44$                     & $77.73 \pm 0.41$             & $76.36 \pm 0.40$             \\
        \bottomrule
    \end{tabular}
\end{table}

\paragraph{Result analysis.}
The results in \Cref{relative:tab:cifar100-imagenet} highlight how the \glsxtrshortpl{rr} allow {stitching} modules with different encoding dimensionality, since {the decoder receives a \gls{rr}} with guaranteed equal size equal to the number of anchors. Furthermore, the results demonstrate the ability to generalize and perform Zero-Shot Stitching on \gls{cifarh}, although that data was never seen by the encoder since it is a frozen transformer trained on \gls{imagenet1k}.
Interestingly, \texttt{\glsxtrshort{rexnet100}} is the only transformer whose latent dimensionality is higher than the number of anchors, and the biggest drop in stitching performance happens when the decoder is trained on it. This suggests the number of anchors is an important hyperparameter; we refer to \Cref{relative:fig:anchor-num} for a more in-depth analysis.

Overall, these results prove that \glsxtrshortpl{rr} can bridge general purpose encoders and pre-trained task-specific decoders.

\Chapter{Direct Translation}{Latent Space Translation via Semantic Alignment\footnote{\fullcite{maiorca2023}}}
\label{chap:translation}

\begin{quotation}
    \noindent
    In this Chapter, we address the \gls{lcp} as defined in \Cref{chap:problemformalization}, incorporating an additional assumption: either
    \gls{Tx} or \gls{Ty} is the identity function. Referring to \Cref{fig:formalization}, this means that we assume the transformation $\gls{transformation} \in \gls{transformationclass}$, that maps the latent manifold embeddings from one space to another, can be directly approximated by some $\gls{transformationapprox}$.
    In the following, we show that good approximations $\gls{transformationapprox} \approx \gls{transformation}$
    are simpler than previously thought, i.e., at most affine transformations, and can often be estimated using standard, well-understood algebraic procedures with closed-form solutions.
\end{quotation}

\section{Introduction}

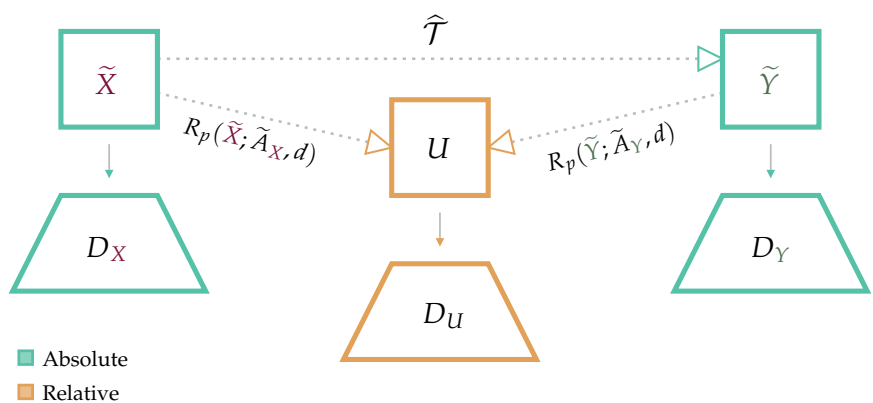
\begin{figure}[ht]
    \centering
    \begin{overpic}[trim=0 0cm 0 0cm,clip,width=0.8\linewidth]{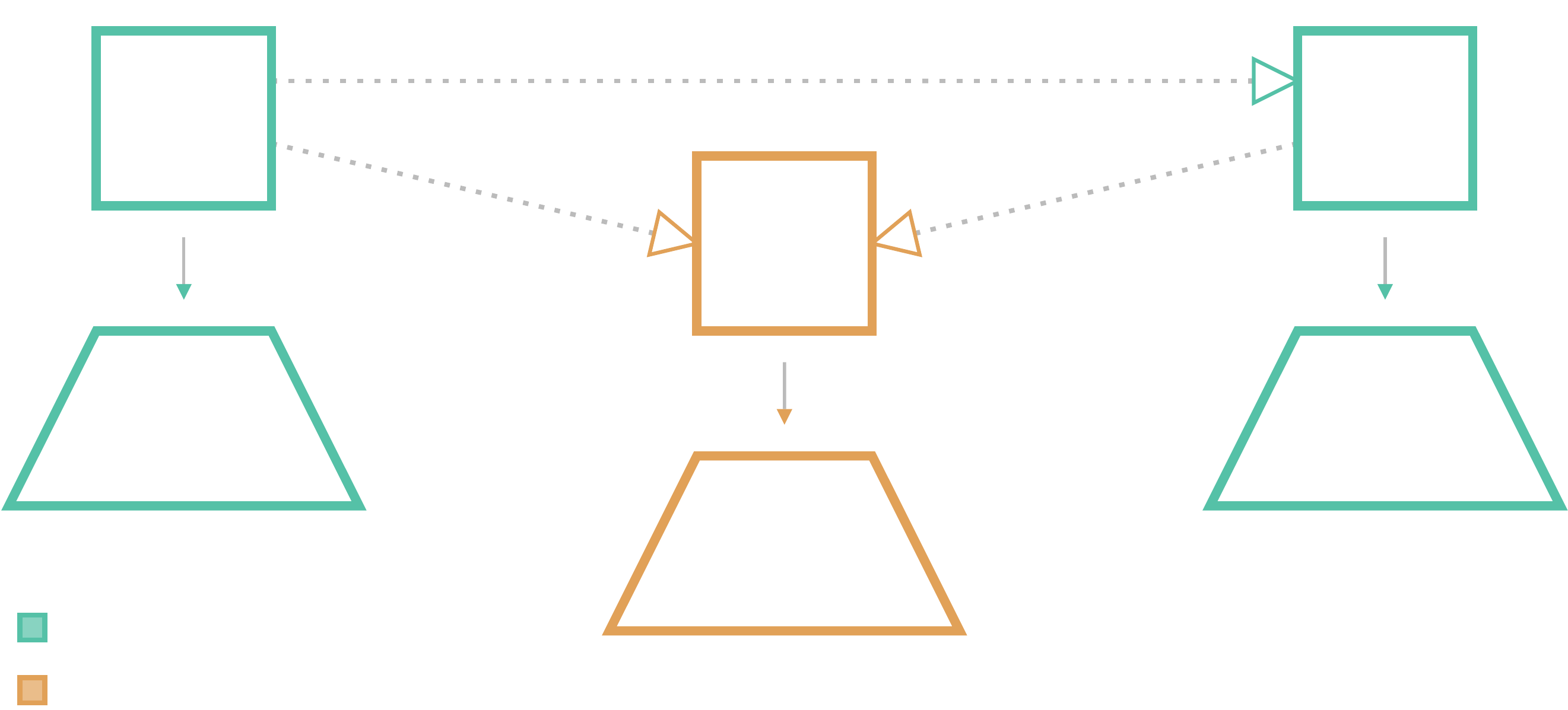}
        \put(10, 36){\gls{ZX}}
        \put(87, 36){\gls{ZY}}
        \put(48.5, 28.5){\gls{U}}
        \put(9, 17){$\gls{dec}_{\gls{X}}$}
        \put(86, 17){$\gls{dec}_{\gls{Y}}$}
        \put(48, 9.25){$\gls{dec}_{\gls{U}}$}
        \put(48, 42){\gls{transformationapprox}}
        \put(20, 31.5){\rotatebox{-13}{\footnotesize$\gls{relativeprojection}(\gls{ZX}; \gls{ZA}_{\gls{X}}, \gls{sim})$}}
        \put(62, 27){\rotatebox{15}{\footnotesize$\gls{relativeprojection}(\gls{ZY}; \gls{ZA}_{\gls{Y}}, \gls{sim})$}}
        \put(4, 4.25){\scriptsize Absolute}
        \put(4, 0.25){\scriptsize Relative}
    \end{overpic}
    \caption[Zero-shot stitching between absolute spaces]{
    Zero-shot stitching between absolute spaces utilizing \glsxtrshortpl{rr} and the method presented in this Chapter (the estimation of ${\gls{transformationapprox}} \approx \gls{transformation}$).
    The proposed approach does not require a decoder $\gls{dec}_{\gls{U}}$ specifically trained on \glsxtrshortpl{rr}. Instead, we directly translate latent spaces, enabling the use of arbitrarily pre-trained decoders originally trained on absolute spaces, i.e., $\gls{dec}_{\gls{X}}$~and~$\gls{dec}_{\gls{Y}}$.
    }
    \label{translation:fig:teaser}
\end{figure}

One of the key findings from \Cref{chap:relative} is the empirical evidence demonstrating that the signal encoded in the angle norms, with respect to a reduced set of data points \gls{anchors}, suffices to represent the latent manifold  $\gls{phiZX}(\gls{ZMx})$ embedded within the latent space. This representation is accurate enough to allow downstream performance comparable to using absolute embeddings, in the specific tasks considered.

Building on this intuition of the existence of a relatively simple transformation relating the latent manifolds, we show the effectiveness and applications of \emph{directly translating between different latent spaces}.
Specifically, we show that it is feasible to directly approximate a transformation \gls{transformation} with some \gls{transformationapprox}, given that a {partial (and possibly sparse) correspondence} between data points $\gls{parallelanchors} \subseteq \gls{pi}$ is established.
Unexpectedly, the process of seamlessly combining different \glsxtrshortpl{nn} -- each pre-trained on different datasets, modalities, architectures, or domains -- turns out to be surprisingly straightforward.

For instance, we show how it enables the ability to effectively integrate any pre-trained text encoder with any image classification head, and vice versa; without requiring any additional re-training or assumptions, e.g., without assuming the decoders are trained on \glsxtrshortpl{rr} as in \Cref{chap:relative}.
The method difference is emphasized in \Cref{translation:fig:teaser}, Zero-Shot Stitching (\Cref{sec:stitchingdefinition}) with \glsxtrshortpl{rr} assumes the use of a single decoder specifically trained on a relative space; meanwhile, the method presented in this Chapter allows to zero-shot stitch and reuse decoders originally trained on the absolute spaces.

Our main contributions can be summarized as follows:
\begin{itemize}
    \item We explore the direct translation between latent spaces of distinct \glsxtrshortpl{nn} to solve the \gls{lcp}, as defined in \Cref{chap:problemformalization} and illustrated in \Cref{fig:formalization}. In particular, leveraging a semantic correspondence between the input spaces $\gls{parallelanchors} \subseteq \gls{pi}$, we directly approximate \gls{transformation} for the first time across different trainings, architectures, and modalities. We obtain excellent stitching performances even in cross-modal settings, where we \emph{apply arbitrary text classifiers on top of pre-trained image encodings} (and vice versa).
    \item We show that different downstream tasks, namely classification and generation, require modeling different transformations to obtain the most out of the translation between their latent spaces.
\end{itemize}

\section{Latent Space Translation}
\label{translation:sec:translation}

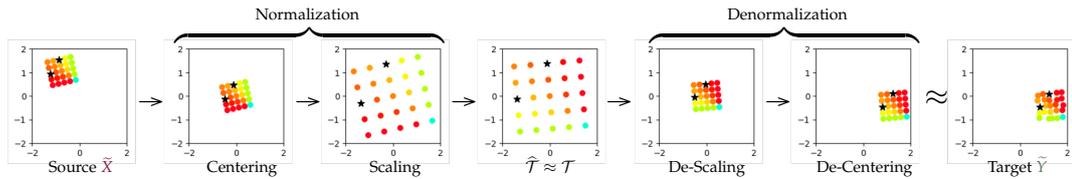
\begin{figure}
    \centering
    \begin{overpic}[trim=0 0cm 0 -1cm,clip,width=1\linewidth,tics=4,]{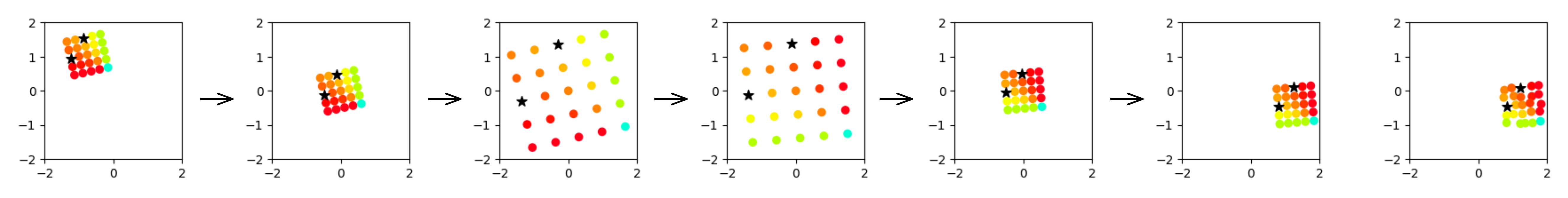}
        \put(4,-.25){\tiny{ Source ${\gls{ZX}}$}}
        \put(91,-.25){\tiny{ Target ${\gls{ZY}}$}}
        \put(16,14){\rotatebox{180}{\makebox(25,3){\upbracefill}}}
        \put(23.5,14){\tiny Normalization}

        \put(60,14){\rotatebox{180}{\makebox(25,3){\upbracefill}}}
        \put(67.25,14){\tiny Denormalization}

        \put(19,-.25){\tiny Centering}
        \put(34.1,-.25){\tiny Scaling}
        \put(48.5,-.25){\tiny $\gls{transformationapprox} \approx {\gls{transformation}}$}
        \put(61.8,-.25){\tiny De-Scaling}
        \put(75.5,-.25){\tiny De-Centering}

        \put(85.5,6){$\approx$}
    \end{overpic}
    \caption[Direct translation illustration on a synthetic example.]{Method illustration on a synthetic example. Given a source space ${\gls{ZX}}$, the steps to translate it to a target ${\gls{ZY}}$ are sequentially applied as described in \Cref{translation:sec:translation}. Note that the translation is not perfect due to an arbitrary distortion of the data.}
    \label{translation:fig:method}
\end{figure}

\subsection{Assumptions}
In this Chapter, we address the \gls{lcp} described in \Cref{chap:problemformalization} and \Cref{fig:formalization}, with the additional assumption that either \gls{Tx} or \gls{Ty} is the identity. This means that we are directly trying to approximate $\gls{transformationapprox} \approx \gls{transformation} \in \gls{transformationclass}$.
Without loss of generality, we always assume that  \gls{Ty} is the identity, thus $\gls{transformationapprox} \approx \gls{Tx} = \gls{transformation}$.
Furthermore, we assume that \gls{transformationapprox} is at most an affine transformation. Please refer to \Cref{chap:problemformalization} and \Cref{fig:formalization} for a formal definition of the \gls{lcp}.

\subsection{Method}
Consider two latent spaces, ${\gls{ZX}} \in {\gls{R}}^{n \times d_1}$ and ${\gls{ZY}} \in {\gls{R}}^{n \times d_2}$.
Our objective is to estimate the transformation $\gls{transformationapprox} \approx {\gls{transformation}} \in \gls{transformationclass}$ that translates $\gls{phiZX}(\gls{ZMx}) \subseteq \gls{ZX}$ into $\gls{phiZY}(\gls{ZMy}) \subseteq \gls{ZY}$, i.e.:$\;$ ${\gls{phiZY}(\gls{ZMy})} = \gls{transformationapprox}\gls{phiZX}(\gls{ZMx})$, exploiting the \emph{semantic alignment} \gls{C} observed through $\gls{parallelanchors} \subseteq \gls{pi}$ between the input spaces \gls{X} and \gls{Y}.

Throughout this work, we identify two main steps in the translation process: pre-processing the spaces and estimating the transformation ${\gls{transformationapprox}}$, as outlined in \Cref{translation:fig:method}.

\paragraph{Pre-processing.}
Generally, the two spaces \gls{ZX} and \gls{ZY} may have different dimensionalities -- in those cases, we zero-pad the smaller one to match the dimension of the other without changing its underlying structure \citep{Williams_etal_2021_shapes}. Moreover, we standardize each feature to have zero mean and unit variance (standard scaling) if not otherwise specified, whose statistics are computed only on the anchor sets for both source and target space, to perform the necessary denormalization.

\paragraph{Estimating ${\gls{transformationapprox}} \approx \gls{transformation}$.}
\label{translation:sec:estimatingt}
In \Cref{chap:relative}, it is empirically shown that the spaces often differ by an angle-norm preserving transformation. Nevertheless, we broaden our investigation by considering different ways of obtaining ${\gls{transformationapprox}}$ to evaluate the robustness of that assumption. Throughout our experiments, we primarily operate under the assumption that ${\gls{transformationapprox}}$ can be constrained to encode, at most, an affine transformation:~${\gls{transformationapprox}}({\gls{ZX}}) = \mathbf{R} {\gls{ZX}} + \mathbf{b}$.

This general formulation, without additional constraints, corresponds to our \texttt{affine} method in the experiments, and it is optimized via gradient descent. The other transformations are trivially obtained by progressively adding constraints on this one:
\begin{itemize}
    \item \texttt{linear}. To model a linear transformation, we can just set the bias term to zero $\mathbf{b} = \vec{0}$ and optimize via Least Square. Here, we are both simplifying the class of transformations and switching from a gradient descent optimization to a closed-form procedure.
    \item \texttt{l-ortho}. Additionally, we could require $\mathbf{R}$ to be orthogonal to encode an isometry. In this case, we obtain this by applying Singular Value Decomposition (SVD) on the corresponding $\mathbf{R}$ obtained by the \texttt{linear} solution. Through this, we aim to understand the implications of enforcing orthogonality on a transformation that was originally not constrained to be so, in a setting similar to \cite{xing-etal-2015-normalized}.
    \item \texttt{ortho}. To obtain the optimal orthogonal $\mathbf{R}$, we apply Procrustes analysis \citep{gower1975a}. Please refer to \Cref{related:align} for further details.
\end{itemize}

The transformation ${\gls{transformationapprox}}$ is estimated from samples in semantic correspondence $\gls{parallelanchors} \subseteq \gls{pi}$, i.e., the parallel anchors defined in \Cref{sec:latentcommunicationproblem}.

This methodology facilitates efficient and precise zero-shot translation between disparate latent spaces, providing a robust and versatile foundation for model reuse and interoperability in diverse machine learning contexts.

\begin{figure}
    \centering
    \begin{overpic}[trim=-0cm -1cm 0 -0cm, clip,width=0.95\linewidth]{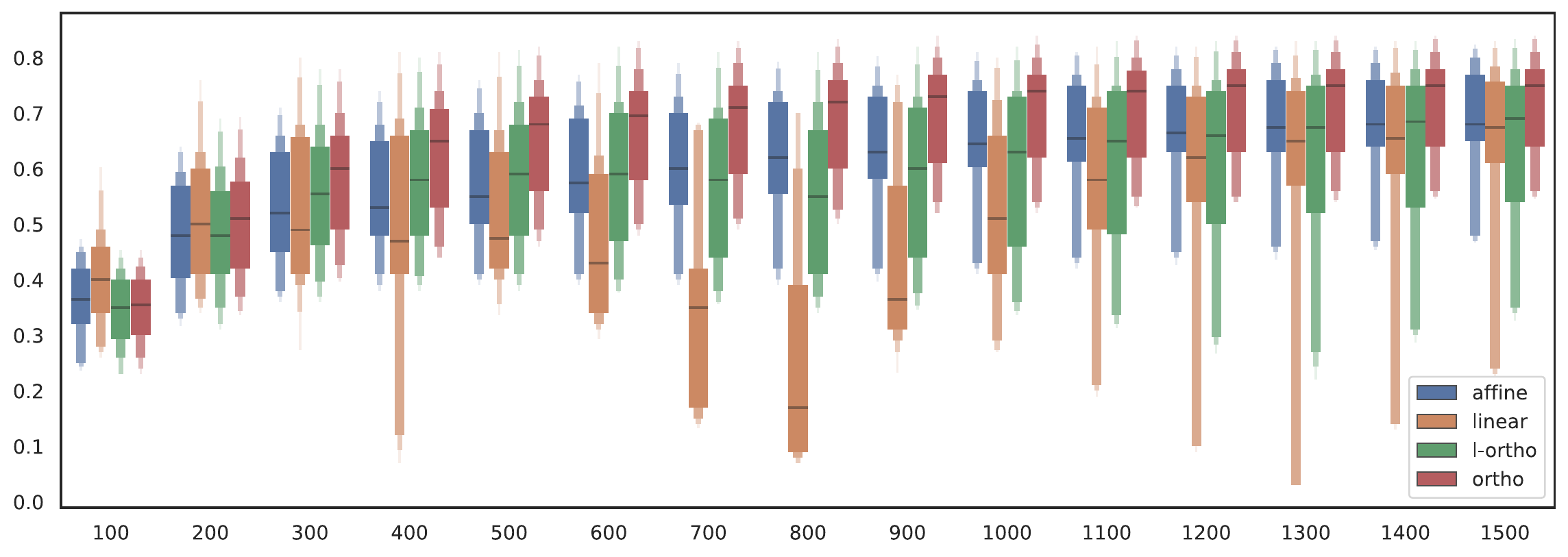}
        \put(-2.5, 15){\rotatebox{90}{Accuracy}}
        \put(42, -.5){Number of anchors}
    \end{overpic}
    \caption[Performance comparison of \texttt{affine}, \texttt{linear}, \texttt{l-ortho}, and \texttt{ortho}]{Performance comparison of \texttt{affine}, \texttt{linear}, \texttt{l-ortho}, and \texttt{ortho} at varying number of anchors on classification accuracy. Results on \gls{cifarh} fine-grained. The same analysis for the generation case is in \Cref{translation:sup:translation:fig:anchors-num} in the Appendix.}
    \label{translation:fig:anchors-num}
\end{figure}

\section{Latent Communication via Translation}\label{translation:sec:latent-translation}

In this Section, we evaluate the capabilities and effectiveness of our translation method through various scenarios, highlighting its applicability in diverse contexts. We present empirical results in three different settings: (i) cross-architecture; (ii) cross-modality; (iii) autoencoding.
In each case, the translation performance of each method for obtaining the transformation ${\gls{transformationapprox}}$ is evaluated against two baselines, the naive absolute one and the relative one.

\paragraph{Stitching Procedure.}
\label{translation:sec:exp_baseline}
In line with the \textit{Zero-Shot Stitching} concept we introduced in \Cref{sec:stitchingdefinition}, we combine independent encoders and decoders (e.g., classifiers, generators)  without further training or fine-tuning. This study does not necessitate a decoder trained on relative representations; instead, we directly employ the original decoders trained on absolute spaces.
Each one of the benchmarks we conduct follows the same procedure unless otherwise specified: we measure the mean performance over all the possible combinations of (encoder, decoder) for each test set in different~settings:
\begin{itemize}
    \item \textit{no-stitch}. The end-to-end performance of the decoder applied to the original space it was trained on. This is useful to establish un upper-bound in performances.
    \item \textit{absolute}. The result of using the encodings without any transformation, we consider this as a probe for any pre-existing compatibility among encodings and, therefore, a lower-bound.
    \item \textit{translation}. These are the results of the application of our latent translation method, with the estimation of ${\gls{transformationapprox}}$ via \texttt{affine}, \texttt{linear}, \texttt{l-ortho} and \texttt{ortho}.
\end{itemize}

In each instance, we use the same parallel anchors \gls{parallelanchors}, that are uniformly chosen, in a quantity comparable with the dimensionality of the absolute representation.

\subsection{Cross-Architecture}
\label{translation:sec:stitch-cross-architecture}
Firstly, we test our method in a cross-architecture setting, Zero-Shot Stitching together encodings coming from a variety of pre-trained networks and their associated absolute decoders (classifiers). This scenario provides an extensive testing ground for our method and demonstrates its robustness across different architectures. Please refer to \Cref{translation:sup:cross-architecture:generation} in the Appendix for further results on cross-architecture stitching in generation tasks.

\begin{table}[ht]
    \centering
    \caption[Cross-architecture stitching with various ${\gls{transformationapprox}}$ and standard scaling]{Cross-architecture stitching with various methods for estimating ${\gls{transformationapprox}}$ and applying standard scaling. The stitched decoders are \glsxtrshortpl{svm} with a linear kernel. 5 runs for each encoder-decoder pair. (C) and (F) next to \gls{cifarh} indicate, respectively, coarse-grained and fine-grained. Please refer to the Appendix in \Cref{translation:sup:stitching:mlp:std} for additional results with \glsxtrshortpl{mlp} as classification heads.}
    \label{translation:tab:stitching:svm}
    \resizebox{\textwidth}{!}{%
        \begin{tabular}{lllllllll}
            \toprule
                                                                    & \multicolumn{1}{l}{\texttt{Dataset}} & \multicolumn{1}{c}{\texttt{no-stitch}} & \multicolumn{1}{c}{\texttt{absolute}} & \multicolumn{1}{c}{\texttt{relative}} & \multicolumn{1}{c}{\texttt{affine}} & \multicolumn{1}{c}{\texttt{linear}} & \multicolumn{1}{c}{\texttt{l-ortho}} & \multicolumn{1}{c}{\texttt{ortho}} \\
            \midrule
            \multirow{5}{*}{\rotatebox{90}{\textbf{Vision}}}        & \gls{cifart}                         & $0.95 \pm 0.03$                        & $0.16 \pm 0.22$                       & $0.80 \pm 0.22$                       & $0.92 \pm 0.05$                     & $0.88 \pm 0.11$                     & $0.90 \pm 0.09$                      & $0.93 \pm 0.04$                    \\
                                                                    & \gls{cifarh}-\texttt{C}              & $0.85 \pm 0.07$                        & $0.11 \pm 0.21$                       & $0.54 \pm 0.25$                       & $0.78 \pm 0.09$                     & $0.73 \pm 0.16$                     & $0.77 \pm 0.11$                      & $0.81 \pm 0.07$                    \\
                                                                    & \gls{cifarh}-\texttt{F}              & $0.76 \pm 0.09$                        & $0.07 \pm 0.21$                       & $0.30 \pm 0.24$                       & $0.68 \pm 0.11$                     & $0.62 \pm 0.19$                     & $0.64 \pm 0.16$                      & $0.71 \pm 0.09$                    \\
                                                                    & \glsxtrshort{fmnist}                 & $0.88 \pm 0.01$                        & $0.15 \pm 0.20$                       & $0.63 \pm 0.23$                       & $0.86 \pm 0.01$                     & $0.83 \pm 0.06$                     & $0.82 \pm 0.05$                      & $0.85 \pm 0.02$                    \\
                                                                    & \gls{mnist}                          & $0.96 \pm 0.01$                        & $0.15 \pm 0.21$                       & $0.50 \pm 0.22$                       & $0.94 \pm 0.01$                     & $0.89 \pm 0.08$                     & $0.81 \pm 0.11$                      & $0.91 \pm 0.02$                    \\
            \midrule \multirow{4}{*}{\rotatebox{90}{\textbf{Text}}} & \gls{trec}                           & $0.87 \pm 0.12$                        & $0.20 \pm 0.06$                       & $0.36 \pm 0.13$                       & $0.82 \pm 0.12$                     & $0.74 \pm 0.25$                     & $0.57 \pm 0.25$                      & $0.79 \pm 0.11$                    \\
                                                                    & \gls{agnews}                         & $0.73 \pm 0.09$                        & $0.25 \pm 0.02$                       & $0.39 \pm 0.13$                       & $0.65 \pm 0.08$                     & $0.62 \pm 0.08$                     & $0.61 \pm 0.10$                      & $0.66 \pm 0.10$                    \\
                                                                    & \gls{dbpedia}                        & $0.78 \pm 0.23$                        & $0.07 \pm 0.01$                       & $0.16 \pm 0.10$                       & $0.66 \pm 0.24$                     & $0.62 \pm 0.23$                     & $0.57 \pm 0.23$                      & $0.66 \pm 0.22$                    \\
                                                                    & \gls{imdb}                           & $0.61 \pm 0.04$                        & $0.50 \pm 0.01$                       & $0.51 \pm 0.02$                       & $0.59 \pm 0.04$                     & $0.57 \pm 0.04$                     & $0.56 \pm 0.03$                      & $0.59 \pm 0.04$                    \\
            \bottomrule
        \end{tabular}
    }
\end{table}

\paragraph{Experimental setting.} We consider a variety of \glsxtrlong{cv} (\gls{mnist}, \gls{fmnist}, \gls{n24news}, \gls{cifart}, \gls{cifarh}) and \glsxtrlong{nlp} (\gls{trec}, \gls{dbpedia}, \gls{n24news}, \gls{agnews}, \gls{imdb}) datasets. For the text domain we consider 6 different language models as encoders (\glsxtrshort{bertbc}, \glsxtrshort{bertbu}, \glsxtrshort{electrabd}, \glsxtrshort{robertab}, \glsxtrshort{albertbv2}, \glsxtrshort{xlmrobertab}, and the text encoder of \glsxtrshort{clip}), and for the image domain 6 encoders (\glsxtrshort{rexnet100}, \glsxtrshort{vitsp16224}, \glsxtrshort{vitbp16224},     \glsxtrshort{vitbp16384}, \glsxtrshort{vitbr50384}, and the image encoder of \glsxtrshort{clip}), all pre-trained and frozen. The full encoder list can be found in \Cref{translation:sup:hf:models} in the Appendix.
For each dataset and for each encoder, we train an \glsxtrshort{svm} classification head (decoder) on top of their specific encodings.
We then proceed with the standard stitching procedure outlined in \Cref{translation:sec:exp_baseline} and collect the results. Please see \Cref{translation:sup:cross-architecture:generation} in the Appendix for cross-architecture stitching in generation tasks, where we extend this analysis by verifying that our method works even across autoencoders of different bottleneck sizes.

\begin{table}[ht]
    \centering
    \caption[Cross-architecture stitching with various ${\gls{transformationapprox}}$ and l2 normalization]{Cross-architecture stitching with various methods for estimating ${\gls{transformationapprox}}$ and applying L2 normalization. The stitched decoders are \glsxtrshortpl{svm} with linear kernel. 5 runs for each encoder-decoder pair. (C) and (F) next to \gls{cifarh} indicate, respectively, coarse-grained and fine-grained. Please refer to \Cref{translation:sup:stitching:mlp:l2} in the Appendix for additional results with \glsxtrshortpl{mlp} as classification heads. }
    \label{translation:tab:stitching:svm:l2}
    \resizebox{\textwidth}{!}{%
        \begin{tabular}{lllllllll}
            \toprule
                                                                    & \multicolumn{1}{l}{\texttt{Dataset}} & \multicolumn{1}{c}{\texttt{no-stitch}} & \multicolumn{1}{c}{\texttt{absolute}} & \multicolumn{1}{c}{\texttt{relative}} & \multicolumn{1}{c}{\texttt{affine}} & \multicolumn{1}{c}{\texttt{linear}} & \multicolumn{1}{c}{\texttt{l-ortho}} & \multicolumn{1}{c}{\texttt{ortho}} \\
            \midrule
            \multirow{5}{*}{\rotatebox{90}{\textbf{Vision}}}        & \gls{cifart}                         & $0.95 \pm 0.03$                        & $0.16 \pm 0.22$                       & $0.80 \pm 0.22$                       & $0.93 \pm 0.04$                     & $0.78 \pm 0.27$                     & $0.88 \pm 0.12$                      & $0.91 \pm 0.09$                    \\
                                                                    & \gls{cifarh}-\texttt{C}              & $0.85 \pm 0.07$                        & $0.11 \pm 0.21$                       & $0.54 \pm 0.25$                       & $0.79 \pm 0.07$                     & $0.65 \pm 0.25$                     & $0.73 \pm 0.17$                      & $0.79 \pm 0.10$                    \\
                                                                    & \gls{cifarh}-\texttt{F}              & $0.76 \pm 0.09$                        & $0.07 \pm 0.21$                       & $0.30 \pm 0.24$                       & $0.69 \pm 0.10$                     & $0.52 \pm 0.25$                     & $0.62 \pm 0.19$                      & $0.68 \pm 0.13$                    \\
                                                                    & \glsxtrshort{fmnist}                 & $0.88 \pm 0.01$                        & $0.15 \pm 0.20$                       & $0.63 \pm 0.23$                       & $0.86 \pm 0.01$                     & $0.65 \pm 0.23$                     & $0.83 \pm 0.06$                      & $0.84 \pm 0.05$                    \\
                                                                    & \gls{mnist}                          & $0.96 \pm 0.01$                        & $0.15 \pm 0.21$                       & $0.50 \pm 0.22$                       & $0.94 \pm 0.01$                     & $0.61 \pm 0.23$                     & $0.90 \pm 0.08$                      & $0.90 \pm 0.04$                    \\
            \midrule \multirow{4}{*}{\rotatebox{90}{\textbf{Text}}} & \gls{trec}                           & $0.87 \pm 0.12$                        & $0.20 \pm 0.06$                       & $0.36 \pm 0.13$                       & $0.82 \pm 0.12$                     & $0.44 \pm 0.20$                     & $0.74 \pm 0.23$                      & $0.77 \pm 0.12$                    \\
                                                                    & \gls{agnews}                         & $0.73 \pm 0.09$                        & $0.25 \pm 0.02$                       & $0.39 \pm 0.13$                       & $0.66 \pm 0.08$                     & $0.56 \pm 0.10$                     & $0.62 \pm 0.08$                      & $0.64 \pm 0.10$                    \\
                                                                    & \gls{dbpedia}                        & $0.78 \pm 0.23$                        & $0.07 \pm 0.01$                       & $0.16 \pm 0.10$                       & $0.66 \pm 0.24$                     & $0.44 \pm 0.20$                     & $0.62 \pm 0.23$                      & $0.60 \pm 0.22$                    \\
                                                                    & \gls{imdb}                           & $0.61 \pm 0.04$                        & $0.50 \pm 0.01$                       & $0.51 \pm 0.02$                       & $0.59 \pm 0.04$                     & $0.55 \pm 0.03$                     & $0.58 \pm 0.04$                      & $0.59 \pm 0.04$                    \\
            \bottomrule
        \end{tabular}
    }
\end{table}

\paragraph{Result analysis.}
The stitching results are in \Cref{translation:tab:stitching:svm}. As expected, the \textit{absolute} encodings obtain a score comparable to random guessing while also considering fewer encoder combinations out of the possible ones due to the dimensionality mismatch between some of them.
These results show that the transformation relating to these pre-trained encoders is indeed mostly orthogonal:
(i) \texttt{ortho} and \texttt{affine}, the narrowest and the broadest transformation classes considered, are the better-performing translation methods. But while the former is obtained via a simple and efficient closed-form algorithm, the latter is SGD-optimized (\Cref{translation:sec:estimatingt}).
(ii) the \texttt{l-ortho} version improves or has small drops in performances over the \texttt{linear} transformation it is obtained from, confirming that the least squares procedure converges to an $\mathbf{R}$ which is almost orthogonal.
Note that these results demonstrate the feasibility of combining pre-trained models without the need for retraining or fine-tuning, with negligible drops in performances across the board, and without any additional assumption on the decoders. Please refer to \Cref{translation:sup:stitching:mlp:std,translation:sup:stitching:mlp:l2} in the Appendix for results with different decoders. In the Appendix (\Cref{translation:sup:fig:cross-domain}), we extend the cross-architecture transfer to decoders trained on different domains (styles) of the same \gls{cifart} dataset: the original one and a grayscale one.

\paragraph{Sensibility to Anchor Quantity.}
The number of anchors is an essential parameter in our approach. In \Cref{translation:fig:anchors-num}, we evaluate how the quantity of these anchors impacts the residual error and the overall performance of our method for this experimental setting. This analysis offers insights into the optimal number of anchors necessary for efficient latent space translation.

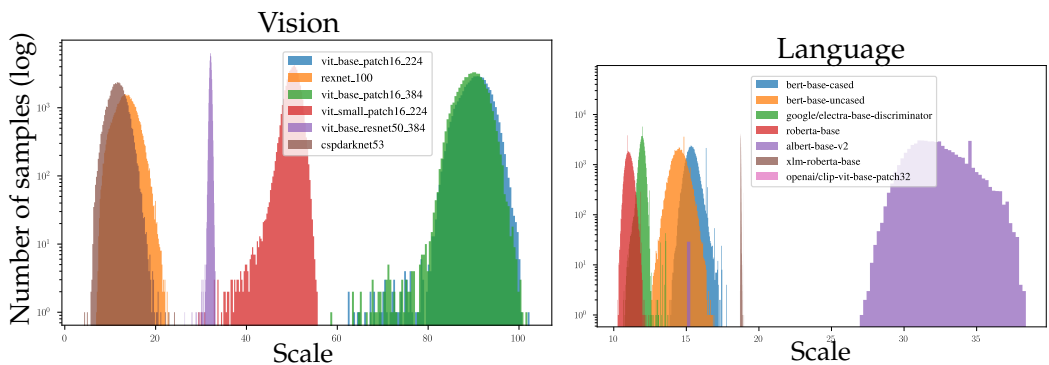
\begin{figure}[ht]
    \centering
    \begin{overpic}[trim={0 0 0 -1cm},clip,width=0.48\linewidth]{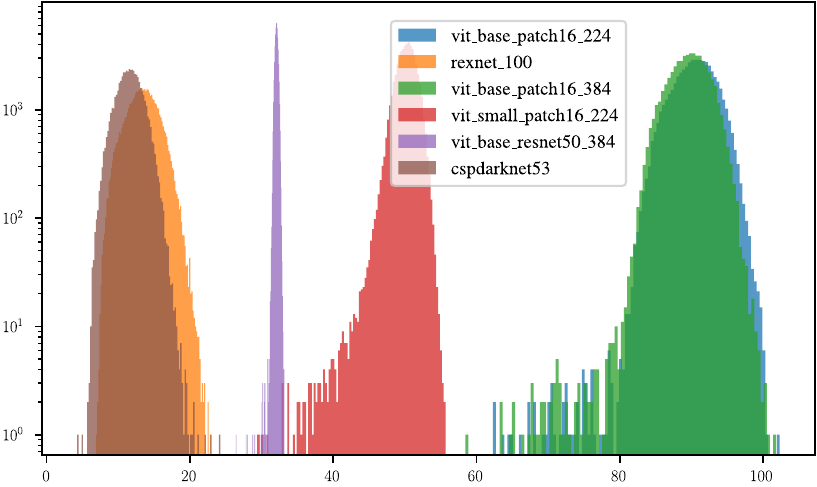}
        \put(-5, 4){\rotatebox{90}{\small Number of samples ($\log$)}}
        \put(46, -3.25){\small Scale}
        \put(43,61){Vision}
    \end{overpic}
    \,
    \begin{overpic}[trim={0 0 0 -1cm},clip,width=0.44\linewidth]{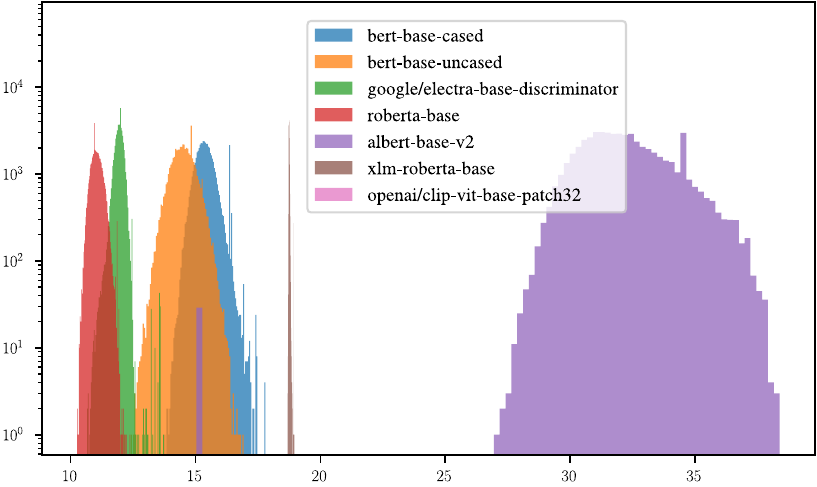}
        \put(46, -3.25){\small Scale}
        \put(43,61){Language}
    \end{overpic}
    \caption[Scale distribution in encodings of different pre-trained encoders]{Scale distribution in encodings of different pre-trained encoders on the \gls{n24news} dataset.}
    \label{translation:fig:norm-ranges}
\end{figure}

\paragraph{Role of Scaling.}
Our approach is designed to accommodate generic (re)scaling methods as pre-processing steps. We advocate for the use of standard scaling, as it shows reliable performance in our experiments, indicating that the scale of the data points is useful in estimating the latent transformation ${\gls{transformationapprox}}$.

However, for completeness, we also consider L2 normalization, which is the standard normalization in \glsxtrshortpl{rr}. This normalization method generalizes the class of transformations handled by our method and introduces an element of complete scale invariance. It is important to note that when this level of generalization is introduced, a scale-invariant decoder is required, since the norm information is effectively removed. In \Cref{chap:relative}, this is implicitly accomplished by training a decoder on \glsxtrshortpl{rr}.
In our setting, since we do not train the decoder, we just assume it is scale invariant; in \Cref{translation:sup:scale-invariance} we elaborate why this is a reasonable assumption that happens in practice.

This investigation exemplifies the flexibility of our approach, capable of adapting to different normalization and pre-processing strategies based on the specific requirements of the task at hand.
The results presented in \Cref{translation:tab:stitching:svm:l2}, when compared with \Cref{translation:tab:stitching:svm}, indicate a stronger reliance on the information encoded in the norm in the text modality. This is aligned with existing literature in the \gls{nlp} domain \citep{oyama2023norm}, which suggests that the scale of the encodings contains information (e.g., it is correlated with the token frequency).

These results in diverse scenarios showcase the flexibility and adaptability of our method, especially its robustness in translating between latent spaces of different dimensionality and domains.

\subsection{Cross-Modality}\label{translation:sec:cross-modality}

This scenario illustrates the applicability of our method in cross-modality settings, where we aim to translate between text and image latent spaces.

\paragraph{Experimental setting.}
We adopt \gls{n24news}, a multimodal news classification dataset that contains both text and associated pictures. We apply the standard encoding procedure to these two features separately, using different pre-trained uni-modal encoders. Then, we train a classification head (an \glsxtrshort{svm}, please refer to Appendix \Cref{translation:sup:cross-modality} for further results employing an \glsxtrshort{mlp} as classification head) on top of each one. Lastly, we zero-shot stitch each encoder with a classification head different from its corresponding one, measuring its classification accuracy, without further training or fine-tuning.

\begin{figure}[ht]
    \centering
    \begin{minipage}{.61\columnwidth}
        \begin{overpic}[trim=-1cm -1cm 0 0,width=1\linewidth]{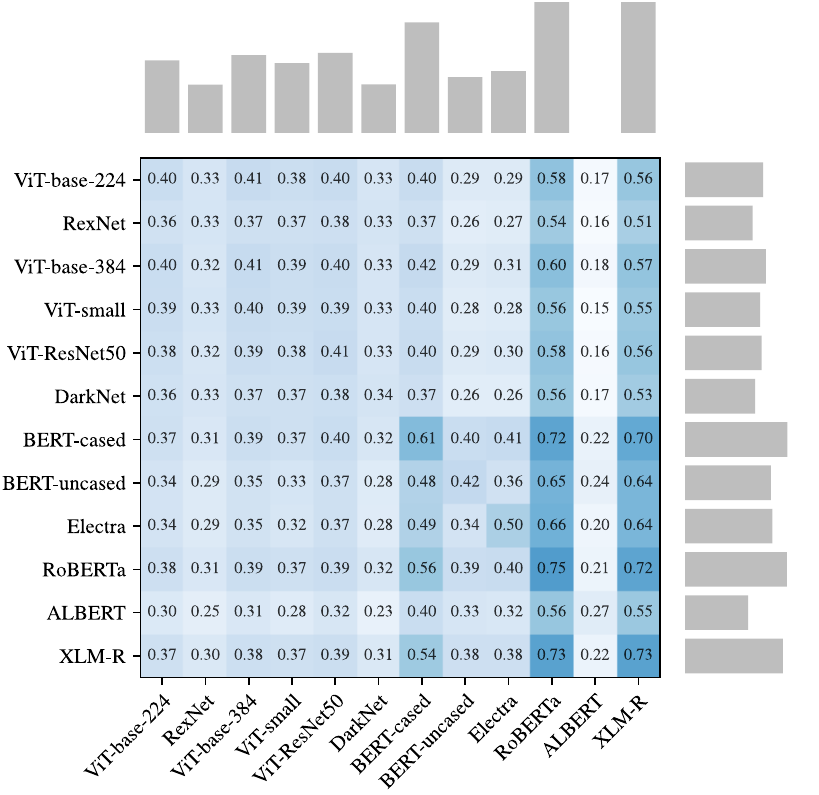}
            \put(0.5, 44){\rotatebox{90}{ Decoder}}
            \put(45, 2){ Encoder}
        \end{overpic}
    \end{minipage}
    \begin{minipage}{.375\columnwidth}
        \scriptsize
        \begin{tabular}{llcc}
            \toprule
            {}                                               & \textbf{Encoder}           & \textbf{Score} & \textbf{Scale} \\
            \midrule
            \multirow{5}{*}{\rotatebox{90}{\textbf{Vision}}} & \glsxtrshort{vitbp16224}   & 0.40           & 90.45          \\
                                                             & \glsxtrshort{rexnet100}    & 0.33           & 13.46          \\
                                                             & \glsxtrshort{vitbp16384}   & 0.41           & 89.66          \\
                                                             & \glsxtrshort{vitsp16224}   & 0.39           & 50.17          \\
                                                             & \glsxtrshort{vitbr50384}   & 0.41           & 32.10          \\
                                                             & \glsxtrshort{cspdarknet53} & 0.34           & 11.62          \\
            \midrule

            \multirow{6}{*}{\rotatebox{90}{\textbf{Text}}}   & \glsxtrshort{bertbc}       & 0.61           & 15.43          \\
                                                             & \glsxtrshort{bertbu}       & 0.42           & 14.54          \\
                                                             & \glsxtrshort{electrabd}    & 0.50           & 11.94          \\
                                                             & \glsxtrshort{robertab}     & 0.75           & 11.06          \\
                                                             & \glsxtrshort{albertbv2}    & 0.27           & 32.27          \\
                                                             & \glsxtrshort{xlmrobertab}  & 0.73           & 18.75          \\
            \bottomrule
        \end{tabular}
    \end{minipage}

    \caption[Performance comparison between different encoders and data modalities]{Performance comparison between different encoders and data modalities on  the \gls{n24news} multimodal dataset. On the right the accuracy of models trained end-to-end on a single data modality (Score) and their average norm (Scale). On the left the stitching performance between pairs of encoders and decoder. This shows the importance of translating from good encoders, that can even improve unimodal decoder performances. Results obtained with $2000$ anchors and ortho, with an \glsxtrshort{svm} as classification head. In the Appendix \Cref{translation:sup:cross-modality}, additional results using \glsxtrshortpl{mlp} as decoders.}
    \label{translation:fig:cross-modality-stitching}
\end{figure}

\paragraph{Scale distributions.}\label{translation:sec:scale-distribution}
In \Cref{translation:fig:norm-ranges}, we present the scale distribution of the embeddings produced by several encoders on the \gls{n24news} dataset.
This empirical analysis shows a consistent pattern among encoders: the scale distribution of their embeddings follows a Gaussian one with a single mode and a well-defined mean, which are usually compatible with standard scaling. This consistent behavior across encoders is likely attributed to their architectural choices, such as the normalization techniques, regularizations and the optimization problems they are designed to solve.

\paragraph{Result analysis.}
The discrepancy in the mean accuracy represented by the marginal bar plots in \Cref{translation:fig:cross-modality-stitching} is a signal that can be used to identify spaces more suited to be \textit{decoded into} and the ones that are stronger in \textit{encoding from}. In fact, the language models as source space for the translation exhibit stronger performance than the vision encoders. We relate this behavior to the higher generality of the text domain data used during pre-training with respect to the image domain one \citep{Zhai_2022_CVPR}. A remarkable finding in this setting is the improvement in classification performance when a modality-specific classifier trained on images is fed zero-shot with corresponding text encodings translated to the image domain via our method. This result underlines the significance of a good encoder and demonstrates the broad applicability of our technique. In practice, this means we can seamlessly apply image classifiers on textual data, and vice versa.

These results show that our method: (i) obtains effective zero-shot translation over different modalities; (ii) improves unimodal decoders when translating from a better encoder than the one it was trained on.

\subsection{Autoencoding}

In this setting, our method is applied to align latent spaces of different trainings of the same \gls{ae}. The novelty of this scenario lies in the generation setting itself, as most prior works (\Cref{related:align}) primarily focus on classification tasks.
One key observation explored in \Cref{chap:bridge} is that the \textit{task} at hand (e.g., classification, generation) defines a certain \textit{class of transformations} \gls{transformationclass} (e.g., rotations) which act among the latent spaces.  To ensure the best possible performance and efficiency, it is essential to limit the search for the transformation to the appropriate class.

\begin{figure}[ht]

    \hfill \begin{overpic}[trim=-0.27cm 0cm 0cm 0cm, clip,width=.96\linewidth,]{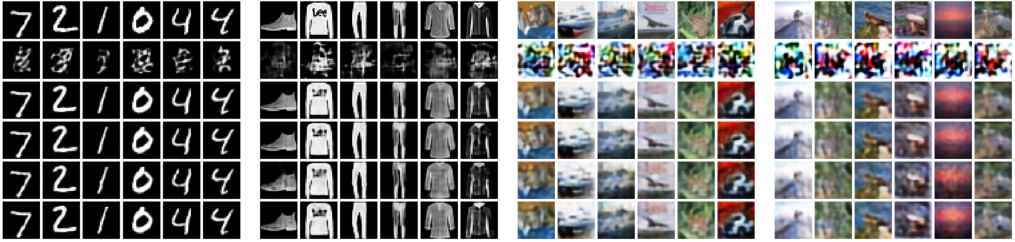}
        \put(.1, 20.7){\rotatebox{0}{\tiny \texttt{S}}}
        \put(-1.5, 16.7){\rotatebox{0}{\tiny \texttt{Abs.}}}
        \put(-3.4, 13){\rotatebox{0}{\tiny \texttt{affine}}}
        \put(-3.4, 9){\rotatebox{0}{\tiny \texttt{linear}}}
        \put(-4.25, 5.2){\rotatebox{0}{\tiny \texttt{l-ortho}}}
        \put(-2.75, 1.5){\rotatebox{0}{\tiny \texttt{ortho}}}
    \end{overpic}

    \caption[Translation reconstruction examples grouped by dataset]{Reconstruction examples grouped by dataset. Each column is a different image, from top to bottom: original image, \texttt{absolute} stitching, \texttt{affine} stitching \texttt{linear} stitching, \texttt{l-ortho} stitching, and \texttt{ortho} stitching. No additional normalization applied on the decoder part. Please refer to \Cref{translation:sup:fig:aes-decoders-with-l2-norm,translation:sup:fig:more-aes-decoders-with-l2-norm} in the Appendix for decoders trained with L2 normalization.}
    \label{translation:fig:aes-decoders-without-l2-norm}
\end{figure}
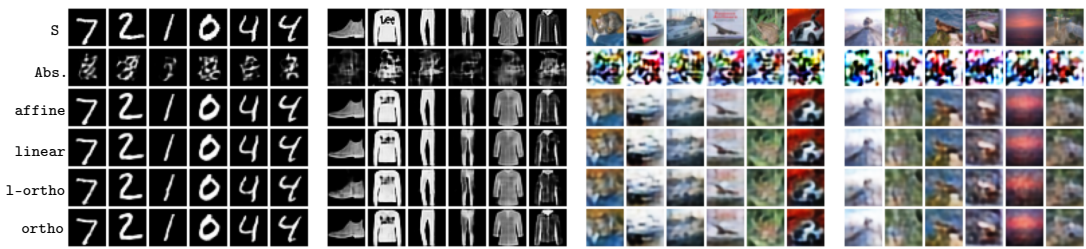

\paragraph{Experimental setting.}
We utilize four datasets for these experiments, namely  \gls{mnist},  \gls{fmnist}, \gls{cifart} and \gls{cifarh}. For each dataset, we train two standard \gls{cnn}-based \gls{ae}, with convolutions in the encoder and deconvolutions in the decoder, please refer to the Appendix for further implementation details. The two \glspl{ae} are identical in structure, differing only in the random seed used for weight initialization and data shuffling. To perform Zero-Shot Stitching, we first translate each data point from the latent space of the first encoder to the latent space of the second using $1000$ parallel anchors. We then apply the second decoder to the translated data, without any additional training or fine-tuning.

\paragraph{Result analysis.}

\begin{table}
    \small
    \caption[Zero-shot stitching for generation with various ${\gls{transformationapprox}}$]{Zero-shot stitching for generation with various methods for estimating ${\gls{transformationapprox}}$. The representation is normalized using Standard Scaling, and no additional normalization is applied to the stitched decoders. We report the latent cosine similarity (\textit{lcos}) and MSE (\textit{lmse}) between the target encoding and the translated one, but also the reconstruction MSE (\textit{rmse}) between the input and the output. The absolute space dimension is 500, and we used 1000 anchors. Please refer to \Cref{translation:sup:table:aes-decoder-with-norm-1000-anchors} for results on decoders scale-invariant by design (with L2 normalization on the encodings). }
    \label{translation:table:aes-decoder-without-norm-1000-anchors}
    \centering
    \scriptsize
    \begin{tabular}{lrrrrrrrrrrrr}
        \toprule
                          & \multicolumn{3}{c}{    \gls{mnist} } & \multicolumn{3}{c}{    \gls{fmnist} } & \multicolumn{3}{c}{   \gls{cifart} } & \multicolumn{3}{c}{    \gls{cifarh}}                                                                                                                 \\
        \cmidrule(lr){2-4} \cmidrule(lr){5-7} \cmidrule(lr){8-10} \cmidrule(lr){11-13}
                          & \emph{lcos}                          & \emph{lmse}                           & \emph{rmse}                          & \emph{lcos}                          & \emph{lmse} & \emph{rmse} & \emph{lcos} & \emph{lmse} & \emph{rmse} & \emph{lcos} & \emph{lmse} & \emph{rmse} \\
        \midrule
        \texttt{absolute} & 0.09                                 & 0.27                                  & 0.14                                 & 0.17                                 & 0.23        & 0.23        & 0.30        & 0.29        & 0.34        & 0.34        & 0.53        & 0.40        \\
        \texttt{affine}   & 0.94                                 & 0.08                                  & 0.02                                 & 0.94                                 & 0.06        & 0.03        & 0.96        & 0.03        & 0.05        & 0.96        & 0.04        & 0.05        \\
        \texttt{linear}   & 0.92                                 & 0.09                                  & 0.02                                 & 0.93                                 & 0.07        & 0.04        & 0.94        & 0.03        & 0.05        & 0.94        & 0.04        & 0.06        \\
        \texttt{l-ortho}  & 0.79                                 & 0.14                                  & 0.02                                 & 0.78                                 & 0.12        & 0.05        & 0.85        & 0.05        & 0.06        & 0.84        & 0.07        & 0.07        \\
        \texttt{ortho}    & 0.90                                 & 0.10                                  & 0.02                                 & 0.90                                 & 0.08        & 0.04        & 0.94        & 0.03        & 0.06        & 0.93        & 0.04        & 0.06        \\
        \bottomrule
    \end{tabular}
\end{table}

This experiment analyzes the alignment of latent spaces in different training regimens of the same \gls{ae}.
The performance evaluation, as shown in \Cref{translation:table:aes-decoder-without-norm-1000-anchors}, demonstrates that all methods \texttt{affine}, \texttt{linear}, \texttt{l-ortho}, and \texttt{ortho} yield satisfactory results. Moreover, qualitative results depicted in \Cref{translation:fig:aes-decoders-without-l2-norm} reveals minimal visual differences in the stitching outcomes across various datasets using different methods. Please refer to \Cref{translation:sup:fig:aes-decoders-with-l2-norm,translation:sup:fig:more-aes-decoders-with-l2-norm} for other qualitative results.
In fact, these results suggest that the latent spaces of image \glspl{ae} are not exclusively correlated by orthogonal transformations.
Consequently, in order to constrain and improve their approximation, more research is necessary to investigate and model the particular class of transformations that control the correlation between \glspl{nn} during image autoencoding.
For additional results pertaining to decoders with L2 normalization on their input, we refer to the \Cref{translation:sup:table:aes-decoder-with-norm-1000-anchors} in the Appendix.

Overall these results, combined with \cite{cannistraci2023} presented in \Cref{chap:bridge} and \Cref{translation:sec:stitch-cross-architecture}, confirm that latent spaces in image \glspl{ae} trained end-to-end are related by a class of transformations larger than orthogonal transformations.

\part[Overcoming Limitations in Latent Communication]{Overcoming Limitations\\ in Latent Communication}

\chapter{Current limitations}\label{chap:limitations}

The methodologies explored in \Cref{chap:relative,chap:translation} have demonstrated significant potential in addressing the \glsfirst{lcp} illustrated in \Cref{fig:formalization} and detailed in \Cref{chap:problemformalization}. Despite these advancements, there exist major constraints within these approaches that merit further discussion.

\paragraph{Assumptions on the transformation class \gls{transformationclass}.}
The approaches delineated in \Cref{chap:relative,chap:translation} presuppose a \emph{known} transformation class
\gls{transformationclass} between latent manifold embeddings $\gls{phiZX}(\gls{ZMx}) \subseteq \gls{ZX}$ and $\gls{phiZY}(\gls{ZMy}) \subseteq \gls{ZY}$.
Specifically, \Cref{chap:relative} assumes that \gls{transformationclass} comprises transformations preserving angle norms,
whereas \Cref{chap:translation} assumes it to be simple, exploring different possibilities (i.e., either affine, linear or orthogonal). However, this assumption does not always align with practical scenarios.
Indeed, the latent manifolds embeddings are subject to changes due to several factors \gls{randomfactors}, as explained in \Cref{sec:randomfactors}, and the precise nature of \gls{transformationclass} connecting these embeddings often remains undetermined a priori.

\paragraph{Observable partial correspondence \gls{pi}.}

Both methodologies assume that a partial correspondence \gls{pi} between the input spaces exists and, most importantly, that it is at least partially observable through the parallel anchors $\gls{parallelanchors} \subseteq \gls{pi}$. This premise, however, is not universally applicable, as the parallel anchors $\gls{parallelanchors}$ are typically not available in large quantities.
The assumption that \gls{parallelanchors} is sufficiently large to define \glsxtrshortpl{rr} without losing information, in \Cref{chap:relative}, and to accurately estimate the transformation $\gls{transformationapprox} \approx \gls{transformation}$, in \Cref{chap:translation}, does not hold in many practical instances.
This limitation is particularly evident in multimodal data contexts. Here, the available partial correspondence $\gls{parallelanchors}$ often falls short of the threshold necessary for the effective application of these methodologies, especially when considering domains different from images-text pairs.

\bigskip
In the following Chapters, we will explore methods to overcome these limitations.
In \Cref{chap:bridge}, we introduce a novel approach to tackle the \gls{lcp} without any specific assumption on the transformation class \gls{transformationclass}. Meanwhile, in \Cref{chap:bootstrapping} we delineate a methodology capable of discovering new parallel anchors from a limited known set, thereby expanding $\gls{parallelanchors} \subseteq \gls{pi}$ and facilitating the communication between these spaces.

\Chapter{Unknown Latent Transformation}{From Bricks to Bridges: Product of Invariances to Enhance Latent Space Communication\footnote{\fullcite{cannistraci2023}}}
\label{chap:bridge}

\begin{quotation}
    \noindent
    In this Chapter, we address the \gls{lcp} outlined in \Cref{chap:problemformalization}, without imposing any additional explicit assumptions on the transformation class \gls{transformationclass} that connects the latent manifold embeddings, $\gls{phiZX}(\gls{ZMx}) \subseteq \gls{ZX}$ and $\gls{phiZY}(\gls{ZMy}) \subseteq \gls{ZY}$.
    Leveraging the \glsxtrshortpl{rr} framework introduced in \Cref{chap:relative}, we define both $\gls{Tx}$ and \gls{Ty} as multiple relative projections, each characterized by distinct similarity functions \gls{sim}.
    This approach enables the construction of a product space of invariant components, obviating the need for pre-existing knowledge about the specific invariances to be incorporated.
\end{quotation}

\section{Introduction}
\label{bridge:sec:intro}

\begin{figure}[t]
    \centering
    \includegraphics[width=.65\linewidth]{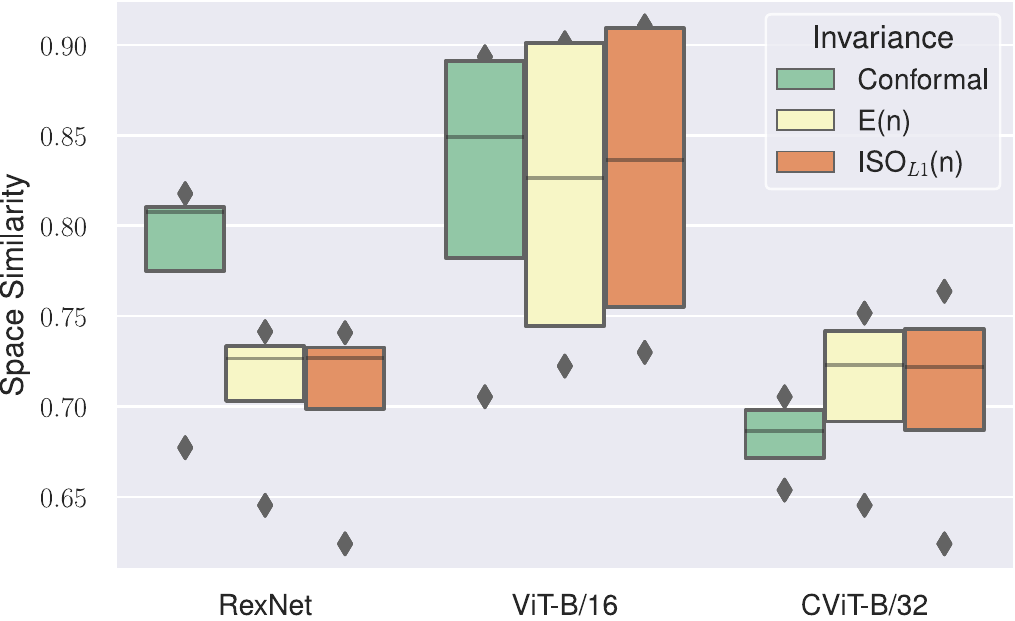}
    \caption[CKA similarity of pretrained models on \glsxtrlong{fmnist}]{CKA similarity \citep{Kornblith2019-sz} of pretrained models on \gls{fmnist}, measured on projections of the latent space onto subspaces invariant to specific classes of transformations (Conformal\protect\footnotemark, Euclidean, Orthogonal).
        In each bar, we report the distribution of similarity to the other models while infusing a specific invariance.
        The score diversity highlights the absence of a universal transformation connecting all latent spaces.}
    \label{bridge:fig:teaser}
\end{figure}

Achieving invariance to specific groups of transformations within latent spaces
is at the core of solving the \gls{lcp}.
In fact, the \glsxtrshortpl{rr} framework presented in \Cref{chap:relative} enables communication between latent spaces by infusing an invariance to angle-norm preserving transformations in them.
However, as shown in \Cref{bridge:fig:teaser}, the transformations relating different latent manifold embeddings are not always consistently within a specific class of transformations \gls{transformationclass}.
Determining a priori which \gls{transformationclass} relates distinct latent manifold embeddings is challenging due to complex interactions in the data, and multiple nuisance factors that are typically irrelevant but can nevertheless affect the representation, i.e., the factors \gls{randomfactors} outlined in \Cref{sec:randomfactors}.

To address this challenge, we expand upon the method of \gls{rr}, presenting a framework to \textit{efficiently incorporate a set of invariances into the learned latent space}. This is achieved by constructing a product space of invariant components on top of the latent representations of, possibly pretrained, neural models. Each component of this product space is a \gls{rr} produced with a different similarity function \gls{sim}. Thus, we can infuse invariances to specific transformation classes into each component of the product space.

Our main contributions can be summarized as follows:
\begin{itemize}
    \item We show that the transformation class \gls{transformationclass} that relates latent manifold embeddings learned by distinct \glsxtrshortpl{nn} -- trained on data semantically related by \gls{pi} -- may vary and directly depends on the factors \gls{randomfactors}.
    \item We introduce a framework to construct a product space of invariant components and improve the \gls{lcp} solution proposed in \Cref{chap:relative}; achieving the best performance without any prior knowledge of the transformation class \gls{transformationclass} or the factors \gls{randomfactors} that may affect it.
    \item  We validate our findings on classification and reconstruction tasks, observing consistent latent similarity and downstream performance improvements in the Zero-Shot Stitching setting  (\Cref{sec:stitchingdefinition}). The experimental analysis comprises three modalities (vision, text, and graphs), twelve pretrained foundational models, eight benchmarks, and several architectures trained from scratch.
\end{itemize}

\footnotetext{More precisely, global orthogonal transformations composed with local rescalings.}

\section{Infusing invariances}
\label{bridge:sec:method}

In this Chapter, our focus is to leverage different choices of the similarity function \gls{sim} in the \gls{rr} framework to induce a \emph{set of invariances} into the representations to capture complex transformations between latent spaces. Meanwhile, in \Cref{chap:relative}, $\gls{sim}$ was always the cosine similarity, inducing a \gls{rr} invariant to {angle-norm preserving transformations}.

\paragraph{Overview.} As illustrated in \Cref{fig:formalization} and \Cref{chap:problemformalization}, when considering different encoders $\gls{enc}_{\gls{X}},\gls{enc}_{\gls{Y}}$, we are interested in modeling the class of transformations $\gls{transformationclass}$ that relates their latent manifold embeddings $\gls{phiZX}(\gls{ZMx}) \subseteq \gls{ZX}$ and $\gls{phiZY}(\gls{ZMy}) \subseteq \gls{ZY}$.
In general, this transformation can be induced by any change in the factors \gls{randomfactors}  which could affect $\gls{phiZX}(\gls{ZMx})$ and $\gls{phiZY}(\gls{ZMy})$ in unpredictable ways, as observed in \Cref{bridge:fig:teaser}, e.g., their training dynamics, by architectural changes, or even domain changes.
The resulting $\gls{transformationclass}$ could be something known, e.g., rotations, or a nontrivial, complex class of transformations.

\subsection{Method}
What we look for are the transformations \gls{Tx} and \gls{Ty}, as per \Cref{fig:formalization}, that independently projects the latent manifold embeddings $ \gls{phiZX}(\gls{ZMx})$ and $\gls{phiZY}(\gls{ZMy})$ into a universal space $\gls{U}$, where they become the same $\gls{Tx}(\gls{phiZX}(\gls{ZMx})) = \gls{Ty}(\gls{phiZY}(\gls{ZMy})) \subseteq \gls{U}$.
This is achieved by independently enforcing an \emph{invariance to \gls{transformationclass}} in each space, i.e.,
\begin{equation}
    \gls{Tx}(\gls{zmx}) = \gls{Tx}(\gls{transformation}\gls{zmx})
    \qquad
    \forall\, \gls{transformation} \in \gls{transformationclass}
    \quad
    \text{and}
    \quad
    \forall\, \gls{zmx} \in \gls{phiZX}(\gls{ZMx}),
\end{equation}
and similarly for $\gls{Ty}$. Generalizing the \glsxtrshortpl{rr} to arbitrary similarity functions \gls{sim}, or distance metrics, gives us a straightforward way to define representations invariant to specific classes of transformations.

However, \gls{transformationclass} is typically unknown a priori, and it is also improbable to accurately characterize it as a singular, well-defined class of transformations (as observed in \Cref{bridge:fig:teaser} and \Cref{bridge:sec:exp-spaces-similarity}).
To overcome this, we approximate ${\gls{U}}$  with a product space $\gls{productspace}:= \prod_{i=1}^N {\gls{U}}_i $,  where each component is obtained by projecting samples of \gls{ZX} and \gls{ZY} in a \gls{rr} equipped with a different similarity function $\gls{sim}_{i}$. %
Each ${\gls{U}}_i$ will have properties induced by a similarity function $\gls{sim}_{i}$ invariant to a specific, known, class of transformations ${{\gls{transformationclass}}_i}$ (e.g., dilations). By combining this set of invariances, we want to construct \gls{Tx} and \gls{Ty} such that they are invariant to $\gls{transformationclass}$.
We define $\gls{Tx}$, and equivalently $\gls{Ty}$, formally as the \textit{product projection} from $\gls{ZX}$ to ${\gls{productspace}}$: %
\begin{definition}[Product projection]
    Given a latent space \gls{ZX} produced by some encoding function $\gls{enc}_{\gls{X}}$, a set of encoded anchors $\gls{ZA}_{\gls{X}}$, the definition of relative projection \gls{relativeprojection} in \Cref{relative:eq:relative-projection}, and a set of similarity functions $\mathcal{D}$ each one invariant to a specific known class of transformations $\gls{transformationclass}_i$ (e.g., rotations),
    i.e.,
    $
        \gls{relativeprojection}(\gls{zx}; \gls{ZA}_{\gls{X}}, \gls{sim}_i)
        =
        \gls{relativeprojection}(  \gls{transformation} \gls{zx};  \gls{transformation} \gls{ZA}_{\gls{X}}, \gls{sim}_i )
        \;\forall {\gls{transformation}} \in {{\gls{transformationclass}}}_i$.
    We define \gls{Tx} as a product projection:
    \begin{equation}
        \gls{Tx}(\gls{zx}) = {\gls{aggregation}} \circ \gls{relativeprojection}(\gls{zx}; \gls{ZA}_{\gls{X}}, \gls{sim}_i) \qquad \forall\, \gls{sim}_i \in \mathcal{D}
    \end{equation}
    where ${\gls{aggregation}}$ is an aggregation function (further details in \Cref{bridge:sec:merging_subspaces}) responsible for merging the relative spaces induced by each ${\gls{sim}}_i \in \mathcal{D}$. %
\end{definition}

\paragraph{Distance-induced invariances.}
We leverage the \gls{rr} framework considering the following similarity functions $\gls{sim}$: \gls{cos}, \gls{eu}, \gls{l1}, and \gls{linf}, each one inducing invariances to a specific, known class of transformations. In \Cref{bridge:tab:invariances}, we summarize the invariances guaranteed by different distance metrics concerning the following standard classes of transformations: \gls{is}, \gls{ot}, \gls{tr}, \gls{pt}, \gls{at} and \gls{lt}.

\begin{table}[ht]
    \caption[Invariances summary]{{Invariances summary.} Overview of the different distance-induced invariances.}
    \label{bridge:tab:invariances}
    \centering
    \small
    \begin{tabular}{cccccccc}
        \toprule
        Similarity & \gls{is}                                                & \gls{ot}                                                & \gls{tr}                                                & \gls{pt}                                                & \gls{at}                             & \gls{lt}                             \\
        \midrule
        Absolute   & $\textcolor[rgb]{0.5, 0, 0}{\times}$                    & $\textcolor[rgb]{0.5, 0, 0}{\times}$                    & $\textcolor[rgb]{0.5, 0, 0}{\times}$                    & $\textcolor[rgb]{0.5, 0, 0}{\times}$                    & $\textcolor[rgb]{0.5, 0, 0}{\times}$ & $\textcolor[rgb]{0.5, 0, 0}{\times}$ \\
        \gls{cos}  & $\textcolor{green}{ \textcolor[rgb]{0, 0.5, 0}{\surd}}$ & $\textcolor{green}{ \textcolor[rgb]{0, 0.5, 0}{\surd}}$ & $\textcolor[rgb]{0.5, 0, 0}{\times}$                    & $\textcolor{green}{ \textcolor[rgb]{0, 0.5, 0}{\surd}}$ & $\textcolor[rgb]{0.5, 0, 0}{\times}$ & $\textcolor[rgb]{0.5, 0, 0}{\times}$ \\
        \gls{eu}   & $\textcolor[rgb]{0.5, 0, 0}{\times}$                    & $\textcolor{green}{ \textcolor[rgb]{0, 0.5, 0}{\surd}}$ & $\textcolor{green}{ \textcolor[rgb]{0, 0.5, 0}{\surd}}$ & $\textcolor{green}{ \textcolor[rgb]{0, 0.5, 0}{\surd}}$ & $\textcolor[rgb]{0.5, 0, 0}{\times}$ & $\textcolor[rgb]{0.5, 0, 0}{\times}$ \\
        \gls{l1}   & $\textcolor[rgb]{0.5, 0, 0}{\times}$                    & $\textcolor[rgb]{0.5, 0, 0}{\times}$                    & $\textcolor{green}{ \textcolor[rgb]{0, 0.5, 0}{\surd}}$ & $\textcolor{green}{ \textcolor[rgb]{0, 0.5, 0}{\surd}}$ & $\textcolor[rgb]{0.5, 0, 0}{\times}$ & $\textcolor[rgb]{0.5, 0, 0}{\times}$ \\
        \gls{linf} & $\textcolor[rgb]{0.5, 0, 0}{\times}$                    & $\textcolor[rgb]{0.5, 0, 0}{\times}$                    & $\textcolor{green}{ \textcolor[rgb]{0, 0.5, 0}{\surd}}$ & $\textcolor{green}{ \textcolor[rgb]{0, 0.5, 0}{\surd}}$ & $\textcolor[rgb]{0.5, 0, 0}{\times}$ & $\textcolor[rgb]{0.5, 0, 0}{\times}$ \\
        \bottomrule
    \end{tabular}
\end{table}

Note how, in general, it is not straightforward to characterize the set of invariances induced by a similarity function. For example, the $L_\infty$ distance does not enforce isometry invariance (in the $L_2$ sense of rigidity) in the representation, but induces an invariance to perturbations in dimensions that are not the maximum one. Formalizing and analyzing such types of invariances presents challenges since these transformations cannot be neatly classified into a specific simple class of transformations.

\subsection{Aggregation functions.} \label{bridge:sec:merging_subspaces} %
This section summarizes the different aggregation strategies \gls{aggregation} to construct the product space $\gls{productspace}$:
\begin{itemize}
    \item \textit{Concatenation} (Concat): the subspaces are independently normalized and concatenated.
    \item \textit{Aggregation by sum} (\glsxtrshort{mlp}+Sum): similar to DeepSet \citep{zhang2019deep}, the subspaces are independently normalized and non-linearly projected. The resulting components are summed.
    \item \textit{Self-attention} (SelfAttention): the subspaces are independently normalized and aggregated via a self-attention layer.
\end{itemize}
When not specified, all the results are obtained using the \textit{Aggregation by sum} strategy. For the implementation details of each strategy, please refer to \cite{cannistraci2023}.

The product space \gls{productspace} is a \textit{robust} latent representation, made of \textit{invariant} components which are combined to capture \textit{nontrivial, complex} transformations, improving \gls{lcp} solutions and boosting the performance on downstream tasks.

\section{Experiments}
\label{bridge:sec:exp}
In this Section, we perform qualitative and quantitative experiments to analyze the effectiveness of our framework in constructing representations invariant to complex \gls{transformationclass}. Specifically, \Cref{bridge:sec:exp-spaces-similarity} provides empirical motivation, implicitly analyzing the transformations classes that emerge between different pretrained models on multiple datasets and modalities (i.e., vision and text); meanwhile, \Cref{bridge:sec:exp-downstream-task} evaluates the Zero-Shot Stitching performance of our framework across text, vision, and graph modalities;
finally, \Cref{bridge:sec:exp-subspace-selection} examines attention weights and their role in selecting the optimal relative subspace.
Refer to \cite{cannistraci2023} for further experiments and details.

\subsection{Latent space analysis} \label{bridge:sec:exp-spaces-similarity}\label{bridge:sec:exp-pretrained-similarity}

\paragraph{Experimental setting.} In this Section, we analyze the similarity of latent spaces produced by pretrained foundational models in both the vision and text domains. For the vision domain, we evaluated five distinct foundational models (either convolutional or transformer-based) using the {\gls{cifart}}, {\gls{cifarh}}, {\gls{mnist}}, and \gls{fmnist} datasets. Meanwhile, in the text domain, we assessed seven different foundational models using the \gls{dbpedia}, \gls{trec} (coarse), and \gls{n24news} (Text) datasets.
\begin{figure}[ht]
    \centering
    \begin{overpic}[trim=0 0 0 0,clip,width=.475\linewidth]{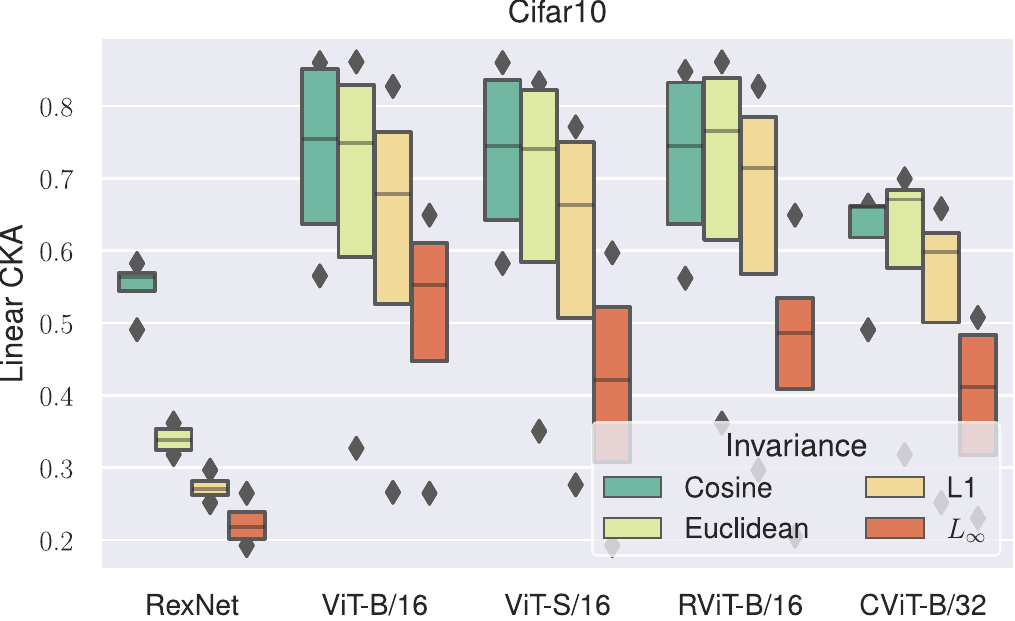}
    \end{overpic}
    \hfill
    \begin{overpic}[trim=0 0 0 0,clip,width=.475\linewidth]{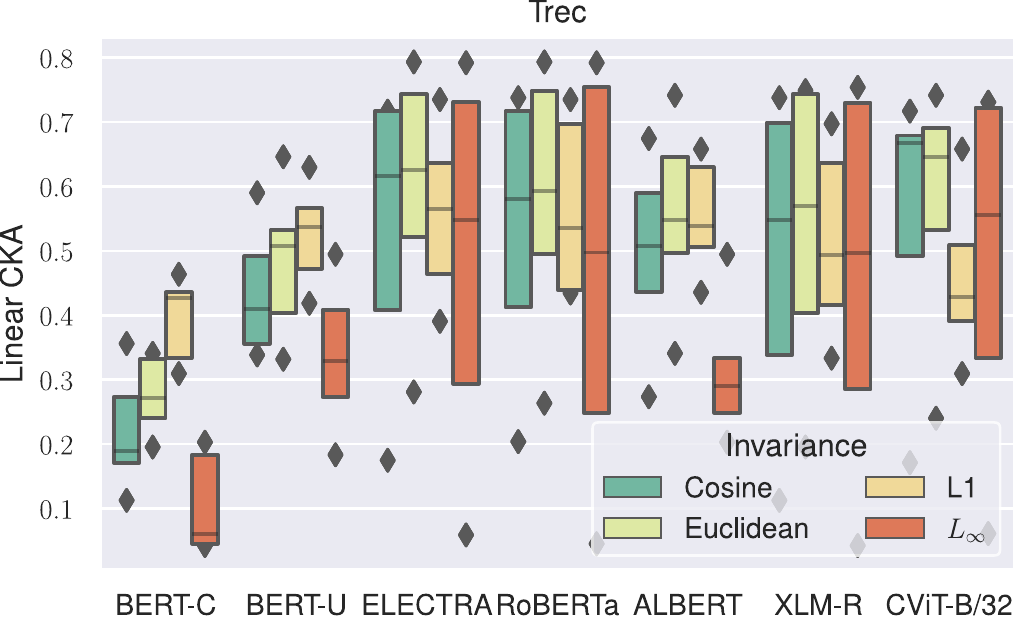}
    \end{overpic}
    \caption[Latent Spaces Cross-Architecture Similarity]{{Latent Spaces Cross-Architecture Similarity.} Linear \gls{cka} similarity of latent spaces across several
        pretrained models and datasets.
        In each bar, we report the space similarities distribution to the other models while infusing a specific invariance.
        There is no singular projection that consistently outperforms others across all configurations.}
    \label{bridge:fig:pretrained-main-cka}
\end{figure}

\paragraph{Result Analysis.} In \Cref{bridge:fig:pretrained-main-cka}, we report the Linear \gls{cka} correlations for various models and datasets for vision (\textit{left}) and text (\textit{right}) modalities. This analysis highlights the absence of a universally shared transformation class that connects latent spaces of foundation models across distinct conditions.
For example, on {\gls{cifart}} (\emph{left}), the highest similarity is achieved with different projection functions when using different architectures.
Interestingly, from \Cref{bridge:fig:pretrained-main-cka}, it is possible to see that similar architectures (i.e., ViT-based models) exhibit similar trends; hinting at the possibility that the choice of architecture plays a major role in influencing the emerging transformation class \gls{transformationclass}.

\paragraph{Takeaway.} The transformation class \gls{transformationclass} that correlates different latent manifold embeddings produced by different models depends on the dataset, architecture, modality, and possibly other factors \gls{randomfactors}.

\subsection{Zero-Shot Stitching} \label{bridge:sec:exp-downstream-task}

The Zero-Shot Stitching (\Cref{sec:stitchingdefinition}) methodology allows combining components of different \glsxtrshortpl{nn} to obtain a new model, where each element of the stitched model functions as an autonomous frozen module: the encoder handles data embedding, while the dedicated relative decoder manages the downstream task.

\begin{table}[ht]
    \caption[Graph and Text Stitching Performance]{{Graph and Text Stitching Performance.} Zero-shot accuracy scores across various decoders, seeds, and datasets. In the text domain, results are obtained from stitching across pretrained models, while in the graph domain, we train \gls{gnn} models from scratch and evaluate the stitching across seeds. Using the \textit{Aggregation by sum} (\textit{last row}) we consistently achieve the best performance.}
    \label{bridge:tab:text-graph-partial-stitching}
    \small
    \centering
    \begin{tabular}{lccc}
        \toprule
                                            & \multicolumn{2}{c}{Text}                    & Graph                                               \\
        \cmidrule(l){2-3}
        \cmidrule(l){4-4}
                                            & \multicolumn{2}{c}{\glsxtrshort{albertbv2}} & \glsxtrshort{gnn}                                   \\
        \cmidrule(l){2-3}
        \cmidrule(l){4-4}
        Projection                          & \gls{dbpedia}                               & \gls{cora}               & \gls{cora}               \\
        \midrule
        Cosine                              & $0.50\pm 0.02$                              & $0.54 \pm 0.03$          & $\textbf{0.53} \pm 0.06$ \\
        Euclidean                           & $0.50 \pm 0.00$                             & $0.60 \pm 0.03$          & $0.27 \pm 0.06$          \\
        $L_1$                               & $\textbf{0.52} \pm 0.01$                    & $\textbf{0.65} \pm 0.02$ & $0.26 \pm 0.06$          \\                $L_{\infty}$                     & $0.18 \pm 0.02$                     & $0.29 \pm 0.06$          & $0.12 \pm 0.03$          \\
        \cmidrule(l){1-4}
        Cosine,Euclidean,$L_1$,$L_{\infty}$ & $\textbf{0.53} \pm 0.01$                    & $\textbf{0.65} \pm 0.02$ & $\textbf{0.77} \pm 0.01$ \\
        \bottomrule
    \end{tabular}
\end{table}

\paragraph{Experimental setting.} We perform Zero-Shot Stitching classification using text, vision, and graph modalities with various models and datasets.
For the \textit{Vision} and \textit{Text} domains, we used the same datasets and pretrained models employed in \Cref{bridge:sec:exp-pretrained-similarity}. For the \textit{Graph} domain, we employed the \gls{cora} dataset \citep{corapubmed} and a \gls{gnn} architecture trained from scratch.
Relative decoders are trained with three different seed values, and the resulting representations are transformed into \glsxtrshortpl{rr} by projecting the encodings onto 1280 randomly selected but fixed anchors.

\begin{table}[ht]
    \caption[Image Stitching Performance Cross-Architecture and Cross-Seed]{{Image Stitching Performance Cross-Architecture and Cross-Seed.} Zero-shot accuracy score across different pretrained models, seeds, and datasets. The proposed method with \textit{Aggregation by sum}
        consistently achieves the highest accuracy score or comparable results, even without prior knowledge of the optimal projection to employ.}
    \label{bridge:tab:images-stitching}
    \centering
    \footnotesize
    \resizebox{\textwidth}{!}{%
        \begin{tabular}{llcccc}
            \toprule
                                                         &                                     & \multicolumn{4}{c}{Accuracy ↑}                                                                                  \\
            \cmidrule(l){3-6}
            Encoder                                      & Projection                          & \gls{cifarh}                   & \gls{cifart}             & \gls{mnist}              & \gls{fmnist}             \\
            \midrule
            \multirow[t]{5}{*}{\glsxtrshort{clip}}       & Cosine                              & $0.52 \pm 0.03$                & $\textbf{0.87} \pm 0.02$ & $0.61 \pm 0.06$          & $0.68 \pm 0.02$          \\
                                                         & Euclidean                           & $0.53 \pm 0.02$                & $\textbf{0.87} \pm 0.02$ & $\textbf{0.66} \pm 0.05$ & $\textbf{0.70} \pm 0.03$ \\
                                                         & $L_1$                               & $\textbf{0.53} \pm 0.04$       & $\textbf{0.87} \pm 0.02$ & $\textbf{0.66} \pm 0.05$ & $\textbf{0.70} \pm 0.03$ \\
                                                         & $L_{\infty}$                        & $0.27 \pm 0.04$                & $0.52 \pm 0.04$          & $0.57 \pm 0.03$          & $0.55 \pm 0.01$          \\
            \cmidrule(l){2-6}
                                                         & Cosine,Euclidean,$L_1$,$L_{\infty}$ & $\textbf{0.58} \pm 0.03$       & $\textbf{0.88} \pm 0.02$ & $\textbf{0.68} \pm 0.05$ & $\textbf{0.70} \pm 0.01$ \\
            \midrule
            \multirow[t]{5}{*}{\glsxtrshort{vitbr50384}} & Cosine                              & $\textbf{0.79} \pm 0.03$       & $0.94 \pm 0.01$          & $0.69 \pm 0.04$          & $0.76 \pm 0.03$          \\
                                                         & Euclidean                           & $\textbf{0.79} \pm 0.03$       & $0.94 \pm 0.01$          & $\textbf{0.71} \pm 0.04$ & $0.77 \pm 0.03$          \\
                                                         & $L_1$                               & $0.77 \pm 0.04$                & $\textbf{0.95} \pm 0.01$ & $\textbf{0.71} \pm 0.04$ & $\textbf{0.79} \pm 0.03$ \\
                                                         & $L_{\infty}$                        & $0.31 \pm 0.03$                & $0.75 \pm 0.04$          & $0.61 \pm 0.05$          & $0.60 \pm 0.03$          \\
            \cmidrule(l){2-6}
                                                         & Cosine,Euclidean,$L_1$,$L_{\infty}$ & $\textbf{0.81} \pm 0.04$       & $\textbf{0.95} \pm 0.01$ & $\textbf{0.72} \pm 0.04$ & $0.76 \pm 0.04$          \\
            \midrule
            \multirow[t]{5}{*}{\glsxtrshort{vitbp16224}} & Cosine                              & $0.75 \pm 0.05$                & $\textbf{0.96} \pm 0.01$ & $0.59 \pm 0.05$          & $0.79 \pm 0.03$          \\
                                                         & Euclidean                           & $\textbf{0.76} \pm 0.05$       & $\textbf{0.96} \pm 0.01$ & $0.65 \pm 0.06$          & $\textbf{0.81} \pm 0.02$ \\
                                                         & $L_1$                               & $\textbf{0.76} \pm 0.06$       & $\textbf{0.96} \pm 0.01$ & $\textbf{0.66} \pm 0.07$ & $\textbf{0.81} \pm 0.02$ \\
                                                         & $L_{\infty}$                        & $0.42 \pm 0.02$                & $0.70 \pm 0.05$          & $0.42 \pm 0.05$          & $0.52 \pm 0.04$          \\
            \cmidrule(l){2-6}
                                                         & Cosine,Euclidean,$L_1$,$L_{\infty}$ & $\textbf{0.81} \pm 0.05$       & $\textbf{0.96} \pm 0.01$ & $\textbf{0.66} \pm 0.04$ & $0.80 \pm 0.04$          \\
            \bottomrule
        \end{tabular}}
\end{table}

\begin{table}[ht]
    \caption[Stitching Index Across Architectures and Seeds on \gls{cora}]{{Stitching Index Across Architectures and Seeds on \gls{cora}.} Composing different projections using the \textit{Aggregation by sum} (\textit{last row}) enables Zero-Shot Stitching \textit{without} any performance drop in this setting, ensuring competitive end-to-end performance.}
    \label{bridge:tab:stitching-index-graphs}
    \centering
    \small
    \begin{tabular}{lcc}
        \toprule
        Projection                          & Accuracy ↑               & Stitching index ↑ \\
        \midrule
        Absolute                            & $0.14 \pm 0.04$          & $0.18$            \\
        Cosine                              & $\textbf{0.53} \pm 0.06$ & $0.71$            \\
        Euclidean                           & $0.27 \pm 0.06$          & $0.58$            \\
        $L_1$                               & $0.26 \pm 0.06$          & $0.58$            \\
        $L_{\infty}$                        & $0.12 \pm 0.03$          & $\textbf{1.00}$   \\
        \midrule
        Cosine,Euclidean,$L_1$,$L_{\infty}$ & $\textbf{0.77} \pm 0.01$ & $\textbf{1.00}$   \\
        \bottomrule
    \end{tabular}
\end{table}

\paragraph{Results Analysis.} \Cref{bridge:tab:images-stitching,bridge:tab:text-graph-partial-stitching} present the performance of various projection functions for different modalities.
As previously observed in \Cref{bridge:sec:exp-spaces-similarity}, the experiments reveal the absence of a single optimal projection function across architectures, modalities, and even within individual datasets.
Our proposed method consistently achieves superior accuracy across most scenarios. It is important to emphasize that the dimensionality of each independent projection and the aggregated product space remains constant, ensuring fair comparison.

To compare the performance with the end-to-end reference (reported in \cite{cannistraci2023}), we also propose an additional evaluation metric named the \textit{Stitching Index} computed as the ratio between the stitching score and the end-to-end score. It measures how closely the stitching accuracy aligns with the original score, i.e., a stitching score of one indicates there is no drop in performance when stitching modules.
Results in \Cref{bridge:tab:stitching-index-graphs} highlight that our method enables Zero-Shot Stitching \textit{without} any performance drop in this setting, while still ensuring competitive end-to-end performance.

\paragraph{Takeaway.}
A product space with invariant components can improve the Zero-Shot Stitching performance, solving the \gls{lcp} defined in \Cref{chap:problemformalization}, without any prior knowledge of the class of transformation \gls{transformationclass} that relates the manifold embeddings.

\subsection{Subspace selection} \label{bridge:sec:exp-subspace-selection}
In the preceding sections, we discussed integrating individual and multiple invariances into the representation through various projection functions and appropriate aggregation strategies.
In this Section, we aim to analyze and understand if tuning only the aggregation strategy at stitching time is a reasonable cost for selecting the optimal subspace.
We focus on the \textit{Self-attention} aggregation, which is a single self-attention layer as described in \Cref{bridge:sec:merging_subspaces}, and fine-tune only the \gls{qkv} parameters (i.e., the ones responsible for subspace blending). Each subspace is generated by its own projection function. We remark that stitching-time fine-tuning is exclusive to this experiment.

\begin{figure}[ht]
    \centering
    \begin{overpic}[trim=0 0 0 -.6cm,clip,width=.95\linewidth]{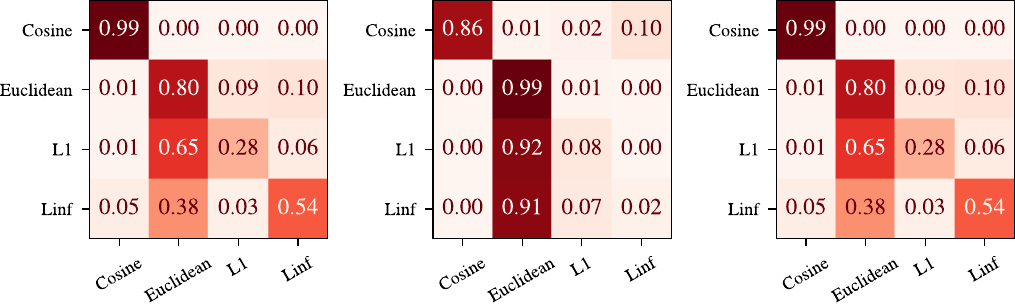}
        \put(16, 31.15){\texttt{Original}}
        \put(48.25, 31.15){\texttt{QKV opt}}
        \put(82.5, 31.15){\texttt{\glsxtrshort{mlp} opt}}
    \end{overpic}
    \caption[Comparison of attention weights before and after fine-tuning]{
        Comparison of attention weights for the stitched model with \glsxtrshort{rexnet100} as encoder and \glsxtrshort{vitbp16224} as decoder on \gls{cifarh}, before and after fine-tuning. \textit{(left)} the attention weights of the initial zero-shot stitched model, which remain unchanged \textit{(right)} when fine-tuning the classifier (\texttt{\glsxtrshort{mlp} opt}). Conversely, fine-tuning the \gls{qkv} projections (\texttt{QKV opt})  leads to a notable shift in attention weights \textit{(center)}, assigning lower importance to the subspace that performs worst individually.}
    \label{bridge:fig:attention-opt-cm}
\end{figure}

\paragraph{Experimental setup.} We identify two crucial components within the stitched model: (i) the linear projections associated with \gls{qkv} in the attention mechanism, which is responsible for selecting and blending subspaces, and (ii) the \glsxtrshort{mlp} in the classification head following the attention mechanism, which classifies the aggregated embeddings.
We examine two distinct approaches: the first approach fine-tunes only the first component (\texttt{QKV opt}), while the second one fine-tunes the second component (\texttt{\glsxtrshort{mlp} opt}). All the experiments in this Section are conducted on the \gls{cifarh} dataset, employing the \glsxtrshort{rexnet100} as encoder and the \glsxtrshort{vitbp16224} as decoder.

\begin{table}[ht]
    \caption[Classification accuracy with pretrained stitched models]{
        Classification accuracy for the stitched model with \glsxtrshort{rexnet100} as encoder and \glsxtrshort{vitbp16224} as decoder on \gls{cifarh}, using different projection functions and aggregation strategies. Fine-tuning the subspace selection and blending part (\texttt{QKV opt}) has a more significant effect on performance than fine-tuning only the larger \glsxtrshort{mlp} (\texttt{\glsxtrshort{mlp} opt}).}
    \label{bridge:tab:attention-opt-table}
    \centering
    \small
    \begin{tabular}{llc}
        \toprule
        Projection                             & Aggregation                                    & Accuracy ↑      \\
        \midrule
        Cosine                                 & -                                              & $\textbf{0.50}$ \\
        Euclidean                              & -                                              & $0.38$          \\
        $L_1$                                  & -                                              & $0.24$          \\
        $L_{\infty}$                           & -                                              & $0.21$          \\
        \midrule
        Cosine, Euclidean, $L_1$, $L_{\infty}$ & SelfAttention                                  & $0.17$          \\
        Cosine, Euclidean, $L_1$, $L_{\infty}$ & \glsxtrshort{mlp}+Sum                          & $\textbf{0.45}$ \\
        \midrule
        Cosine, Euclidean, $L_1$, $L_{\infty}$ & SelfAttention + \texttt{QKV opt}               & $\textbf{0.75}$ \\
        Cosine, Euclidean, $L_1$, $L_{\infty}$ & SelfAttention + \texttt{\glsxtrshort{mlp} opt} & $0.52$          \\
        \bottomrule
    \end{tabular}
\end{table}

\paragraph{Result Analysis.} \Cref{bridge:tab:attention-opt-table} summarizes downstream classification accuracy for the stitched model using various projection functions and aggregation strategies. Incorporating multiple invariances and aggregating them via \textit{Self-attention} (\textit{fifth row}) does not perform well; meanwhile, using the \textit{\glsxtrshort{mlp}+Sum} or the Cosine projection alone is more effective. This is expected, considering the attention mechanism is primarily trained to improve end-to-end performance rather than to maximize compatibility between different spaces.
Incorporating the adaptation strategies at stitching time significantly boosts performance, either focusing on the subspace selection and blending (\texttt{QKV opt}) or the classification head (\texttt{\glsxtrshort{mlp} opt}).
We find that an informed fine-tuning of the parameters responsible for the subspace blending (i.e., only the \gls{qkv} projections) significantly impacts performances, even more than tuning the whole classifier.
\Cref{bridge:fig:attention-opt-cm} illustrates the attention weights averaged over the test dataset:
the \emph{left} figure shows the attention weights of the zero-shot stitched model, that remain unchanged when fine-tuning only the classifier, as reported on the \emph{right}.
Meanwhile, the \emph{center} figure shows that fine-tuning the \gls{qkv} projection shifts the attention weights to allocate less importance to worse-performing projections~(i.e.,~$L_\infty$).

\paragraph{Takeaway.} Appropriate subspace selection and aggregation are crucial to further enhance latent communication between neural models.

\Chapter{Limited Semantic Correspondence}{Bootstrapping Parallel Anchors for Relative Representations\footnote{\fullcite{cannistraci2023a}}}
\label{chap:bootstrapping}

\begin{quotation}
    \noindent
    The previous \Cref{chap:relative,chap:translation,chap:bridge} presented potential solutions to solve the \glsxtrshort{lcp} (\Cref{chap:problemformalization}). Nevertheless, they all rely on a certain amount of parallel anchors $\gls{parallelanchors} \subseteq \gls{pi}$ to be given, which can be impractical to obtain in certain scenarios. To overcome this limitation, in this Chapter, we propose an optimization-based method to discover new parallel anchors from a limited known set \gls{parallelseeds}, denoted as \emph{seed}.
    Our approach expands the semantic correspondence between different domains, enabling the solution of the \glsxtrshort{lcp} in scenarios where it was previously not possible.
\end{quotation}

\section{Introduction}
\label{bootstrapping:introduction}

The previous Chapters presented potential solutions to solve the \gls{lcp}, either by projecting the latent manifold embeddings into a universal space \gls{U} or by directly approximating a specific transformation \gls{transformation} that relates $\gls{phiZX}(\gls{ZMx}) \subseteq \gls{ZX}$ and $\gls{phiZY}(\gls{ZMy}) \subseteq \gls{ZY}$.
Nevertheless, they all rely on the existence of an abstract correspondence \gls{C}, observable directly in the input spaces through \gls{pi} and provided in input as a certain amount of parallel anchors $\gls{parallelanchors} \subseteq \gls{pi}$.
However, obtaining a sufficient number of parallel anchors in specific applications can be challenging or impossible, hindering the use of the aforementioned methods.

In this Chapter, we focus on the scenario where only a very limited number of parallel anchors is available, denoted as \textit{seed} $\gls{seeds}$, and we aim to expand this initial set through an \gls{ao} process. Our method achieves competitive performance in NLP and Vision domains while significantly reducing the number of required parallel anchors by \textit{one order of magnitude}.

\begin{table}[ht]
    \caption[Qualitative and quantitative comparisons optimizing the \gls{word2vec} space]{Qualitative (\textit{left}) and quantitative (\textit{right}) comparisons of the three methods when optimizing over the \gls{word2vec} space, to discover the parallel anchors \gls{parallelanchors} between \gls{word2vec} and \gls{fasttext}. All metrics are calculated with $K=10$ averaged over 20k words across 5 random seeds. Refer to \Cref{relative:appendix:word-embeddings} for the metric definitions.}
    \label{bootstrapping:fig:w2v-latent-rotation-comparison-complete}
    \noindent\makebox[\textwidth][c]{%
        \begin{minipage}{.275\textwidth}
            \begin{overpic}[trim=-1cm 1cm -0.5cm 0cm,width=1\linewidth]{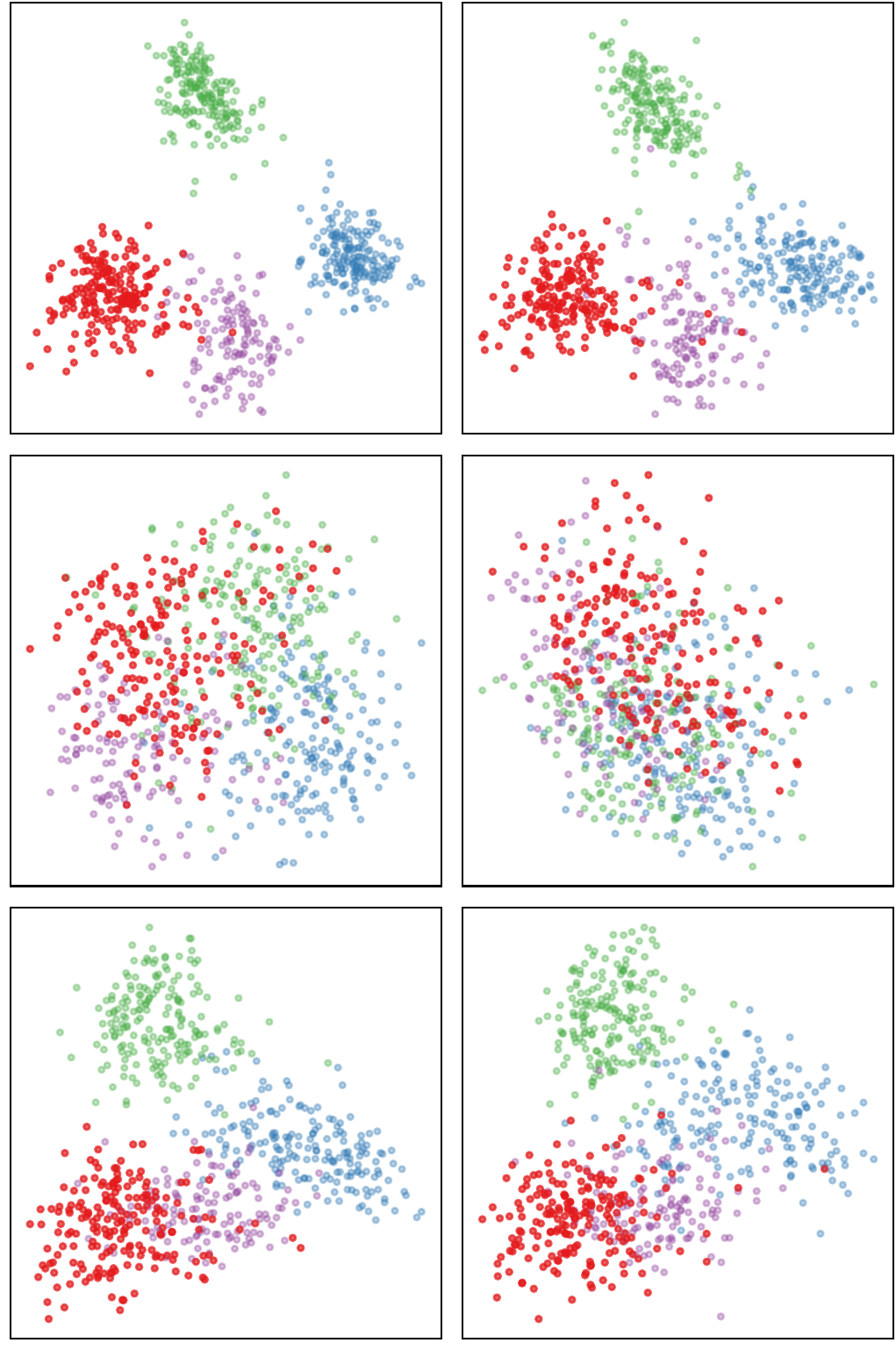}
                \put(5.5, 101.5){\gls{fasttext}}
                \put(40.5, 101.5){\gls{word2vec}}
                \put(-1, 79.5){\rotatebox{90}{\small GT}}
                \put(-1, 43){\rotatebox{90}{\small Seed}}
                \put(-1, 9){\rotatebox{90}{\small AO}}
            \end{overpic}
        \end{minipage}
        \hspace{0.1cm}
        \begin{minipage}{.675\textwidth}
            \footnotesize
            \begin{tabular}{lllllll}
                \toprule
                                                                           &  & Source                          & Target         & \multicolumn{1}{c}{Jaccard ↑} & \multicolumn{1}{c}{MRR ↑} & \multicolumn{1}{c}{Cosine ↑} \\
                \midrule
                \multirow{4}{*}{{\STAB{\rotatebox[origin=c]{90}{GT{}}}}}   &  & \multirow{2}{*}{\texttt{{FT}}}  & \texttt{{FT}}  & $1.00 \pm 0.00$               & $1.00 \pm 0.00$           & $1.00 \pm 0.00$              \\
                                                                           &  &                                 & \texttt{{W2V}} & $0.34 \pm 0.01$               & $0.94 \pm 0.00$           & $0.86 \pm 0.00$              \\[0.5ex]
                                                                           &  & \multirow{2}{*}{\texttt{{W2V}}} & \texttt{{FT}}  & $0.39 \pm 0.00$               & $0.98 \pm 0.00$           & $0.86 \pm 0.00$              \\
                                                                           &  &                                 & \texttt{{W2V}} & $1.00 \pm 0.00$               & $1.00 \pm 0.00$           & $1.00 \pm 0.00$              \\
                \midrule
                \multirow{4}{*}{{\STAB{\rotatebox[origin=c]{90}{Seed{}}}}} &  & \multirow{2}{*}{\texttt{{FT}}}  & \texttt{{FT}}  & $1.00 \pm 0.00$               & $1.00 \pm 0.00$           & $1.00 \pm 0.00$              \\
                                                                           &  &                                 & \texttt{{W2V}} & $0.06 \pm 0.01$               & $0.11 \pm 0.01$           & $0.85 \pm 0.01$              \\[0.5ex]
                                                                           &  & \multirow{2}{*}{\texttt{{W2V}}} & \texttt{{FT}}  & $0.06 \pm 0.01$               & $0.15 \pm 0.02$           & $0.85 \pm 0.01$              \\
                                                                           &  &                                 & \texttt{{W2V}} & $1.00 \pm 0.00$               & $1.00 \pm 0.00$           & $1.00 \pm 0.00$              \\
                \midrule
                \multirow{4}{*}{{\STAB{\rotatebox[origin=c]{90}{AO{}}}}}   &  & \multirow{2}{*}{\texttt{{FT}}}  & \texttt{{FT}}  & $1.00 \pm 0.00$               & $1.00 \pm 0.00$           & $1.00 \pm 0.00$              \\
                                                                           &  &                                 & \texttt{{W2V}} & $0.52 \pm 0.00$               & $0.99 \pm 0.00$           & $0.94 \pm 0.00$              \\[0.5ex]
                                                                           &  & \multirow{2}{*}{\texttt{{W2V}}} & \texttt{{FT}}  & $0.50 \pm 0.01$               & $0.99 \pm 0.00$           & $0.94 \pm 0.00$              \\
                                                                           &  &                                 & \texttt{{W2V}} & $1.00 \pm 0.00$               & $1.00 \pm 0.00$           & $1.00 \pm 0.00$              \\
                \bottomrule
            \end{tabular}
            \vspace{0.4cm}
        \end{minipage}
    }
\end{table}

\section{Method}
\label{bootstrapping:sec:method}

In this Section, we introduce an optimization procedure that reduces the required number of parallel anchors by one order of magnitude.
This method does not require complete knowledge of \gls{parallelanchors} but only of few initial \emph{seed} anchors, denoted as
${\gls{parallelseeds}}  \subseteq {\gls{parallelanchors}}$, where $|{\gls{seeds}}|\ll|{\gls{parallelanchors}}|$.
With no prior knowledge of ${\gls{anchors}}_{\gls{Y}}$, we initialize the optimization process by approximating its embeddings ${\mathbf{\gls{ZA}}}_{\gls{Y}}$ with the known seed embeddings:
$\gls{zseeds}_{\gls{Y}} = \bigoplus_{\gls{a} \in \gls{seeds}_{\gls{Y}}} \gls{enc}_{\gls{Y}}(\gls{a})$,
where $\bigoplus$ denotes row-wise concatenation,
concatenated with  $|{\gls{parallelanchors}}| - |{\gls{parallelseeds}}|$ random embeddings $\mathbf{N}$,
i.e.,
${\mathbf{\gls{ZA}}}_{\gls{Y}} =  \gls{zseeds}_{\gls{Y}} \oplus \mathbf{N}$ with $\mathbf{N} \sim \mathcal{N}(0,\mathbf{I})$.
Thus,
we define the following objective function optimizing over ${{\mathbf{\gls{ZA}}}}_{\gls{Y}}$:
\begin{equation}
    \argmin_{\substack{ \widetilde{{\mathbf{A}}}_{\gls{Y}}
            \text{ s.t. }
            {||a||}_2=1 \; \forall a \in  \widetilde{{\mathbf{A}}}_{\gls{Y}}
        }} \; \sum_{\gls{zy} \in {\gls{ZY}}}
    MSE(
    \gls{relativeprojection}(\Pi(\gls{zy}), {\gls{ZA}}_{\gls{X}}, \gls{sim} ),\,
    \gls{relativeprojection}(\gls{zy}, \mathbf{\gls{ZA}}_{\gls{Y}}, \gls{sim})
    )
\end{equation}
where \gls{sim} is the cosine similarity and $\Pi: {\gls{ZY}} \to {\gls{ZX}}$ is a correspondence estimated at each optimization step by the Sinkhorn algorithm \citep{cuturi2013sinkhorn} exploiting the initial seed and the current approximation of the remaining anchors:
\begin{equation}
    \Pi = {sinkhorn}\;
    (
    \gls{relativeprojection}(\gls{ZX}, {\gls{ZA}}_{\gls{X}}, \gls{sim} ),\,
    \gls{relativeprojection}(\gls{ZY}, \mathbf{{\gls{ZA}}}_{\gls{Y}}, \gls{sim} )
    ).
\end{equation}
After convergence, $\mathbf{{\gls{ZA}}}_{\gls{Y}}$ is discretized into ${{\gls{anchors}}}_{\gls{Y}} \subseteq {\gls{Y}}$ considering the nearest embeddings~in~\gls{ZY}.

\section{Experiments} \label{bootstrapping:sec:experiments}
This section assesses the effectiveness of the \gls{ao} method in reducing the reliance on parallel anchors \gls{parallelanchors} to the minimum necessary and automatically expanding the provided semantic correspondence between domains.

\begin{table}[ht]
       \footnotesize
       \centering
       \caption[Evaluation of the \glsxtrshort{ao} method in the vision domain]{Evaluation of the \gls{ao} method in the vision domain on \gls{cifart}. The table reports the mean results for each metric and its standard deviation across 5 different random seeds.}
       \label{bootstrapping:tab:quantitative-analysis-cifar}
       \begin{tabular}{lllllll}
              \toprule
              \multicolumn{1}{c}{{Mode}}                                      & \multicolumn{1}{c}{{Type}}                                   & \multicolumn{1}{c}{{Source}}                & \multicolumn{1}{c}{{Target}} & \multicolumn{1}{c}{{Jaccard ↑}} & \multicolumn{1}{l}{{MRR ↑}} & \multicolumn{1}{c}{{Cosine ↑}} \\
              \midrule
              \multirow{8}{*}{{\STAB{\rotatebox[origin=c]{90}{\textbf{GT}}}}} & \multirow{4}{*}{{\STAB{\rotatebox[origin=c]{90}{Absolute}}}} & \multirow{2}{*}{{\glsxtrshort{vitbp16224}}} & {\glsxtrshort{vitbp16224}}   & $1.00 \pm 0.00$                 & $1.00 \pm 0.00$             & $1.00 \pm 0.00$                \\
                                                                              &                                                              &                                             & {\glsxtrshort{vitsp16224}}   & \multicolumn{1}{c}{-}           & \multicolumn{1}{c}{-}       & \multicolumn{1}{c}{-}          \\[0.5ex]
                                                                              &                                                              & \multirow{2}{*}{{\glsxtrshort{vitsp16224}}} & {\glsxtrshort{vitbp16224}}   & \multicolumn{1}{c}{-}           & \multicolumn{1}{c}{-}       & \multicolumn{1}{c}{-}          \\
                                                                              &                                                              &                                             & {\glsxtrshort{vitsp16224}}   & $1.00 \pm 0.00$                 & $1.00 \pm 0.00$             & $1.00 \pm 0.00$                \\
              \cmidrule{2-7}
                                                                              & \multirow{4}{*}{{\STAB{\rotatebox[origin=c]{90}{Relative}}}} & \multirow{2}{*}{{\glsxtrshort{vitbp16224}}} & {\glsxtrshort{vitbp16224}}   & $1.00 \pm 0.00$                 & $1.00 \pm 0.00$             & $1.00 \pm 0.00$                \\
                                                                              &                                                              &                                             & {\glsxtrshort{vitsp16224}}   & $0.11 \pm 0.00$                 & $0.27 \pm 0.01$             & $0.97 \pm 0.00$                \\[0.5ex]
                                                                              &                                                              & \multirow{2}{*}{{\glsxtrshort{vitsp16224}}} & {\glsxtrshort{vitbp16224}}   & $0.10 \pm 0.00$                 & $0.28 \pm 0.01$             & $0.97 \pm 0.00$                \\
                                                                              &                                                              &                                             & {\glsxtrshort{vitsp16224}}   & $1.00 \pm 0.00$                 & $1.00 \pm 0.00$             & $1.00 \pm 0.00$                \\
              \cmidrule{1-7}
              \multirow{8}{*}{{\STAB{\rotatebox[origin=c]{90}{{Seed}}}}}      & \multirow{4}{*}{{\STAB{\rotatebox[origin=c]{90}{Absolute}}}} & \multirow{2}{*}{{\glsxtrshort{vitbp16224}}} & {\glsxtrshort{vitbp16224}}   & $1.00 \pm 0.00$                 & $1.00 \pm 0.00$             & $1.00 \pm 0.00$                \\
                                                                              &                                                              &                                             & {\glsxtrshort{vitsp16224}}   & \multicolumn{1}{c}{-}           & \multicolumn{1}{c}{-}       & \multicolumn{1}{c}{-}          \\[0.5ex]
                                                                              &                                                              & \multirow{2}{*}{{\glsxtrshort{vitsp16224}}} & {\glsxtrshort{vitbp16224}}   & \multicolumn{1}{c}{-}           & \multicolumn{1}{c}{-}       & \multicolumn{1}{c}{-}          \\
                                                                              &                                                              &                                             & {\glsxtrshort{vitsp16224}}   & $1.00 \pm 0.00$                 & $1.00 \pm 0.00$             & $1.00 \pm 0.00$                \\
              \cmidrule{2-7}
                                                                              & \multirow{4}{*}{{\STAB{\rotatebox[origin=c]{90}{Relative}}}} & \multirow{2}{*}{{\glsxtrshort{vitbp16224}}} & {\glsxtrshort{vitbp16224}}   & $1.00 \pm 0.00$                 & $1.00 \pm 0.00$             & $1.00 \pm 0.00$                \\
                                                                              &                                                              &                                             & {\glsxtrshort{vitsp16224}}   & $0.03 \pm 0.00$                 & $0.03 \pm 0.01$             & $0.97 \pm 0.00$                \\[0.5ex]
                                                                              &                                                              & \multirow{2}{*}{{\glsxtrshort{vitsp16224}}} & {\glsxtrshort{vitbp16224}}   & $0.03 \pm 0.00$                 & $0.04 \pm 0.01$             & $0.96 \pm 0.00$                \\
                                                                              &                                                              &                                             & {\glsxtrshort{vitsp16224}}   & $1.00 \pm 0.00$                 & $1.00 \pm 0.00$             & $1.00 \pm 0.00$                \\
              \cmidrule{1-7}
              \multirow{8}{*}{{\STAB{\rotatebox[origin=c]{90}{{AO}}}}}        & \multirow{4}{*}{{\STAB{\rotatebox[origin=c]{90}{Absolute}}}} & \multirow{2}{*}{{\glsxtrshort{vitbp16224}}} & {\glsxtrshort{vitbp16224}}   & $1.00 \pm 0.00$                 & $1.00 \pm 0.00$             & $1.00 \pm 0.00$                \\
                                                                              &                                                              &                                             & {\glsxtrshort{vitsp16224}}   & \multicolumn{1}{c}{-}           & \multicolumn{1}{c}{-}       & \multicolumn{1}{c}{-}          \\[0.5ex]
                                                                              &                                                              & \multirow{2}{*}{{\glsxtrshort{vitsp16224}}} & {\glsxtrshort{vitbp16224}}   & \multicolumn{1}{c}{-}           & \multicolumn{1}{c}{-}       & \multicolumn{1}{c}{-}          \\
                                                                              &                                                              &                                             & {\glsxtrshort{vitsp16224}}   & $1.00 \pm 0.00$                 & $1.00 \pm 0.00$             & $1.00 \pm 0.00$                \\
              \cmidrule{2-7}
                                                                              & \multirow{4}{*}{{\STAB{\rotatebox[origin=c]{90}{Relative}}}} & \multirow{2}{*}{{\glsxtrshort{vitbp16224}}} & {\glsxtrshort{vitbp16224}}   & $1.00 \pm 0.00$                 & $1.00 \pm 0.00$             & $1.00 \pm 0.00$                \\
                                                                              &                                                              &                                             & {\glsxtrshort{vitsp16224}}   & $0.10 \pm 0.01$                 & $0.23 \pm 0.03$             & $0.97 \pm 0.00$                \\[0.5ex]
                                                                              &                                                              & \multirow{2}{*}{{\glsxtrshort{vitsp16224}}} & {\glsxtrshort{vitbp16224}}   & $0.10 \pm 0.00$                 & $0.28 \pm 0.01$             & $0.97 \pm 0.00$                \\
                                                                              &                                                              &                                             & {\glsxtrshort{vitsp16224}}   & $1.00 \pm 0.00$                 & $1.00 \pm 0.00$             & $1.00 \pm 0.00$                \\
              \bottomrule
       \end{tabular}
\end{table}

\paragraph{Experimental Setting}
We utilize 15 seed anchors to approximate 300 parallel anchors that serve as ground truth in all downstream tasks. Specifically, we compare the performance of our method against two different baselines: (i) \textit{GT}, the Ground Truth employs all the anchors that our method aims to semantically approximate; and, (ii) \textit{Seed}, exploits only the seed anchors.
Refer to \Cref{relative:appendix:word-embeddings} for the metric definitions, and to \cite{cannistraci2023a} for complete details.

\paragraph{Retrieval Task.}\label{bootstrapping:sec:retrieval}
\gls{ao} effectively discovers new parallel anchors in the \gls{nlp} and Vision domains, as demonstrated in \Cref{bootstrapping:tab:quantitative-analysis-cifar,bootstrapping:fig:w2v-latent-rotation-comparison-complete}.
Specifically, we explore different word embeddings and pre-trained foundational visual encoders, and assess the quality of the discovered anchors through a retrieval task.
Results demonstrate that, when given the same number of starting anchors, our method outperforms the approach that relies solely on those without optimizing.
Our results are \textit{comparable} to those obtained employing all the ground truth parallel anchors.

\paragraph{Zero-Shot Stitching task.}\label{bootstrapping:sec:stitching}

\begin{table}[ht]
       \centering
       \scriptsize
       \caption[Cross-lingual Zero-Shot Stitching performance evaluation]{
              Cross-lingual Zero-Shot Stitching performance evaluation.
              The table reports the mean weighted F1 and \glsxtrshort{mae} on \gls{amazon} fine-grained across 5 random seeds.}
       \label{bootstrapping:tab:stitching-multilingual-opt}
       \resizebox{\textwidth}{!}{%
              \begin{tabular}{cccccccc}
                     \cmidrule(l){3-8}
                                         &         & \multicolumn{2}{c}{GT} & \multicolumn{2}{c}{Seed} & \multicolumn{2}{c}{AO}                                                           \\ \cmidrule(l){3-8}
                     Decoder             & Encoder & Fscore                 & \glsxtrshort{mae}        & Fscore                 & \glsxtrshort{mae} & Fscore          & \glsxtrshort{mae} \\
                     \midrule
                     \multirow{2}{*}{en} & en      & 0.64 $\pm$ 0.01        & 0.43 $\pm$ 0.01          & 0.50 $\pm$ 0.01        & 0.69 $\pm$ 0.01   & 0.62 $\pm$ 0.01 & 0.44 $\pm$ 0.01   \\
                                         & es      & 0.51 $\pm$ 0.01        & 0.67 $\pm$ 0.02          & 0.44 $\pm$ 0.01        & 0.80 $\pm$ 0.01   & 0.48 $\pm$ 0.01 & 0.70 $\pm$ 0.02   \\
                     \multirow{2}{*}{es} & en      & 0.50 $\pm$ 0.02        & 0.72 $\pm$ 0.04          & 0.41 $\pm$ 0.01        & 0.92 $\pm$ 0.02   & 0.46 $\pm$ 0.01 & 0.76 $\pm$ 0.02   \\
                                         & es      & 0.60 $\pm$ 0.01        & 0.45 $\pm$ 0.01          & 0.48 $\pm$ 0.01        & 0.70 $\pm$ 0.01   & 0.61 $\pm$ 0.01 & 0.44 $\pm$ 0.01   \\
                     \bottomrule
              \end{tabular}}
\end{table}

Furthermore, \Cref{bootstrapping:tab:stitching-multilingual-opt} demonstrates that our method can discover parallel anchors across different domains: the method finds aligned Amazon reviews in different languages with unavailable ground truth. Using only 15 out-of-domain anchors \gls{OODanchors} (refer to \Cref{relative:sec:ood-anchors} for their definition), our method enables Zero-Shot Stitching (\Cref{sec:stitchingdefinition}), allowing to train a classifier on one language and perform predictions on another one without any fine-tuning.

\part[Applying Latent Communication]{Applying Latent Communication}

\chapter{Case Studies}
\label{chap:casestudies}

In this Chapter, we examine three case studies that highlight the potential impacts of solving the \gls{lcp}. The first case, discussed in \Cref{chap:asif}, shows that it is possible to create a multimodal model from unimodal models, solving CLIP-like tasks without ever training a multimodal model. The second case, discussed in \Cref{chap:chart}, analyzes the possibility to merge latent spaces with differing sample and class compositions. This investigation provides insights into the theoretical and practical aspects of latent space manipulation, showcasing methods for the coherent integration of diverse datasets, which is crucial for the enhancement of model generalization and data utilization. Finally, the third case, outlined in \Cref{chap:rl}, explores the Zero-Shot Stitching between policies and encoders trained on different variants of the Car Racing environment. This case study illustrates the application of \gls{lcp} solutions in \gls{rl}, particularly in the transfer and generalization of policies across varied environment, underscoring the potential for policy reuse in \gls{rl} methodologies.

Through these case studies, this Chapter aims to showcase the broad applicability and significance of solving the \gls{lcp}, ranging from multimodal data processing to \gls{rl}. Please refer to \Cref{sec:otherworks} for additional applications of \gls{lcp} solutions.

\section[ASIF: Coupled Data Turns Unimodal Models to Multimodal Without Training]{ASIF: Coupled Data Turns Unimodal Models to Multimodal Without Training\protect\footnote{\fullcite{norelli2022}}}
\label{chap:asif}

Large multimodal models such as CLIP~\citep{clip} are rapidly becoming the standard for foundation models in computer vision. This is largely due to their zero-shot and open-world capabilities that enable diverse suites of downstream tasks, from classification to detection and visual search.
Still, training \glspl{nn} at such scale presents several challenges beside the obvious infrastructure and training costs. In fact, it requires collecting massive training sets, making it difficult to interpret the predictions of the model in light of their training data. Additionally, the training assets are often not owned by the institution training the model.

\begin{figure}[ht]
    \centering
    \begin{overpic}[trim=0cm 0cm 0cm 0cm,clip,width=0.8\linewidth]{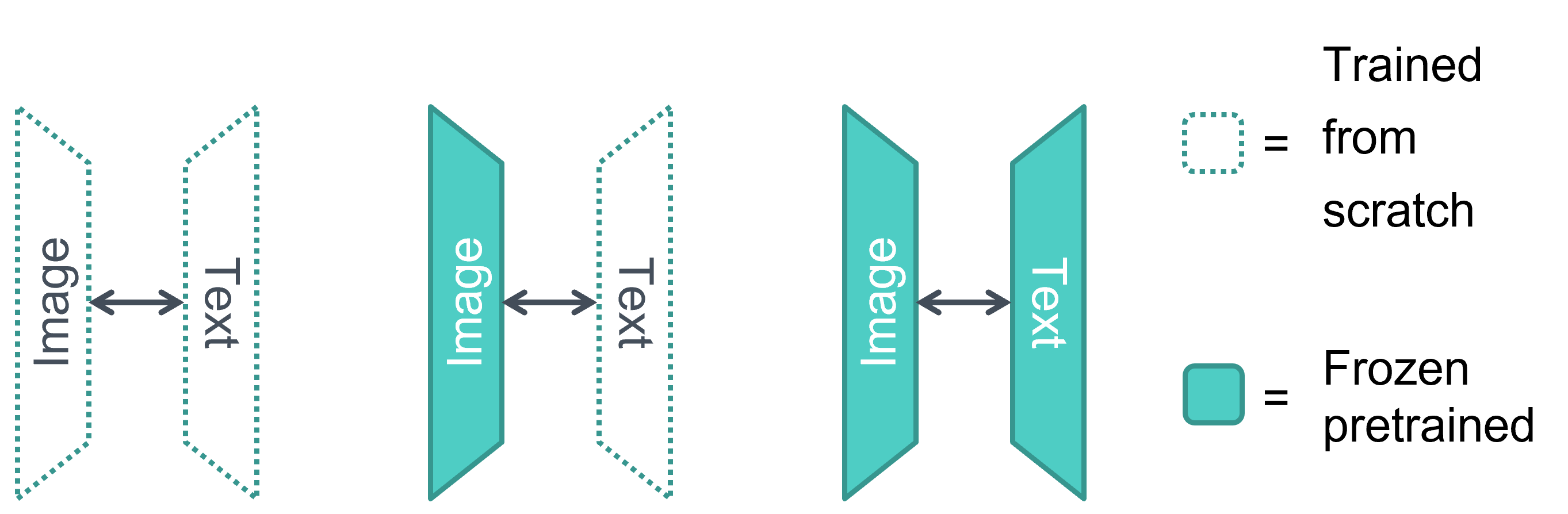}
        \put(4,28){\small CLIP}
        \put(31.8,28){\small LiT}
        \put(56.5,28){\small ASIF}
    \end{overpic}
    \caption[ASIF aligns latent spaces of frozen pre-trained encoders]{ASIF aligns latent spaces of frozen pre-trained encoders.}\label{asif:fig-models}
\end{figure}

In \cite{norelli2022}, we present ASIF, building on the \gls{rr} framework introduced in \Cref{chap:relative}, to turn pre-trained uni-modal image and text encoders into a multi-modal model using a \textit{relatively small}\footnote{CLIP \citep{clip} experiments used from 400M to 15M captioned images as training samples, LiT \citep{lit} from 901M to 10M. ASIF uses 1.6M.} collection of image-text pairs and no additional training, as depicted in \Cref{asif:fig-models}. The resulting model is functionally equivalent to CLIP, effectively producing aligned representations of images and their captions.
The key insight is that captions of similar images are themselves similar, and therefore a representation crafted using just similarities to ground-truth multimodal pairs is quasi modality-invariant.

\begin{table*}[ht]
    \caption[Zero shot classification accuracy of different multimodal designs]{ {Zero shot classification accuracy of different multimodal designs.} CLIP and LiT implementations vary by dataset and the visual transformer used as image encoder.
    }
    \label{asif:tab-main}
    \centering
    \resizebox{\textwidth}{!}{%
        \begin{tabular}{lccccc}
            \toprule
            {Method}                  & {Dataset size} & {ImageNet} & {CIFAR100} & {Pets} & {ImageNet v2} \\
            \midrule
            CLIP~\cite{clip}          & 400M (private) & 68.6       & 68.7       & 88.9   & -             \\
            CLIP~\cite{clip}          & 15M (public)   & 31.3       & -          & -      & -             \\
            LiT~\cite{lit}            & 10M (public)   & 66.9       & -          & -      & -             \\
            CLIP~\cite{lit}           & 901M (private) & 50.6       & 47.9       & 70.3   & 43.3          \\
            LiT~\cite{lit}            & 901M (private) & 70.1       & 70.9       & 88.1   & 61.7          \\
            \midrule
            ASIF (sup vis. encoder)   & 1.6M (public)  & 60.9       & 50.2       & 81.5   & 52.2          \\
            ASIF (unsup vis. encoder) & 1.6M (public)  & 53.0       & 46.5       & 74.7   & 45.9          \\
            \bottomrule
        \end{tabular}}
\end{table*}

As shown in \Cref{asif:tab-main},
despite the simplicity of the approach,
a multimodal dataset that is up to 250 times smaller than in prior work,
and the lack of actually training the model on multimodal data; ASIF achieves zero-shot classification accuracy on downstream datasets that is comparable to CLIP \citep{clip,lit}. For a comprehensive overview and discussion, please consult \cite{norelli2022}. The key points are summarized as follows:
\begin{itemize}
    \item The introduction of ASIF, a method that transforms two pre-existing frozen unimodal encoders into an interpretable multimodal model.
    \item The demonstration of ASIF efficacy in zero-shot image classification tasks, exhibiting comparable performance to CLIP while requiring significantly fewer image-text pairs.
\end{itemize}

\clearpage
\section[From Charts to Atlas: Merging Latent Spaces into One]{From Charts to Atlas: Merging Latent Spaces into One\protect\footnote{\fullcite{crisostomi2023from}}}
\label{chap:chart}

In \cite{crisostomi2023from}, we investigate a natural follow-up question: when, and under what assumptions, \emph{can two latent spaces be merged into one?}
In principle, given two comparable representations that may partially overlap, or be entirely disjoint, it should be possible to generate a unified representation in which both coexist consistently. We refer to this problem as \emph{Latent Space Aggregation}. Space aggregation raises several questions on (i) the representational power of the unified representation space, (ii) its ability to accommodate both spaces without collisions, and (iii) its robustness to complementary information present in only one of the two spaces. In fact, naively aggregating the sample representations by computing their mean {\em in absolute coordinates} would not account for the different latent configurations caused by the factors \gls{randomfactors}, resulting in an inconsistent aggregation of different entities based on spurious random factors.

\begin{figure}[ht]
  \centering
  \begin{overpic}[width=\linewidth]{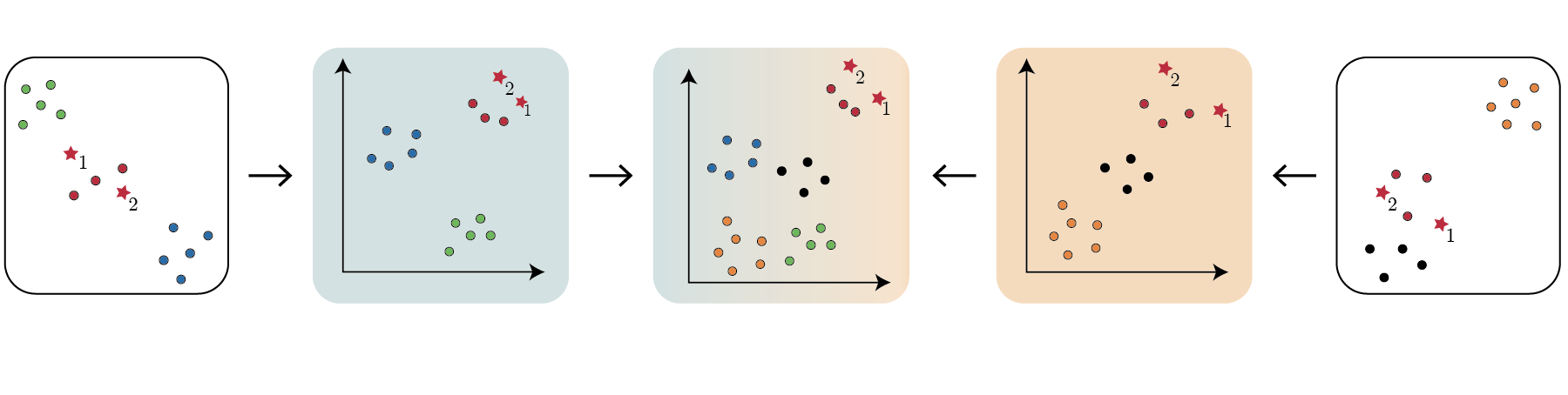}
    \put(6.5,-1){${\gls{ZX}}$}
    \put(91.5,-1){${\gls{ZY}}$}
    \put(27.5,-1){${\gls{ZX}}_\text{rel}$}
    \put(71.5,-1){${\gls{ZY}}_\text{rel}$}
    \put(43,-1){RLSA(${\gls{ZX}}_\text{rel}$, ${\gls{ZY}}_\text{rel}$)}

    \put(62, 23) {\small ${\gls{sim}}(\cdot,$ \includegraphics[height=5pt]{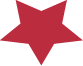}$_2)$}
    \put(40, 23){\small ${\gls{sim}}(\cdot,$ \includegraphics[height=5pt]{papers/chartatlas/figures/star.png}$_2)$}
    \put(18, 23){\small ${\gls{sim}}(\cdot,$ \includegraphics[height=5pt]{papers/chartatlas/figures/star.png}$_2)$}
    \put(75, 3.5){\small ${\gls{sim}}(\cdot,$ \includegraphics[height=5pt]{papers/chartatlas/figures/star.png}$_1)$}
    \put(52, 3.5){\small ${\gls{sim}}(\cdot,$ \includegraphics[height=5pt]{papers/chartatlas/figures/star.png}$_1)$}
    \put(31, 3.5){\small ${\gls{sim}}(\cdot,$ \includegraphics[height=5pt]{papers/chartatlas/figures/star.png}$_1)$}
  \end{overpic}
  \caption[\glsxtrlong{rlsa} description]{\glsxtrlong{rlsa} description. Given two absolute spaces ${\gls{ZX}}$ and ${\gls{ZY}}$, we first project these spaces into two comparable \gls{rr}s ${\gls{ZX}}_\text{rel}$, ${\gls{ZY}}_\text{rel}$. Then, we combine these representations into a single, unified relative space RLSA(${\gls{ZX}}_\text{rel}$, ${\gls{ZY}}_\text{rel}$).}
  \label{fig:teaser}
\end{figure}

Motivated by the above challenges, we propose \gls{rlsa} illustrated in \Cref{fig:teaser}.
The approach involves two steps: we first switch to a \gls{rr} (\Cref{chap:relative}) where the latent spaces are represented with respect to a set of anchors \gls{anchors}, and then aggregate the obtained representations of shared samples by computing their mean.
The first step makes the spaces comparable, enabling a meaningful aggregation of samples that are common to multiple latent spaces, at the same time avoiding collisions. %

\begin{table}
  \caption[\glsxtrlong{rlsa} classification accuracy comparison]{
    \glsxtrlong{rlsa} classification accuracy comparison. Each quarter shows a dataset-model combination, with end-to-end model accuracy on the right. For each $S$ (shared classes), $N$ (novel classes) combination, we report the accuracy of a classifier trained on the aggregated space over all the classes (\emph{total}), along with accuracy when considering only \emph{non-shared} and \emph{shared} classes. \emph{Improv} is the improvement over the end-to-end model, reported in the header, while \emph{vanilla} the accuracy of naive merging.
  }
  \label{tab:part-shared-part-novel-accuracy}
  \centering
  \resizebox{\textwidth}{!}{%
    \begin{tabular}{cccccccccccccc} %

      \toprule
       & $S$   & $N$  & tasks & vanilla                                 & non-shared & shared                                    & total & improv                                 & vanilla & non-shared & shared & total & improv                                 \\
      \midrule
      {\multirow{9}{*}{ \rotatebox[origin=c]{90}{\texttt{CIFAR100}} }}
       &       &      &       & \multicolumn{4}{c}{\texttt{VanillaCNN}} & 0.39       & \multicolumn{4}{c}{\texttt{EfficientNet}} & 0.70                                                                                                                            \\    \cmidrule(lr){5-9} \cmidrule(lr){10-14}
       & $80$  & $10$ & 2     & 0.36                                    & 0.60       & 0.39                                      & 0.43  & \cellcolor[rgb]{0.89, 0.94, 0.91}+0.04 & 0.68    & 0.80       & 0.71   & 0.73  & \cellcolor[rgb]{0.9, 0.94, 0.92}+0.02  \\
       & $60$  & $10$ & 4     & 0.39                                    & 0.64       & 0.45                                      & 0.53  & \cellcolor[rgb]{0.82, 0.9, 0.85}+0.14  & 0.72    & 0.82       & 0.76   & 0.79  & \cellcolor[rgb]{0.86, 0.92, 0.88}+0.08 \\
       & $40$  & $10$ & 6     & 0.42                                    & 0.64       & 0.50                                      & 0.58  & \cellcolor[rgb]{0.78, 0.87, 0.82}+0.19 & 0.75    & 0.87       & 0.80   & 0.84  & \cellcolor[rgb]{0.82, 0.9, 0.85}+0.14  \\
       & $20$  & $10$ & 8     & 0.47                                    & 0.65       & 0.52                                      & 0.62  & \cellcolor[rgb]{0.75, 0.86, 0.8}+0.23  & 0.80    & 0.88       & 0.84   & 0.87  & \cellcolor[rgb]{0.8, 0.88, 0.83}+0.17  \\
      \cmidrule(lr){2-14}
       & $80$  & $5$  & 4     & 0.37                                    & 0.77       & 0.41                                      & 0.49  & \cellcolor[rgb]{0.85, 0.91, 0.87}+0.10 & 0.71    & 0.84       & 0.72   & 0.75  & \cellcolor[rgb]{0.89, 0.93, 0.9}+0.05  \\
       & $60$  & $5$  & 8     & 0.39                                    & 0.71       & 0.45                                      & 0.55  & \cellcolor[rgb]{0.8, 0.89, 0.84}+0.16  & 0.76    & 0.85       & 0.78   & 0.81  & \cellcolor[rgb]{0.85, 0.91, 0.87}+0.11 \\
       & $40$  & $5$  & 12    & 0.44                                    & 0.74       & 0.49                                      & 0.64  & \cellcolor[rgb]{0.74, 0.85, 0.78}+0.25 & 0.80    & 0.90       & 0.80   & 0.86  & \cellcolor[rgb]{0.81, 0.89, 0.84}+0.16 \\
       & $20$  & $5$  & 16    & 0.51                                    & 0.76       & 0.55                                      & 0.72  & \cellcolor[rgb]{0.68, 0.82, 0.74}+0.33 & 0.83    & 0.93       & 0.83   & 0.90  & \cellcolor[rgb]{0.77, 0.87, 0.81}+0.20 \\
      \cmidrule(lr){1-14}
      {\multirow{3}{*}{ \rotatebox[origin=c]{90}{\texttt{TINY}} }}
       &       &      &       & \multicolumn{4}{c}{\texttt{VanillaCNN}} & 0.22       & \multicolumn{4}{c}{\texttt{EfficientNet}} & 0.69                                                                                                                            \\    \cmidrule(lr){5-9} \cmidrule(lr){10-14}
       & $100$ & $25$ & 4     & 0.22                                    & 0.37       & 0.23                                      & 0.30  & \cellcolor[rgb]{0.87, 0.92, 0.89}+0.08 & 0.68    & 0.75       & 0.71   & 0.73  & \cellcolor[rgb]{0.89, 0.93, 0.9}+0.05  \\
       & $50$  & $25$ & 6     & 0.24                                    & 0.36       & 0.36                                      & 0.36  & \cellcolor[rgb]{0.82, 0.9, 0.85}+0.14  & 0.72    & 0.77       & 0.74   & 0.77  & \cellcolor[rgb]{0.86, 0.92, 0.88}+0.08 \\
      \bottomrule
    \end{tabular}
  }
\end{table}

To test the \gls{rlsa} framework, we partition a classification dataset into multiple learning \emph{tasks}. These tasks can vary in terms of class composition, such as covering disjoint subsets of classes, or in sample composition, such as being sampled with different class distributions. These diverse tasks enable us to train task-specific models, extract their latent spaces, and subsequently examine their aggregation.
We consider three different cases: (i) tasks sharing a set of samples, (ii) tasks sharing the same classes but disjoint sample sets, and (iii) tasks disjoint both at the class and at the sample level.
In the first case, we select the anchors from the shared samples, while in the disjoint scenarios they are sampled from unseen samples in the training dataset.
We then analyze the quality of the aggregation by (i) comparing it to the space of an end-to-end model trained on all the tasks, (ii) assessing the performance of a classifier over the aggregated space, as reported in \Cref{tab:part-shared-part-novel-accuracy}, and (iii) quantifying the separability of the classes within it.

For a complete description of the experiments and results, refer to \cite{crisostomi2023from}. To summarize, the main contributions are three-fold:
\begin{enumerate}
  \item The introduction of a novel framework for Latent Space Aggregation, which, for the first time, enables the merging of different latent spaces without requiring weight averaging, sharing, or any model-specific details.
  \item The evaluation of the proposed framework on aggregating tasks sharing samples, classes, or neither, assessing representational power, class separability, and similarity to the global space.
  \item The analysis of the improved performance on class-disjoint tasks, empirically demonstrating that it is a natural consequence of utilizing task-specific embedders.
\end{enumerate}

\clearpage
\section[Zero-Shot Stitching in Reinforcement Learning]{Zero-Shot Stitching in Reinforcement Learning\protect\footnote{\fullcite{ricciardi2023}}}
\label{chap:rl}

In the domain of \gls{rl}, it is commonplace to train agents from scratch in an end-to-end manner, training both the feature extractor and the policy together. This methodology, while effective for singular tasks,
presents scalability challenges both when agents need to adapt to multiple tasks within the same environments and when the same task must be tackled across environment variations.

\begin{figure}[ht]
    \centering
    \includegraphics[width=.25\textwidth]{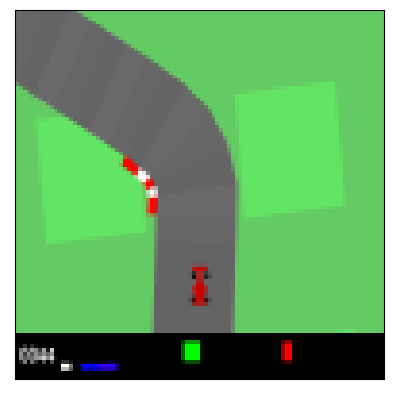}
    \includegraphics[width=.25\textwidth]{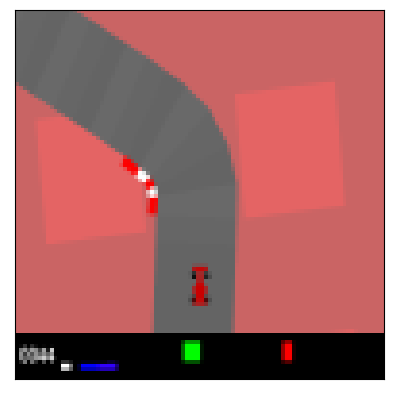}
    \caption[Environment variations in Car Racing]{The modified version of the Car Racing environment, where we can change the color of the background.}
    \label{rl:carracingenv}
\end{figure}

An example of this challenge is illustrated in \Cref{rl:carracingenv}, where we would like to train a policy that \emph{drives the car} in one environment, and reuse it across variations of that environment without retraining.
Ideally, a model trained on a specific task would maintain its policy while being able to substitute its encoder for another, thereby facilitating adaptation to environmental changes -- such as varying weather conditions -- without the need for retraining the policy.
Driven by this insight, our investigation fully described in \cite{ricciardi2023}, and similar concurrent work \citep{jian2023}, leverage \glsxtrshortpl{rr} (\Cref{chap:relative}) to prove the feasibility of Zero-Shot Stitching  (\Cref{sec:stitchingdefinition}) between encoders and policies trained on different environmental variations.

\begin{table}[ht]
    \caption[Episode maximum return comparing in stitching]{Episode maximum return comparing in stitching. Encoder (rows) and policy (columns) colors represent the track background on which that module was trained on.}
    \label{rl:stitchingquantitative}
    \centering
    \resizebox{\textwidth}{!}{
        \begin{tabular}{cccccccccc}
            \toprule
                                                              &                 &                                    &                                  &                                   & \textbf{Policy 	↑}                  &               &               &               &              \\
            \cmidrule{3-4} \cmidrule{5-6} \cmidrule{7-8} \cmidrule{9-10}
                                                              &                 & \multicolumn{2}{c}{\texttt{green}} & \multicolumn{2}{c}{\texttt{red}} & \multicolumn{2}{c}{\texttt{blue}} & \multicolumn{2}{c}{\texttt{yellow}}                                                                \\
            \cmidrule{3-4} \cmidrule{5-6} \cmidrule{7-8} \cmidrule{9-10}
                                                              &                 & \textit{Rel}                       & \textit{Abs}                     & \textit{Rel}                      & \textit{Abs}                        & \textit{Rel}  & \textit{Abs}  & \textit{Rel}  & \textit{Abs} \\
            \midrule
            \multirow{4}{*}{\rotatebox{90}{\textbf{Encoder}}} & \texttt{green}  & 714 $\pm$ 288                      & 863 $\pm$ 109                    & 840 $\pm$ 94                      & 7 $\pm$ 0.7                         & 870 $\pm$ 84  & 26 $\pm$ 4    & 685 $\pm$ 237 & 12 $\pm$ 4   \\
                                                              & \texttt{red}    & 774 $\pm$ 275                      & 19 $\pm$ 3.8                     & 692 $\pm$ 251                     & 829 $\pm$ 116                       & 288 $\pm$ 105 & 7 $\pm$ 0.7   & 638 $\pm$ 235 & 7 $\pm$ 0.7  \\
                                                              & \texttt{blue}   & 256 $\pm$ 163                      & 7 $\pm$ 1                        & 307 $\pm$ 124                     & 7 $\pm$ 0.6                         & 690 $\pm$ 200 & 759 $\pm$ 288 & 556 $\pm$ 71  & 8 $\pm$ 3    \\
                                                              & \texttt{yellow} & 713 $\pm$ 204                      & 7 $\pm$ 0.7                      & 678 $\pm$ 81                      & 33 $\pm$ 4                          & 167 $\pm$ 66  & 26 $\pm$ 4    & 675 $\pm$ 233 & 874 $\pm$ 85 \\
            \bottomrule
        \end{tabular}
    }
\end{table}

The preliminary exploration was carried out within a modified version of the CarRacing environment, featuring a discrete action space and the capability to alter the background color. We trained end-to-end agents, employing a conventional \gls{cnn} as feature extractor, paired with a policy module.
The empirical results, reported in \Cref{rl:stitchingquantitative}, show that by employing \glsxtrshortpl{rr},  a policy trained in conjunction with an encoder under a specific background color setting can be effortlessly used in a different background scenario with the corresponding encoder; resulting in minimal to no degradation in performance.

\part{Conclusions}

\chapter{Conclusions}\label{chap:conclusions}

In this dissertation, we introduced the \emph{\glsfirst{lcp}} framework, as formalized in \Cref{chap:problemformalization} and illustrated in \Cref{fig:formalization}. It is a novel, unifying approach that recognizes the presence of unobservable abstract manifolds, representing the underlying semantics of data. These manifolds become observable when embedded in high-dimensional spaces, such as images, texts, or latent spaces. The objective of the \gls{lcp} is to identify two specific transformations, $\gls{Tx}$ and $\gls{Ty}$, that modify the entire latent spaces to align the manifolds within them. This framework has allowed us to \emph{reinterpret several of our recent works}, \cite{moschella2023, maiorca2023, cannistraci2023, cannistraci2023a, norelli2022, crisostomi2023from, ricciardi2023}, from a new, unifying perspective.

Throughout this manuscript, we have showcased these methodologies to tackle the challenges presented by the \gls{lcp}, building upon the foundational concepts introduced in the initial problem formalization.  Our research spans from the exploration of \emph{universal representations} (\Cref{chap:relative}) and \emph{direct translation} (\Cref{chap:translation}), as well as overcoming inherent methodological limitations.
These efforts include eliminating the need to know the specific transformation class relating different spaces (\Cref{chap:bridge}) and dealing with the limited available semantic correspondence between data domains (\Cref{chap:bootstrapping}).

Indeed, beyond theoretical considerations, solving the \gls{lcp} offers tangible benefits, as described in \Cref{chap:corollaryproblems}.
One of the most salient outcomes is the concept of model compositionality through \emph{Zero-Shot Model Stitching} (\Cref{sec:stitchingdefinition}).
This innovative methodology ensures that neural architectures can function as modular units, facilitating their reuse without the necessity for extensive retraining or fine-tuning.
Furthermore, \gls{lcp} solutions allow the direct comparisons between independently obtained latent spaces.
Therefore, when an appropriate reference model is available, they provide a quantitative \emph{latent measure of performance} (\Cref{sec:latentcomparisondefinition}), which is often differentiable, and is correlated with standard performance measures such as accuracy on downstream tasks.
Finally, it supports the development of \emph{advanced retrieval systems} (\Cref{sec:retrievaldefinition}) that leverage independently computed representations. This enables the retrieval of data points from one space using queries from another, eliminating the need for a shared training dataset.
We illustrate these advantages with three detailed \emph{case studies} in \Cref{chap:casestudies}, covering diverse fields, including \glsxtrlong{cv}, \glsxtrlong{nlp} and \glsxtrlong{rl}.

This dissertation concludes with a brief overview of its impact on the broader field in \Cref{chap:contributions}, along with a discussion of the limitations, future research directions, and opportunities introduced by our works, detailed in \Cref{chap:future}.

\chapter{Contributions to the field}
\label{chap:contributions}

This Section delineates our contributions  to the broader field, underscoring its adoption and impact across a diverse array of academic venues. The UniReps: Unifying Representations in Neural Models workshop at NeurIPS 2023, which our team co-organized, attests the significance and timeliness of our work (\Cref{sec:unireps}).  Moreover, our research has been recognized and further developed in numerous preprints, peer-reviewed journals and articles presented at leading conferences (\Cref{sec:otherworks}).

\section{UniReps Workshop: Unifying Representations in Neural Models} \label{sec:unireps}
The concept of \emph{Latent Communication} has been a central theme at the NeurIPS 2023 workshop titled ``UniReps: Unifying Representations in Neural Models'', which our team co-organized, and where I had the honor of serving as a Program Chair.

The workshop's mission  focused on core questions about when, how, and why different neural models converge on similar representations. This phenomenon has piqued the interest of researchers across Neuroscience, Artificial Intelligence, and Cognitive Science, indicating a thriving interdisciplinary field of study. It focused on three main themes:
(i)~\emph{When.} Understanding the patterns by which these similarities emerge in different neural models and developing methods to measure them;
(ii)~\emph{Why.} Investigating the underlying causes of these similarities in neural representations, considering both artificial and biological models;
(iii)~\emph{What for.} Exploring and showcasing applications in modular deep learning, including model merging, reuse, stitching, efficient strategies for fine-tuning, and knowledge transfer between models and across modalities.

The workshop's success is underscored by its substantial engagement and outcomes, demonstrating widespread interest in \emph{Latent Communication}. It attracted 800+ attendees and received 90+ submissions, supported by a program committee of 150+ experts. Featuring invited talks from leading researchers in both industry (such as DeepMind, Anthropic) and academia (including UCSB, Princeton), the workshop was also sponsored by notable entities like Google DeepMind, Gatsby, and Translated.

\section{Works by other researchers} \label{sec:otherworks}

This Section is dedicated to showcasing the influence and impact of our methodologies, briefly describing how they have been adopted, adapted, and extended by other researchers.

\paragraph{State of the art in \glsxtrshort{wvlp}.}
\cite{chen2023}
``proposes a \emph{relative representation}-based
\gls{wvlp} framework that can both retrieve and
generate weakly aligned image-text pairs for
learning cross-modal representations''
that
``outperforms strong \glsxtrlong{wvlp} baselines and further closes the performance gap
between \glsxtrlong{wvlp} and standard VLP''.

\paragraph{Generalizing Task Semantics Across Language
    Models.}
\cite{wu2023}
``addresses a novel setting of zero-shot continuous prompt transfer, which allows for the
reuse of continuous prompts across different language models'';
suggesting ``an encode-then-search strategy that maps a continuous prompt into a \emph{relative
    space} for transfer between language models''.

\paragraph{Understanding Shared Speech-Text Representations.}
\cite{wang2023a} employs relative representations to reveal that the shared encoder learns a more compact and overlapping speech-text representation than the uni-modal encoders.

\paragraph{Policy Stitching: Learning Transferable Robot Policies.}
\cite{jian2023} generalizes relative representations to enable
``Policy Stitching, a model-free reinforcement learning framework for robot transfer learning
among novel robot and task combinations''
demonstrating clear advantages
``in both zero-shot and few-shot transfer
learning through simulated and physical 3D manipulation tasks''.

\paragraph{Relative Representations for Cognitive Graphs.}
\cite{kiefer2023a} extends
``\emph{relative representations} to discrete state-space models, using Clone-Structured Cognitive Graphs (CSCGs)'';
showing that
``the probability vectors computed during message passing can be used to define relative representations on CSCGs''
enabling effective Latent Communication across agents trained in different settings.

\paragraph{Model Stitching with Static Word Embeddings.}
\cite{ye2024} %
introduces ``MoSECroT, a novel and challenging task for (especially low-resource) languages
where static word embeddings are available'',
and proposes ``a solution that leverages \emph{relative representations} to construct a common space for source and target languages and that allows zero-shot transfer for the target languages''.

\paragraph{Knowledge Distillation with Relative Representations.}
\cite{ramos2023} %
designs
``a knowledge distillation scheme centered around matching the relative representations of a student to those of a teacher'' and show that the proposed method ``outperforms similar relation-based distillation approaches across a variety of benchmarks, with results extending to transfer learning''.

\paragraph{Direct Alignment of Latent Spaces.}
\cite{lahner2023on} %
proposes, concurrently to the work presented in \Cref{chap:translation},
``the theory that semantically related latent spaces even of very different
network architectures are related by a linear transformation''
and demonstrates that
``aligning the latent space with a linear transformation performs best while
not needing more prior knowledge''.

\paragraph{Boosting Visual-Language Models.}
\cite{wang2023b} %
employs relative representations to
``propose a novel hard sample selection technique
for the identification of hard negative samples''
and
``consistently improve CLIP
model checkpoints by finetuning''.

\chapter{Limitations and Future Directions}\label{chap:future}
In this Section, we summarize some open questions and potential avenues for future research based on the contributions presented in this dissertation. Our work has laid a strong foundation in the \gls{lcp} framework, opening several pathways for further exploration and development. In the following paragraphs, we outline some of the most promising directions.

\paragraph{Modular Neural Components.}
Our exploration of \emph{Latent Communication} paves the way for leveraging neural models in a modular, compositional fashion, potentially circumventing the exhaustive fine-tuning or retraining currently prevalent.
Yet, bridging the performance gap across training modalities -- especially between zero-shot and fine-tuning approaches -- remains an elusive challenge.
This disparity is particularly pronounced in industrial ML applications, where performance maximization often comes at the expense of computational efficiency. Our Zero-Shot Stitching methodology offers a partial solution; however, the quest for models that adapt dynamically to changes in feature representation with minimal retraining persists. Future research might focus on \emph{self-adjusting mechanisms} akin to Test-Time Training \citep{sun2020,wang2020}, integrated with Latent Communication strategies, to address this gap.

\paragraph{Latent Communication without Semantic Correspondence.}
The dependency on initial seed anchors for parallel domains limits the current scope of Latent Communication. Removing this constraint could involve developing {unsupervised methods for identifying parallel anchors} or {methods for learning the anchors from the data}.
The Gromov-Wasserstein distance \citep{memoli2011} presents a promising theoretical underpinning for such methods, e.g., potentially revolutionizing cross-domain retrieval systems by eliminating the need for parallel data.

\paragraph{Latent Communication on sequences.}
Currently, we have devised solutions for the \gls{lcp} when samples can be represented as a single embedding, i.e., a point in a high-dimensional space. An interesting direction could be to solve the \gls{lcp} natively on sequences, e.g., by considering the latent space of a sequence of embeddings. This would allow a more natural handling of sequential data, such as textual embeddings, without the need to aggregate all the token embeddings into a single one (e.g., considering the final token, the CLS or the mean of the token embeddings). However, this would require the use of a similarity function that can compare embedding sequences, or the development of novel methods to achieve \gls{lcp} in this context.

\paragraph{\glsxtrlongpl{nn} Inspection.}
Solutions to the \gls{lcp} could be used to analyze the latent spaces of \glspl{nn}, e.g., providing  insights into the evolution of the representations during training. They could be an interesting tool to understand when and how \glspl{nn} develop or refine their ability to represent information. Indeed, it is yet to be understood how the latent spaces evolve during training, and there are a variety of phenomenon (e.g., emergent abilities \citep{wei2022}, neural collapse \citep{papyan2020}, double descent \citep{nakkiran2019}) that could be investigated exploiting the \gls{lcp} framework and a known reference model, inspired by \Cref{sec:latentcomparisondefinition}.

\paragraph{Partial Latent Communication.}
Requiring a full alignment of the manifold embeddings might be too restrictive in some scenarios, e.g., in cases where the two data distributions are only partially semantically overlapping. In these cases, it would be interesting to develop methods for \emph{Partial Latent Communication} to align only a subset of the data manifolds. Similar techniques could be used to ensure the best alignment for a particular subset of the data of interest, such as particular classes or categories.

\paragraph{Representation Interpretability.}
The framework of \glspl{rr} offers a novel lens for representation interpretability, associating specific meanings with each dimension through the anchor semantic. This could be further exploited by more tailored similarity functions, e.g., by performing a change of basis to obtain a more interpretable \gls{rr}. This would allow interpreting the dimensions of the representation as the directions in the data space associated with specific semantic concepts, defined by the anchor, thus providing a more intuitive understanding of the latent space.

\paragraph{Learnable Similarity Functions.}
The framework described in \Cref{chap:relative} allows incorporating invariances into the latent representation, exploiting specific similarity functions. Moreover, in \Cref{chap:bridge} we have shown how to extend it to infuse a set of invariances, instead of a single one. Nevertheless, this can still be limiting when the similarity function that induces an invariance to \gls{transformationclass} cannot be modeled analytically or expressed in closed form. In such cases, an interesting direction would be to \emph{learn} the desired similarity function \gls{sim}.

\paragraph{Geodesic \glsxtrlongpl{rr}.}
Another fascinating line of research to improve the representation expressivity would be to estimate \emph{geodesic} distances over the data manifold, instead of adopting distances in the ambient spaces. This could allow defining \glspl{rr} that better capture the intrinsic structure of the data, especially when the manifold embedding is complex. However, this would require the development of innovative methods for efficient geodesic estimation in high-dimensional spaces.

\paragraph{Higher Order \glsxtrlongpl{rr}.}
The \gls{rr} framework is currently limited to employ pairwise similarity functions, i.e., it can only capture first-order relationships between data points. In practice, this means that there can be only one anchor associated with each dimension.
Extending it to higher-order n-way relationships, e.g., by considering triplets or quadruplets, could allow capturing more complex relationships between data points, and thus provide a more expressive representation.

\paragraph{Anchors selection methods.}
The interplay between anchor composition and the expressiveness of \glspl{rr} warrants further investigation. Questions surrounding optimal anchor selection and the necessary number of anchors remain mostly unanswered. Developing methodologies for selecting anchors -- guided by considerations of data distribution, transformation classes, or specific tasks -- stands as a significant frontier for future research.

\paragraph{Weights Similarity.}
Throughout this dissertation, we have explored the emerging similarities in the latent spaces of \glspl{nn}.
At the same time, a growing body of research has focused on emerging similarities between networks in the weight space, and how to exploit them
\citep{
    ilharco2023,
    ortiz-jimenez2023,
    ainsworthGitReBasinMerging2023,
    wortsman2022,
    singhModelFusionOptimal2020,
    tatroOptimizingModeConnectivity2020,
    entezari2022,
    matenaMergingModelsFisherWeighted2022,
    rameModelRatatouilleRecycling2023,
    frankle2020}.
An exciting research direction would be to investigate the relationship between the latent spaces and the weights of the models, answering the following question: ``Are \glspl{nn} with similar latent spaces also similar in the weight space?''

\paragraph{Automatic Data Curation.}
Data-centric AI and automatic data curation are experiencing rapid growth.
This evolution underscores a realization: the “quantity” is not the sole determinant of AI performance.
Rather, the “quality” of data plays a crucial, if not more significant, role in enhancing the training processes, boosting model performance, and optimizing model size.
Within this context, the \gls{lcp} methodology emerges as a powerful tool.
It introduces a paradigm shift by employing trained models not just for predictions, but as a means to critically assess and ensure the quality of datasets -- for instance, automating the process of identifying and eliminating noisy data alignments.
This direction holds considerable promise for future research, particularly  when dealing with multimodal data, where aligning diverse data modalities presents considerable challenges.

\part[Appendices]{Appendices}

\appendix %

\chapter{Universal Representations}
\section{Anchors analysis}\label{relative:sec:anchor-num}
The cardinality of the anchors set ${\gls{anchors}}$ and the choice of specific anchors is crucial to the quality of the relative representations. At the extreme, selecting one single anchor or the same repeated data points for all anchors, will produce collapsed relative representations. We believe that additional research is required to obtain a better understanding on the optimal choice for ${\gls{anchors}}$. Questions like ``Are anchors set composed only by stopwords worse than the ones composed by meaningful and diverse words?'' require empirical evidence and could help revealing the semantics of the latent space. Indeed, each anchor is associated with a dimension in a relative representation; one could inspect the anchor data point to get a sense of the meaning of that latent dimension.

\begin{figure}[ht]
   \centering
   \begin{overpic}[trim=-.5cm -.4cm -.5cm 0cm,clip,width=0.9\linewidth]{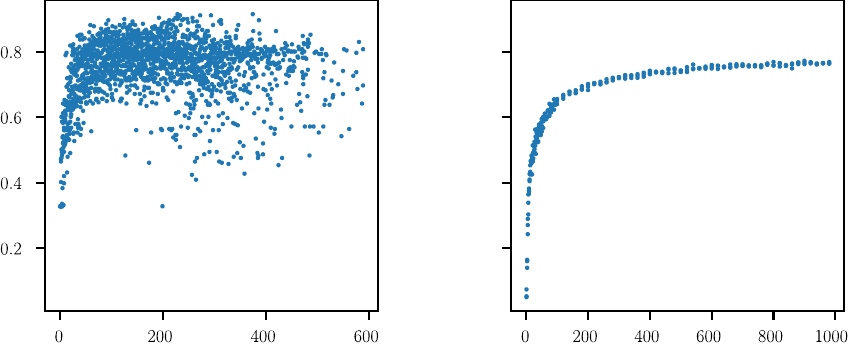}
      \put(0, 15){\rotatebox{90}{Performance}}
      \put(17, 0){Number of anchors}
      \put(70, 0){Number of anchors}
   \end{overpic}
   \caption[Accuracy vs Number of anchors]{Accuracy vs Number of anchors. Each point is a trained model. \textit{Left}: Trained embedder on \gls{cora}, node classification. \textit{Right}: Frozen transformer on \gls{cifarh} coarse-grained, image classification. Left is less stable because the absolute embeddings are trained, and we are working on a domain that is less stable (graphs). Some collapsed examples are not visualized.}
   \label{relative:fig:anchor-num}
\end{figure}

\paragraph{Anchor number.}
Below, we report a preliminary study on the performance sensitivity against the cardinality of the anchors set. In \Cref{relative:fig:anchor-num} we report the performance on the node classification task on \gls{cora}, with a model trained end-to-end adopting the relative representations while training, and on image classification tasks on \gls{cifarh}, with a frozen encoder.
The performance improves monotonically as the number of anchors increase when the absolute representations are frozen \emph{(right)}. Differently, training models end-to-end proves to be more susceptible to model collapse and instabilities, as increasing the number of anchors does not always improve the performance \emph{(left)}. Further research on the relation between the absolute latent space dimensionality and the relative representation dimensionality (i.e., the number of anchors) is needed to clarify how the two quantities impact the performance, when training end-to-end or not.

\paragraph{Anchor selection.}\label{relative:sec:anchors-selection}
\begin{sloppypar}
   In \Cref{relative:tab:quantitative-analysis-word-embeddings-all,relative:tab:quantitative-analysis-cv-all}, we analyze different anchor selection strategies under an experimental setting analogous to the one described in \Cref{relative:sec:word-embeddings}:
\end{sloppypar}
\begin{itemize}
   \item \textbf{uniform} The first selection strategy is the one adopted in \Cref{chap:relative}. We randomly select the anchors with a uniform probability distribution over all the available samples.
   \item \textbf{fps} We select the anchors according to a farthest point sampling strategy.
   \item \textbf{kmeans} We select the anchors as the words more close to the centroids of K-means clustering with $K = \text{number of anchors}$.
   \item \textbf{top\{$k$\}} We select the anchors as the $k$ most frequent words, after skipping the first 400 which are mostly stopwords.
\end{itemize}

{We expect strategies that better cover the absolute space with anchors to be the most effective ones.
Indeed, the results are comparable across selection strategies, but fps reaches everywhere the best Jaccard and MRR scores while k-means the best Cosine ones. We attribute this behavior to their different nature: they both rely on the geometry of the latent spaces they are applied to, but k-means also favors high-density regions, and this can become a negative bias for the task at hand. In general, the uniform sampling is the most straightforward to apply, since it does not require additional computation for the selection process, and still achieves good performances.}

\begin{table}[ht]
   \centering
   \footnotesize
   \caption[Additional results with different anchor selection strategies]{{Extended results from \Cref{relative:sec:word-embeddings} with different anchor selection strategies. The table reports the mean score for each metric and its std across 10 different seeds.}}
   \label{relative:tab:quantitative-analysis-word-embeddings-all}
   \begin{tabular}{lllllll}
      \toprule
      \textbf{Mode}                                                         & \textbf{Type}                                                & \textbf{Source}                   & \textbf{Target}  & \multicolumn{1}{c}{\textbf{Jaccard ↑}} & \multicolumn{1}{c}{\textbf{MRR ↑}} & \multicolumn{1}{c}{\textbf{Cosine ↑}} \\
      \midrule
      \multirow{8}{*}{{\STAB{\rotatebox[origin=c]{90}{\textbf{uniform}}}}}  & \multirow{4}{*}{{\STAB{\rotatebox[origin=c]{90}{Absolute}}}} & \multirow{2}{*}{{\gls{fasttext}}} & {\gls{fasttext}} & $1.00 \pm 0.00$                        & $1.00 \pm 0.00$                    & $1.00 \pm 0.00$                       \\
                                                                            &                                                              &                                   & {\gls{word2vec}} & $0.00 \pm 0.00$                        & $0.00 \pm 0.00$                    & $0.01 \pm 0.00$                       \\[0.5ex]
                                                                            &                                                              & \multirow{2}{*}{{\gls{word2vec}}} & {\gls{fasttext}} & $0.00 \pm 0.00$                        & $0.00 \pm 0.00$                    & $0.01 \pm 0.00$                       \\
                                                                            &                                                              &                                   & {\gls{word2vec}} & $1.00 \pm 0.00$                        & $1.00 \pm 0.00$                    & $1.00 \pm 0.00$                       \\
      \cmidrule{2-7}
                                                                            & \multirow{4}{*}{\STAB{\rotatebox[origin=c]{90}{Relative}}}   & \multirow{2}{*}{{\gls{fasttext}}} & {\gls{fasttext}} & $1.00 \pm 0.00$                        & $1.00 \pm 0.00$                    & $1.00 \pm 0.00$                       \\
                                                                            &                                                              &                                   & {\gls{word2vec}} & $0.34 \pm 0.01$                        & $0.94 \pm 0.00$                    & $0.86 \pm 0.00$                       \\[0.5ex]
                                                                            &                                                              & \multirow{2}{*}{{\gls{word2vec}}} & {\gls{fasttext}} & $0.39 \pm 0.00$                        & $0.98 \pm 0.00$                    & $0.86 \pm 0.00$                       \\
                                                                            &                                                              &                                   & {\gls{word2vec}} & $1.00 \pm 0.00$                        & $1.00 \pm 0.00$                    & $1.00 \pm 0.00$                       \\
      \cmidrule{1-7}
      \multirow{8}{*}{{\STAB{\rotatebox[origin=c]{90}{\textbf{fps}}}}}      & \multirow{4}{*}{{\STAB{\rotatebox[origin=c]{90}{Absolute}}}} & \multirow{2}{*}{{\gls{fasttext}}} & {\gls{fasttext}} & $1.00 \pm 0.00$                        & $1.00 \pm 0.00$                    & $1.00 \pm 0.00$                       \\
                                                                            &                                                              &                                   & {\gls{word2vec}} & $0.00 \pm 0.00$                        & $0.00 \pm 0.00$                    & $0.01 \pm 0.00$                       \\[0.5ex]
                                                                            &                                                              & \multirow{2}{*}{{\gls{word2vec}}} & {\gls{fasttext}} & $0.00 \pm 0.00$                        & $0.00 \pm 0.00$                    & $0.01 \pm 0.00$                       \\
                                                                            &                                                              &                                   & {\gls{word2vec}} & $1.00 \pm 0.00$                        & $1.00 \pm 0.00$                    & $1.00 \pm 0.00$                       \\
      \cmidrule{2-7}
                                                                            & \multirow{4}{*}{\STAB{\rotatebox[origin=c]{90}{Relative}}}   & \multirow{2}{*}{{\gls{fasttext}}} & {\gls{fasttext}} & $1.00 \pm 0.00$                        & $1.00 \pm 0.00$                    & $1.00 \pm 0.00$                       \\
                                                                            &                                                              &                                   & {\gls{word2vec}} & $0.34 \pm 0.01$                        & $0.94 \pm 0.00$                    & $0.81 \pm 0.00$                       \\[0.5ex]
                                                                            &                                                              & \multirow{2}{*}{{\gls{word2vec}}} & {\gls{fasttext}} & $0.41 \pm 0.00$                        & $0.98 \pm 0.00$                    & $0.83 \pm 0.00$                       \\
                                                                            &                                                              &                                   & {\gls{word2vec}} & $1.00 \pm 0.00$                        & $1.00 \pm 0.00$                    & $1.00 \pm 0.00$                       \\
      \cmidrule{1-7}
      \multirow{8}{*}{{\STAB{\rotatebox[origin=c]{90}{\textbf{kmeans}}}}}   & \multirow{4}{*}{{\STAB{\rotatebox[origin=c]{90}{Absolute}}}} & \multirow{2}{*}{{\gls{fasttext}}} & {\gls{fasttext}} & $1.00 \pm 0.00$                        & $1.00 \pm 0.00$                    & $1.00 \pm 0.00$                       \\
                                                                            &                                                              &                                   & {\gls{word2vec}} & $0.00 \pm 0.00$                        & $0.00 \pm 0.00$                    & $0.01 \pm 0.00$                       \\[0.5ex]
                                                                            &                                                              & \multirow{2}{*}{{\gls{word2vec}}} & {\gls{fasttext}} & $0.00 \pm 0.00$                        & $0.00 \pm 0.00$                    & $0.01 \pm 0.00$                       \\
                                                                            &                                                              &                                   & {\gls{word2vec}} & $1.00 \pm 0.00$                        & $1.00 \pm 0.00$                    & $1.00 \pm 0.00$                       \\
      \cmidrule{2-7}
                                                                            & \multirow{4}{*}{\STAB{\rotatebox[origin=c]{90}{Relative}}}   & \multirow{2}{*}{{\gls{fasttext}}} & {\gls{fasttext}} & $1.00 \pm 0.00$                        & $1.00 \pm 0.00$                    & $1.00 \pm 0.00$                       \\
                                                                            &                                                              &                                   & {\gls{word2vec}} & $0.35 \pm 0.00$                        & $0.94 \pm 0.00$                    & $0.87 \pm 0.00$                       \\[0.5ex]
                                                                            &                                                              & \multirow{2}{*}{{\gls{word2vec}}} & {\gls{fasttext}} & $0.39 \pm 0.00$                        & $0.97 \pm 0.00$                    & $0.87 \pm 0.00$                       \\
                                                                            &                                                              &                                   & {\gls{word2vec}} & $1.00 \pm 0.00$                        & $1.00 \pm 0.00$                    & $1.00 \pm 0.00$                       \\
      \cmidrule{1-7}
      \multirow{8}{*}{{\STAB{\rotatebox[origin=c]{90}{\textbf{top1000}}}}}  & \multirow{4}{*}{{\STAB{\rotatebox[origin=c]{90}{Absolute}}}} & \multirow{2}{*}{{\gls{fasttext}}} & {\gls{fasttext}} & $1.00 \pm 0.00$                        & $1.00 \pm 0.00$                    & $1.00 \pm 0.00$                       \\
                                                                            &                                                              &                                   & {\gls{word2vec}} & $0.00 \pm 0.00$                        & $0.00 \pm 0.00$                    & $0.01 \pm 0.00$                       \\[0.5ex]
                                                                            &                                                              & \multirow{2}{*}{{\gls{word2vec}}} & {\gls{fasttext}} & $0.00 \pm 0.00$                        & $0.00 \pm 0.00$                    & $0.01 \pm 0.00$                       \\
                                                                            &                                                              &                                   & {\gls{word2vec}} & $1.00 \pm 0.00$                        & $1.00 \pm 0.00$                    & $1.00 \pm 0.00$                       \\
      \cmidrule{2-7}
                                                                            & \multirow{4}{*}{\STAB{\rotatebox[origin=c]{90}{Relative}}}   & \multirow{2}{*}{{\gls{fasttext}}} & {\gls{fasttext}} & $1.00 \pm 0.00$                        & $1.00 \pm 0.00$                    & $1.00 \pm 0.00$                       \\
                                                                            &                                                              &                                   & {\gls{word2vec}} & $0.27 \pm 0.01$                        & $0.84 \pm 0.01$                    & $0.85 \pm 0.00$                       \\[0.5ex]
                                                                            &                                                              & \multirow{2}{*}{{\gls{word2vec}}} & {\gls{fasttext}} & $0.35 \pm 0.01$                        & $0.97 \pm 0.00$                    & $0.85 \pm 0.00$                       \\
                                                                            &                                                              &                                   & {\gls{word2vec}} & $1.00 \pm 0.00$                        & $1.00 \pm 0.00$                    & $1.00 \pm 0.00$                       \\
      \cmidrule{1-7}
      \multirow{8}{*}{{\STAB{\rotatebox[origin=c]{90}{\textbf{top5000}}}}}  & \multirow{4}{*}{{\STAB{\rotatebox[origin=c]{90}{Absolute}}}} & \multirow{2}{*}{{\gls{fasttext}}} & {\gls{fasttext}} & $1.00 \pm 0.00$                        & $1.00 \pm 0.00$                    & $1.00 \pm 0.00$                       \\
                                                                            &                                                              &                                   & {\gls{word2vec}} & $0.00 \pm 0.00$                        & $0.00 \pm 0.00$                    & $0.01 \pm 0.00$                       \\[0.5ex]
                                                                            &                                                              & \multirow{2}{*}{{\gls{word2vec}}} & {\gls{fasttext}} & $0.00 \pm 0.00$                        & $0.00 \pm 0.00$                    & $0.01 \pm 0.00$                       \\
                                                                            &                                                              &                                   & {\gls{word2vec}} & $1.00 \pm 0.00$                        & $1.00 \pm 0.00$                    & $1.00 \pm 0.00$                       \\
      \cmidrule{2-7}
                                                                            & \multirow{4}{*}{\STAB{\rotatebox[origin=c]{90}{Relative}}}   & \multirow{2}{*}{{\gls{fasttext}}} & {\gls{fasttext}} & $1.00 \pm 0.00$                        & $1.00 \pm 0.00$                    & $1.00 \pm 0.00$                       \\
                                                                            &                                                              &                                   & {\gls{word2vec}} & $0.32 \pm 0.00$                        & $0.92 \pm 0.00$                    & $0.86 \pm 0.00$                       \\[0.5ex]
                                                                            &                                                              & \multirow{2}{*}{{\gls{word2vec}}} & {\gls{fasttext}} & $0.38 \pm 0.00$                        & $0.97 \pm 0.00$                    & $0.86 \pm 0.00$                       \\
                                                                            &                                                              &                                   & {\gls{word2vec}} & $1.00 \pm 0.00$                        & $1.00 \pm 0.00$                    & $1.00 \pm 0.00$                       \\
      \cmidrule{1-7}
      \multirow{8}{*}{{\STAB{\rotatebox[origin=c]{90}{\textbf{top10000}}}}} & \multirow{4}{*}{{\STAB{\rotatebox[origin=c]{90}{Absolute}}}} & \multirow{2}{*}{{\gls{fasttext}}} & {\gls{fasttext}} & $1.00 \pm 0.00$                        & $1.00 \pm 0.00$                    & $1.00 \pm 0.00$                       \\
                                                                            &                                                              &                                   & {\gls{word2vec}} & $0.00 \pm 0.00$                        & $0.00 \pm 0.00$                    & $0.01 \pm 0.00$                       \\[0.5ex]
                                                                            &                                                              & \multirow{2}{*}{{\gls{word2vec}}} & {\gls{fasttext}} & $0.00 \pm 0.00$                        & $0.00 \pm 0.00$                    & $0.01 \pm 0.00$                       \\
                                                                            &                                                              &                                   & {\gls{word2vec}} & $1.00 \pm 0.00$                        & $1.00 \pm 0.00$                    & $1.00 \pm 0.00$                       \\
      \cmidrule{2-7}
                                                                            & \multirow{4}{*}{\STAB{\rotatebox[origin=c]{90}{Relative}}}   & \multirow{2}{*}{{\gls{fasttext}}} & {\gls{fasttext}} & $1.00 \pm 0.00$                        & $1.00 \pm 0.00$                    & $1.00 \pm 0.00$                       \\
                                                                            &                                                              &                                   & {\gls{word2vec}} & $0.34 \pm 0.00$                        & $0.93 \pm 0.00$                    & $0.86 \pm 0.00$                       \\[0.5ex]
                                                                            &                                                              & \multirow{2}{*}{{\gls{word2vec}}} & {\gls{fasttext}} & $0.39 \pm 0.01$                        & $0.97 \pm 0.00$                    & $0.86 \pm 0.00$                       \\
                                                                            &                                                              &                                   & {\gls{word2vec}} & $1.00 \pm 0.00$                        & $1.00 \pm 0.00$                    & $1.00 \pm 0.00$                       \\
      \bottomrule
   \end{tabular}
\end{table}

\begin{table}[ht]
   \small
   \centering
   \caption[Generalization of \Cref{relative:sec:word-embeddings} to a different data modality]{{Generalization of the results from \Cref{relative:sec:word-embeddings} on word embeddings to a different data modality, with different anchor selection strategies (See \Cref{relative:sec:anchors-selection} for their description). The dataset considered is \gls{cifart}, and the table reports the mean score for each metric and its std across 10 different seeds.}}
   \label{relative:tab:quantitative-analysis-cv-all}
   \begin{tabular}{lllllll}
      \toprule
      \multicolumn{1}{c}{\textbf{Mode}}                                    & \multicolumn{1}{c}{\textbf{Type}}                            & \multicolumn{1}{c}{\textbf{Source}}         & \multicolumn{1}{c}{\textbf{Target}} & \multicolumn{1}{c}{\textbf{Jaccard ↑}} & \multicolumn{1}{l}{\textbf{MRR ↑}} & \multicolumn{1}{c}{\textbf{Cosine ↑}} \\
      \midrule
      \multirow{8}{*}{{\STAB{\rotatebox[origin=c]{90}{\textbf{uniform}}}}} & \multirow{4}{*}{{\STAB{\rotatebox[origin=c]{90}{Absolute}}}} & \multirow{2}{*}{{\glsxtrshort{vitbp16224}}} & {\glsxtrshort{vitbp16224}}          & $1.00 \pm 0.00$                        & $1.00 \pm 0.00$                    & $1.00 \pm 0.00$                       \\
                                                                           &                                                              &                                             & {\glsxtrshort{vitsp16224}}          & \multicolumn{1}{c}{-}                  & \multicolumn{1}{c}{-}              & \multicolumn{1}{c}{-}                 \\[0.5ex]
                                                                           &                                                              & \multirow{2}{*}{{\glsxtrshort{vitsp16224}}} & {\glsxtrshort{vitbp16224}}          & \multicolumn{1}{c}{-}                  & \multicolumn{1}{c}{-}              & \multicolumn{1}{c}{-}                 \\
                                                                           &                                                              &                                             & {\glsxtrshort{vitsp16224}}          & $1.00 \pm 0.00$                        & $1.00 \pm 0.00$                    & $1.00 \pm 0.00$                       \\
      \cmidrule{2-7}
                                                                           & \multirow{4}{*}{{\STAB{\rotatebox[origin=c]{90}{Relative}}}} & \multirow{2}{*}{{\glsxtrshort{vitbp16224}}} & {\glsxtrshort{vitbp16224}}          & $1.00 \pm 0.00$                        & $1.00 \pm 0.00$                    & $1.00 \pm 0.00$                       \\
                                                                           &                                                              &                                             & {\glsxtrshort{vitsp16224}}          & $0.11 \pm 0.00$                        & $0.27 \pm 0.01$                    & $0.97 \pm 0.00$                       \\[0.5ex]
                                                                           &                                                              & \multirow{2}{*}{{\glsxtrshort{vitsp16224}}} & {\glsxtrshort{vitbp16224}}          & $0.11 \pm 0.00$                        & $0.30 \pm 0.01$                    & $0.97 \pm 0.00$                       \\
                                                                           &                                                              &                                             & {\glsxtrshort{vitsp16224}}          & $1.00 \pm 0.00$                        & $1.00 \pm 0.00$                    & $1.00 \pm 0.00$                       \\
      \cmidrule{1-7}
      \multirow{8}{*}{{\STAB{\rotatebox[origin=c]{90}{\textbf{fps}}}}}     & \multirow{4}{*}{{\STAB{\rotatebox[origin=c]{90}{Absolute}}}} & \multirow{2}{*}{{\glsxtrshort{vitbp16224}}} & {\glsxtrshort{vitbp16224}}          & $1.00 \pm 0.00$                        & $1.00 \pm 0.00$                    & $1.00 \pm 0.00$                       \\
                                                                           &                                                              &                                             & {\glsxtrshort{vitsp16224}}          & \multicolumn{1}{c}{-}                  & \multicolumn{1}{c}{-}              & \multicolumn{1}{c}{-}                 \\[0.5ex]
                                                                           &                                                              & \multirow{2}{*}{{\glsxtrshort{vitsp16224}}} & {\glsxtrshort{vitbp16224}}          & \multicolumn{1}{c}{-}                  & \multicolumn{1}{c}{-}              & \multicolumn{1}{c}{-}                 \\
                                                                           &                                                              &                                             & {\glsxtrshort{vitsp16224}}          & $1.00 \pm 0.00$                        & $1.00 \pm 0.00$                    & $1.00 \pm 0.00$                       \\
      \cmidrule{2-7}
                                                                           & \multirow{4}{*}{{\STAB{\rotatebox[origin=c]{90}{Relative}}}} & \multirow{2}{*}{{\glsxtrshort{vitbp16224}}} & {\glsxtrshort{vitbp16224}}          & $1.00 \pm 0.00$                        & $1.00 \pm 0.00$                    & $1.00 \pm 0.00$                       \\
                                                                           &                                                              &                                             & {\glsxtrshort{vitsp16224}}          & $0.12 \pm 0.00$                        & $0.37 \pm 0.01$                    & $0.96 \pm 0.00$                       \\[0.5ex]
                                                                           &                                                              & \multirow{2}{*}{{\glsxtrshort{vitsp16224}}} & {\glsxtrshort{vitbp16224}}          & $0.12 \pm 0.00$                        & $0.39 \pm 0.01$                    & $0.96 \pm 0.00$                       \\
                                                                           &                                                              &                                             & {\glsxtrshort{vitsp16224}}          & $1.00 \pm 0.00$                        & $1.00 \pm 0.00$                    & $1.00 \pm 0.00$                       \\
      \cmidrule{1-7}
      \multirow{8}{*}{{\STAB{\rotatebox[origin=c]{90}{\textbf{kmeans}}}}}  & \multirow{4}{*}{{\STAB{\rotatebox[origin=c]{90}{Absolute}}}} & \multirow{2}{*}{{\glsxtrshort{vitbp16224}}} & {\glsxtrshort{vitbp16224}}          & $1.00 \pm 0.00$                        & $1.00 \pm 0.00$                    & $1.00 \pm 0.00$                       \\
                                                                           &                                                              &                                             & {\glsxtrshort{vitsp16224}}          & \multicolumn{1}{c}{-}                  & \multicolumn{1}{c}{-}              & \multicolumn{1}{c}{-}                 \\[0.5ex]
                                                                           &                                                              & \multirow{2}{*}{{\glsxtrshort{vitsp16224}}} & {\glsxtrshort{vitbp16224}}          & \multicolumn{1}{c}{-}                  & \multicolumn{1}{c}{-}              & \multicolumn{1}{c}{-}                 \\
                                                                           &                                                              &                                             & {\glsxtrshort{vitsp16224}}          & $1.00 \pm 0.00$                        & $1.00 \pm 0.00$                    & $1.00 \pm 0.00$                       \\
      \cmidrule{2-7}
                                                                           & \multirow{4}{*}{{\STAB{\rotatebox[origin=c]{90}{Relative}}}} & \multirow{2}{*}{{\glsxtrshort{vitbp16224}}} & {\glsxtrshort{vitbp16224}}          & $1.00 \pm 0.00$                        & $1.00 \pm 0.00$                    & $1.00 \pm 0.00$                       \\
                                                                           &                                                              &                                             & {\glsxtrshort{vitsp16224}}          & $0.11 \pm 0.00$                        & $0.25 \pm 0.01$                    & $0.97 \pm 0.00$                       \\[0.5ex]
                                                                           &                                                              & \multirow{2}{*}{{\glsxtrshort{vitsp16224}}} & {\glsxtrshort{vitbp16224}}          & $0.10 \pm 0.00$                        & $0.27 \pm 0.00$                    & $0.97 \pm 0.00$                       \\
                                                                           &                                                              &                                             & {\glsxtrshort{vitsp16224}}          & $1.00 \pm 0.00$                        & $1.00 \pm 0.00$                    & $1.00 \pm 0.00$                       \\
      \bottomrule
   \end{tabular}
\end{table}

\section{Dataset Information}
In \Cref{relative:tab:dataset-info} we summarize the datasets utilized in \Cref{chap:relative}, and for each one, we specify the number of classes, to give an idea about the classification difficulty.

\begin{table}[ht]
   \centering
   \caption{All the datasets utilized in \Cref{chap:relative} with their number of classes.}\label{relative:tab:dataset-info}
   \begin{tabular}{lll}
      \toprule
                                                      & \textbf{Dataset}    & \textbf{Number of Classes} \\
      \midrule
      \multirow{5}{*}{\rotatebox{90}{\textbf{Image}}} & \gls{mnist}         & 10                         \\
                                                      & \glsxtrlong{fmnist} & 10                         \\
                                                      & \gls{cifart}        & 10                         \\
                                                      & \gls{cifarh}        & 20 (coarse) | 100 (fine)   \\
                                                      & \gls{imagenet1k}    & 1000                       \\
      \midrule
      \multirow{3}{*}{\rotatebox{90}{\textbf{Graph}}} & \gls{cora}          & 7                          \\
                                                      & \gls{citeseer}      & 6                          \\
                                                      & \gls{pubmed}        & 3                          \\
      \midrule
      \multirow{3}{*}{\rotatebox{90}{\textbf{Text}}}  & \gls{trec}          & 6 (coarse) |  50 (fine)    \\
                                                      & \gls{dbpedia}       & 14                         \\
                                                      & \gls{amazon}        & 2 (coarse) | 5 (fine)      \\
      \bottomrule
   \end{tabular}
\end{table}

\section{Implementation Details}\label{relative:sec:implementatin-details}
In this Section, following the corresponing sections in \Cref{chap:relative}, we report implementation details for all the experimental settings considered.

\paragraph{Tools \& Technologies.}
In all the experiments presented in this work, the following tools were used:
\begin{itemize}
   \item \textit{NN-Template} \cite{nn-template}, to easily bootstrap the project and enforce best practices.
   \item \textit{PyTorch Lightning} \citep{lightning}, to ensure reproducible results while also getting a clean and modular codebase.
   \item \textit{Weights and Biases} \citep{wandb}, to log experiments and compare runs across huge sweeps.
   \item \textit{Transformers by HuggingFace} \citep{wolf-etal-2020-transformers}, to get ready-to-use transformers for both text and images.
   \item \textit{Datasets by HuggingFace} \citep{lhoest-etal-2021-datasets}, to access most of the NLP datasets and ImageNet for CV.
   \item \textit{DVC} \citep{dvc}, for data versioning.
   \item \textit{PyTorch Geometric} \citep{pyg}, to handle graph datasets and get ready-to-use GNN architectures.
\end{itemize}

\subsection{Word Embeddings}\label{relative:appendix:word-embeddings}
For both the Figure and the Table in \Cref{relative:sec:word-embeddings}, the number of anchors is set to 300 for a fair comparison with the dimensionality of the original spaces.
For visualization purposes, we needed the figure to both show an easy clusterable and restricted set of word embeddings. They are obtained by subsampling the shared vocabulary with the following procedure: we select 4 random pivot words, and for each of them we consider the top-200 words in their neighborhood. This results in a total of 800 points divided in 4 clusters, the ones used only for the visualization part.
For the quantitative part (table results), we select 20K random words from the shared vocabulary with a fixed seed for reproducibility purposes.

\begin{figure}[ht]
   \centering
   \begin{overpic}[trim=-1cm 1cm -0.5cm 0cm,width=.32\linewidth]{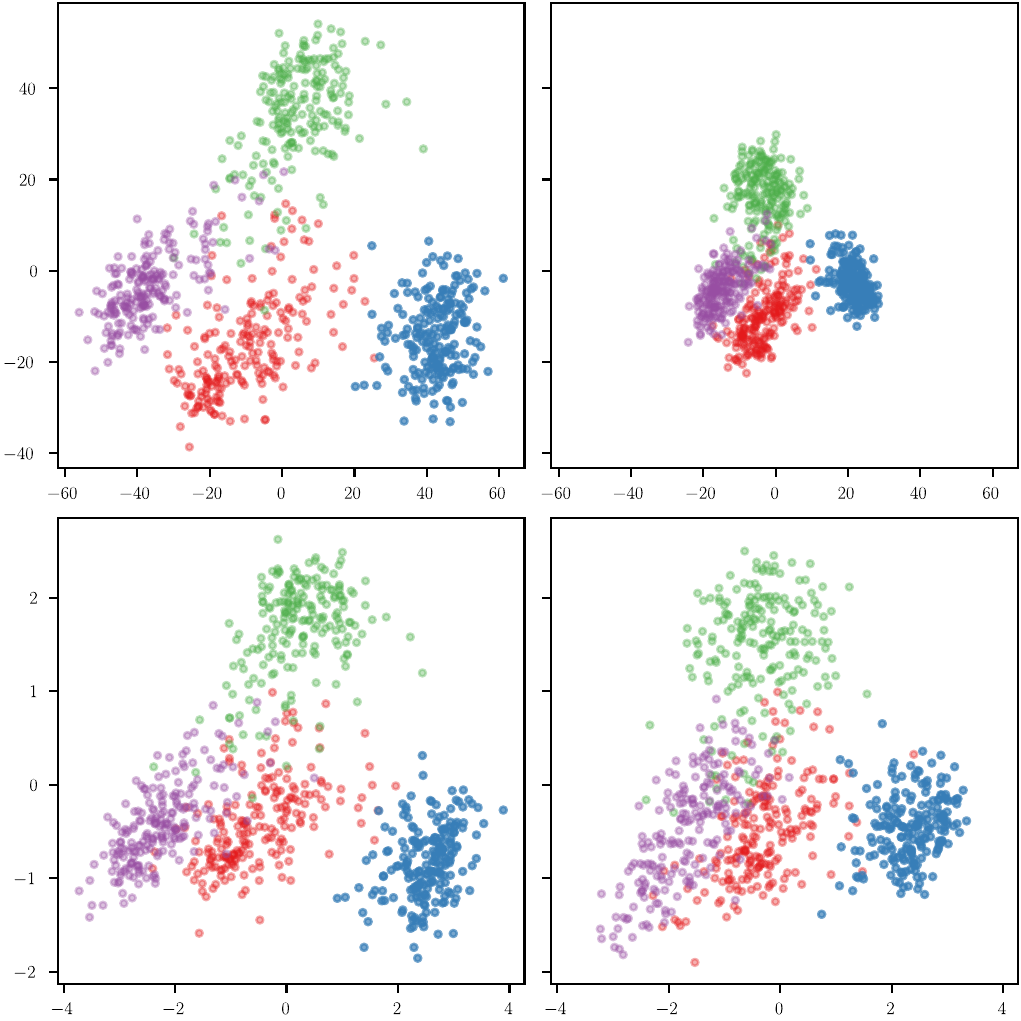}
      \put(13, 89){{\glsxtrshort{vitbp16224}}}
      \put(58.5, 89){{\glsxtrshort{vitsp16224}}}
      \put(-1, 53){\rotatebox{90}{\small Absolute}}
      \put(-1, 5){\rotatebox{90}{\small Relative}}
   \end{overpic}
   \begin{overpic}[trim=-1cm 1cm -0.5cm -.50cm,width=.32\linewidth]{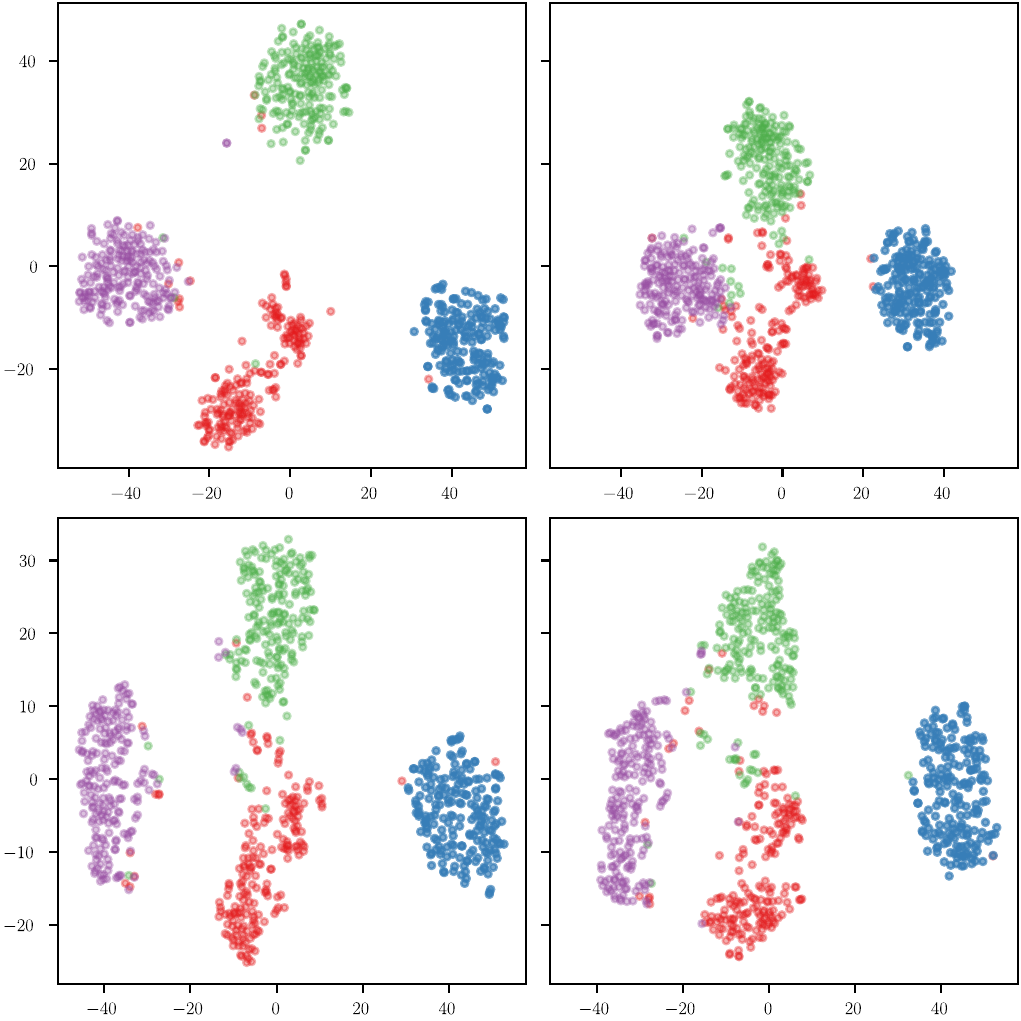}
      \put(13, 89){{\glsxtrshort{vitbp16224}}}
      \put(58.5, 89){{\glsxtrshort{vitsp16224}}}
   \end{overpic}
   \begin{overpic}[trim=-1cm 1cm -0.5cm 0cm,width=.32\linewidth]{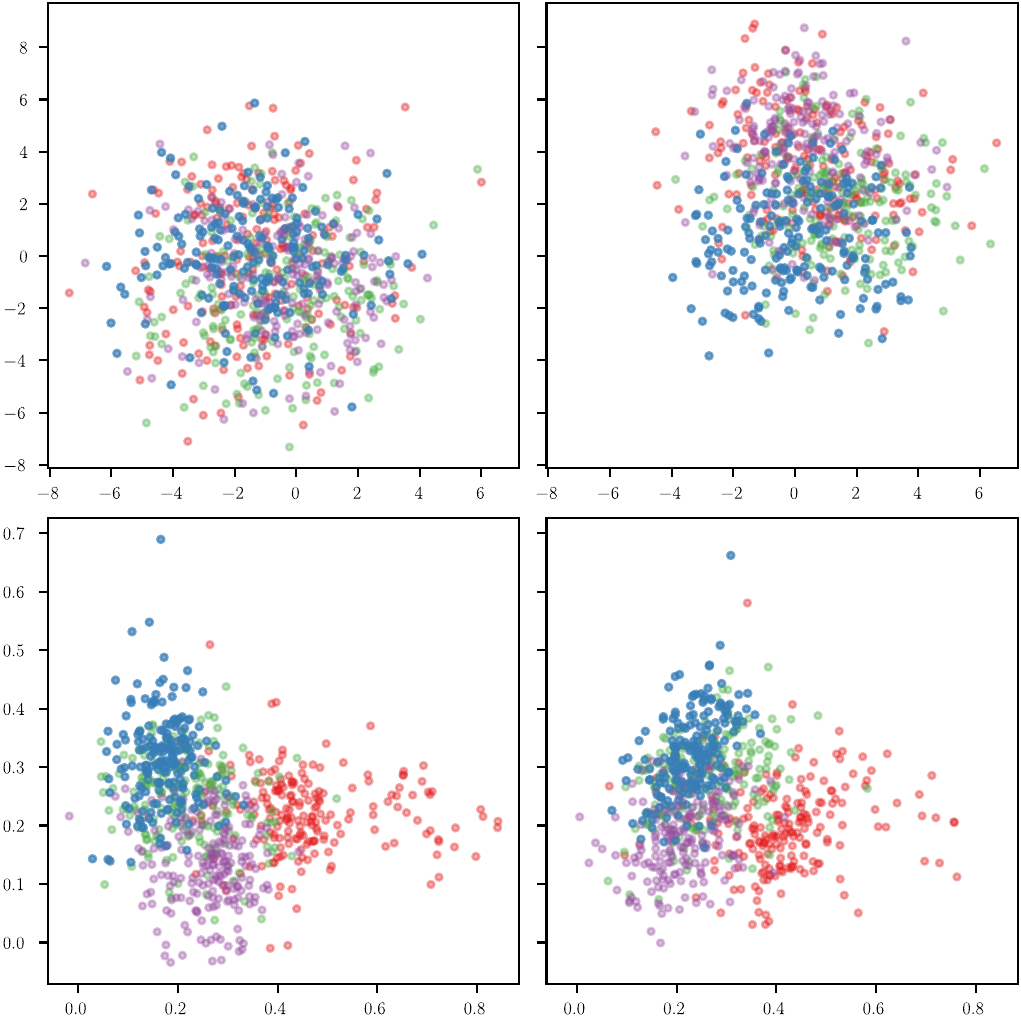}
      \put(13, 89){{\glsxtrshort{vitbp16224}}}
      \put(58.5, 89){{\glsxtrshort{vitsp16224}}}
   \end{overpic}
   \caption[\gls{cifart} embeddings similarity across different models]{{Different dimensionality reduction techniques applied to absolute and relative spaces on \gls{cifart}. From left to right: PCA (Principal Component Analysis), tSNE, and visualizing only their first two dimensions. Only 800 randomly sampled points are shown, belonging to the classes "bird", "ship", "cat", and "frog".}}
   \label{relative:fig:qualitative-retrieval-cv}
\end{figure}
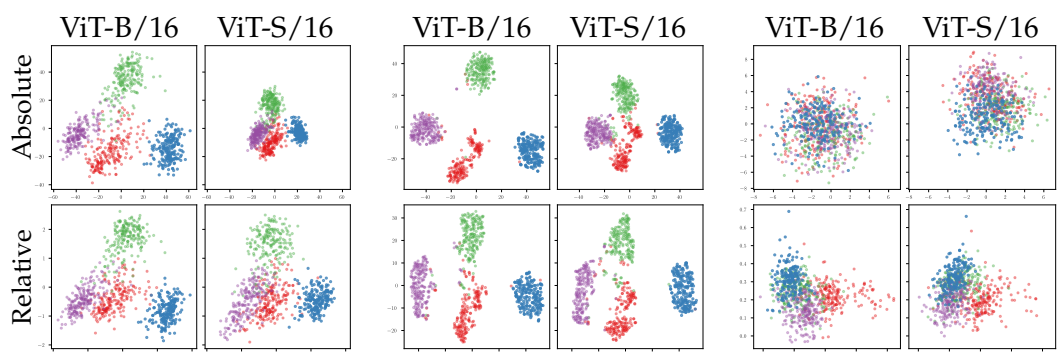

   {
      For the computer vision counterpart (\Cref{relative:fig:qualitative-retrieval-cv,relative:tab:quantitative-analysis-cv-all}), the procedure is similar but with the following differences: (i) the number of anchors is set to 500 to balance between the different encoding dimensions of the two transformers (384 for \glsxtrshort{vitsp16224} and 768 for \glsxtrshort{vitbp16224}); (ii) the subsampling for visualization purposes is done by selecting 4 classes and randomly picking 200 samples for each of them;
   }

\paragraph{Evaluation metrics.}
Consider the
source space ${\gls{ZX}'}$ and target space ${\gls{ZY}'}$
and a set of $\approx$ 20k samples ${\gls{S}} \subseteq (\gls{X} \cap \gls{Y})$ (words for the NLP test, images for the CV one);
for any sample ${\gls{s}} \in {\gls{S}}$, we compute its representation in ${\gls{ZX}'}$ and ${\gls{ZY}'}$ through the functions  $f_{\gls{ZX}'}: {\gls{S}} \to {\gls{ZX}'}$  and $f_{\gls{ZY}'}: {\gls{S}} \to {\gls{ZY}'}$ (e.g. $f$ can be the encoder composed with a relative proejction) and define the metrics as follows:
\begin{align*}
   \operatorname{{Jaccard}({\gls{s}})} & = \frac{|\operatorname{KNN}_k^{\gls{ZX}'}(f_{\gls{ZX}'}({\gls{s}})) \cap \operatorname{KNN}_k^{\gls{ZY}'}(f_{\gls{ZX}'}({\gls{s}}))|}{|\operatorname{KNN}_k^{\gls{ZX}'}(f_{\gls{ZX}'}({\gls{s}})) \cup \operatorname{KNN}_k^{\gls{ZY}'}(f_{\gls{ZX}'}({\gls{s}}))|} \\
   \operatorname{{MRR}({\gls{s}})}     & = \frac{1}{\operatorname{Rank}_{\gls{ZY}'}(f_{\gls{ZX}'}({\gls{s}}), f_{\gls{ZY}'}({\gls{s}}))}                                                                                                                                                                     \\
   \operatorname{{Cosine}({\gls{s}})}  & = \frac{f_{\gls{ZX}'}({\gls{s}}) \cdot f_{\gls{ZY}'}({\gls{s}})}{\|f_{\gls{ZX}'}({\gls{s}})\| \|f_{\gls{ZY}'}({\gls{s}})\|}
\end{align*}
where $\operatorname{KNN}_k^{\gls{S}}(\vv)$ is a function that returns the $k$-top similar samples (according to cosine similarity) to $\vv$ in the space ${\gls{S}}$, and $\operatorname{Rank}_{\gls{S}}(\vv, \vu)$ is a function that returns the index at which $\vu$ is found in the ordered  $\operatorname{KNN}_k^{\gls{S}}(\vv)$. The final score for each metric is the mean over each ${\gls{s}} \in S$.

\subsection{Relative representation space correlations}
\begin{figure}[ht]
   \centering
   \begin{overpic}[trim=0cm 0cm 0cm 0cm,width=.6\linewidth]{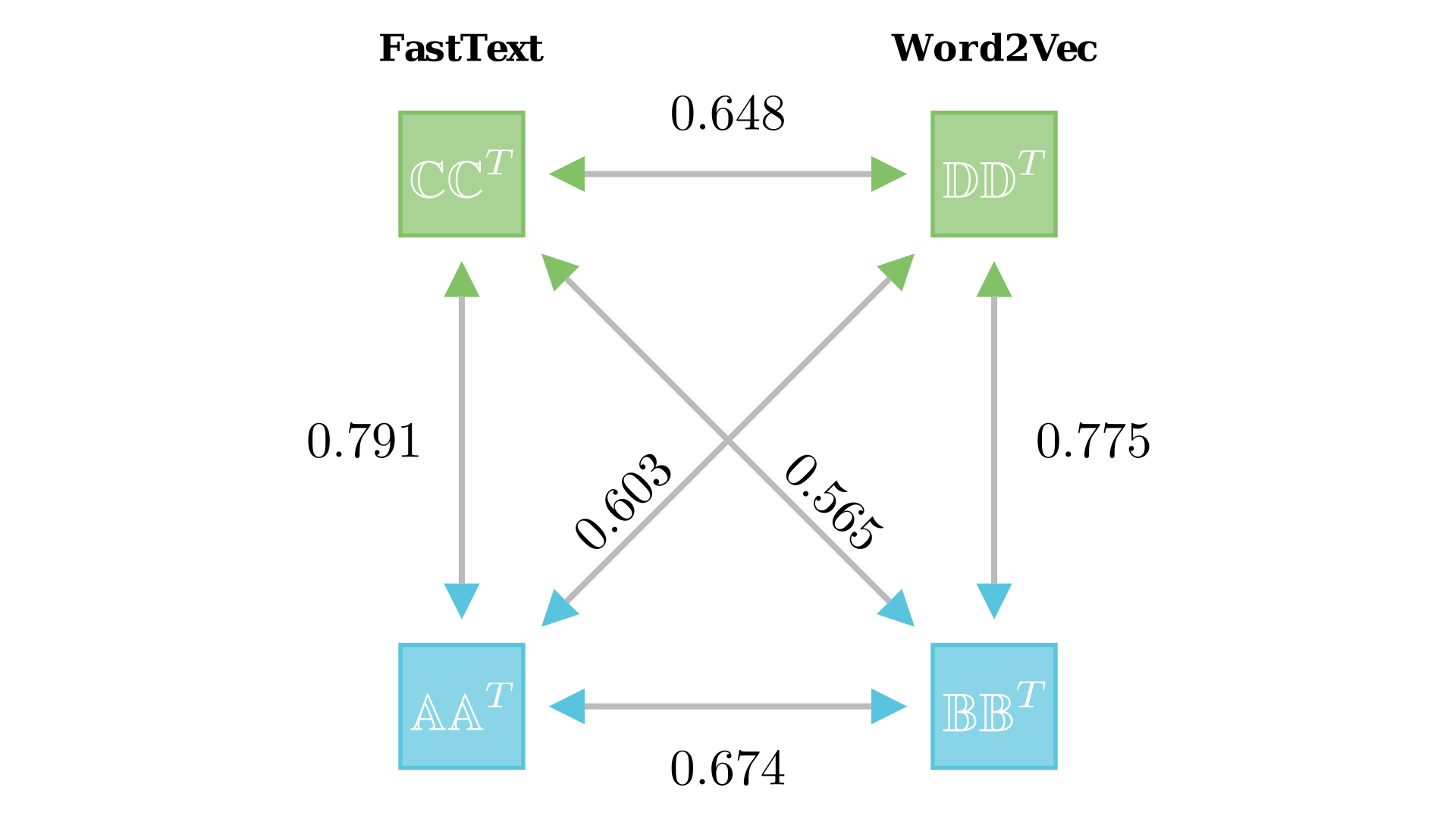}

   \end{overpic}

   \caption[{Self similiarities correlations} between each space]{{Self similiarities correlations} between each space, measured with the Pearson correlation coefficient. In blue, we denote the self similarities in the absolute spaces $\mathbb{A}$, $\mathbb{B}$ of \gls{fasttext} and \gls{word2vec}; in green we depict the relative spaces $\mathbb{C}$, $\mathbb{D}$. The correlation in the vertical arrows indicate how much the underlying metric in the abolute space is preserved by the relative coordinate transformation.}%
   \label{relative:fig:rel_space_correlations}
\end{figure}

{ In this Section, we analyze how similarities in absolute and relative spaces are correlated.
   Let us consider two spaces alignable in the relative space. We denote elements of the spaces with $\mathbb{A} \in \mathbb{R}^{m_1\times n_1}$ and $\mathbb{B}\in \mathbb{R}^{m_2\times n_2}$ and corresponding relative embeddings with $\mathbb{C} \in \mathbb{R}^{m_1\times d}$, $\mathbb{D} \in \mathbb{R}^{m_2\times d}$. Examples of  $\mathbb{A}$ and $\mathbb{B}$ can be the \gls{fasttext} and \gls{word2vec} word embedding spaces. We already observed in \Cref{relative:fig:latent-rotation-comparison} how the spaces $\mathbb{A}$ and $\mathbb{B}$ are well aligned in the relative space. We can go further and analyze how self similarities in each space are preserved by the relative transform. In \Cref{relative:fig:rel_space_correlations}, we show that relative representations not only facilitate latent communication, but also preserve the underlying (absolute) latent space metric up to a certain degree.}

\subsection{Latent distance as a performance proxy}
The hyperperameters used in \Cref{relative:rebase:sec:manifold-performance} are summarized in \Cref{relative:tab:manifold-performance-parameters}.

\begin{table}[ht]
   \centering
   \caption[Hyperparameter grid search performed in \Cref{relative:rebase:sec:manifold-performance}]{The {reference} model and exhaustive hyperparameter combinations pertaining \Cref{relative:rebase:sec:manifold-performance}.}
   \label{relative:tab:manifold-performance-parameters}
   \begin{tabular}{lll}
      \toprule
      Hyperparameter         & Reference Model  & Sweep                                                      \\
      \midrule
      Seed                   & \texttt{1}       & \texttt{0}, \texttt{1}, \texttt{2}, \texttt{3}, \texttt{4} \\
      Epochs                 & \texttt{500}     & \texttt{10}, \texttt{30}, \texttt{50}                      \\
      Number of layers       & \texttt{32}      & \texttt{32}, \texttt{64}                                   \\
      Dropout Probability    & \texttt{0.5}     & \texttt{0.1}, \texttt{0.5}                                 \\
      Hidden Activations     & \texttt{ReLU}    & \texttt{ReLU}, \texttt{Tanh}                               \\
      Convolution Activation & \texttt{ReLU}    & \texttt{ReLU}, \texttt{Tanh}                               \\
      Optimizer              & \texttt{Adam}    & \texttt{Adam}, \texttt{SGD}                                \\
      Learning Rate          & \texttt{0.02}    & \texttt{0.01}, \texttt{0.02}                               \\
      Graph Embedder         & \texttt{GCNConv} & \texttt{GCNConv}, \texttt{GINConv}                         \\
      \bottomrule
   \end{tabular}
\end{table}

\subsection{Training with Absolute vs. Relative Representations}
{The models trained on relative representations do not backpropagate through the anchors, which encourages a smoother optimization of the anchors' representations.}

\paragraph{Image Classification.} The architecture is a standard deep \gls{cnn}. We run a sweep for each dataset where we vary only the random seed (over 10 possible in total). We then aggregate by dataset and encoding type to obtain the final results with their standard deviation.

\paragraph{Graph Classification.} We run a sweep identical to the one in \Cref{relative:tab:manifold-performance-parameters} for the reference model, except that we sweep on the ``Number of layers'' with two values: 32 and 64. Each configuration is repeated with 10 different seeds, then we aggregate by dataset and encoding type to obtain the final results with their standard deviation.

\subsection{Image Reconstruction}
The relative and absolute models appearing in \Cref{relative:fig:encoder-decoder-swap} are vanilla \glsxtrshortpl{ae} and  \glsxtrshortpl{vae}, the same for all the datasets, and have a comparable number of trainable parameters. Their architecture is composed by simple convolutions, deconvolutions and mean squared error as reconstruction loss. The number of anchors is $500$ and the latent dimensionality of the absolute representations is $500$.

\subsection{Text Classification}
We report in \Cref{relative:tab:transformers-nlp,relative:tab:transformers-nlp-mono,relative:tab:wikimatrix} details on the transformers and anchors adopted in \Cref{relative:sec:nlp-app}.

\begin{table}[ht]
   \centering
   \caption[The pretrained transformers used for the \textit{Cross-lingual} setting]{The HuggingFace transformers models employed in \Cref{relative:sec:nlp-app} to tackle the \textit{Cross-lingual} setting.}
   \label{relative:tab:transformers-nlp}
   \begin{tabular}{lll}
      \toprule
      Language & HuggingFace transformers name             & Encoding Dim \\
      \midrule
      English  & \glsxtrlong{robertab}                     & 768          \\
      Spanish  & \texttt{PlanTL-GOB-ES/roberta-base-bne}   & 768          \\
      French   & \texttt{ClassCat/roberta-base-french}     & 768          \\
      Japanese & \texttt{nlp-waseda/roberta-base-japanese} & 768          \\
      \bottomrule
   \end{tabular}
\end{table}

\begin{table}[ht]
   \centering
   \caption[The pretrained transformers used for the \textit{Cross-architecture} setting]{The HuggingFace transformers models employed in \Cref{relative:sec:nlp-app} to tackle the \textit{Cross-architecture} setting.}
   \label{relative:tab:transformers-nlp-mono}
   \begin{tabular}{ll}
      \toprule
      HuggingFace transformers name & Encoding Dim \\
      \midrule
      \glsxtrlong{bertbc}           & 768          \\
      \glsxtrlong{bertbu}           & 768          \\
      \glsxtrlong{electrabd}        & 768          \\
      \glsxtrlong{robertab}         & 768          \\
      \bottomrule
   \end{tabular}
\end{table}

\paragraph{Preprocessing.} Following the original work in which the \gls{amazon} dataset was proposed \citep{amazon-reviews}, we utilize both the \textit{title} and \textit{body} of each review. We differ in not using the category and in how we merge them; namely, we add the title as prefix for the body and add a full stop as separator when needed (avoiding duplicates). To obtain a single latent encoding for each sample, with fixed shape, we take the last hidden state and select the representation corresponding to the \emph{[CLS]} token.

\paragraph{Wikipedia anchors.}
We use WikiMatrix, a corpus of sentences extracted from Wiki\-pedia. The sentences are parallel between pairs of languages (i.e., same sentences translated in two languages), and since we are looking for a collection of parallel anchors between all 4 languages, we decided to use the English language as a pivot to compute the intersection. To get the final results, we considered only the sentences with margin score $\ge 1.06$, getting high-quality sentence alignments.
In \Cref{relative:tab:wikimatrix} we show the total number of parallel sentences when computing the intersections. We randomly selected 768 samples to use as anchors.

\begin{table}[ht]
   \centering
   \caption[WikiMatrix analysis further details]{
      WikiMatrix analysis. Each row shows the number of parallel sentences having a translation available in all the languages of that row. Since we consider all four languages, we have $3338$ parallel sentences available.}
   \label{relative:tab:wikimatrix}
   \begin{tabular}{ll}
      \toprule
      Languages      & Number of Sentences \\ \midrule
      en, es         & 2302527             \\
      en, ja         & 264259              \\
      en, fr         & 1682477             \\
      en, es, fr     & 23200               \\
      en, es, ja     & 147665              \\
      en, fr, ja     & 20990               \\
      en, es, fr, ja & \textbf{3338}       \\ \bottomrule
   \end{tabular}
\end{table}

\subsection{Image Classification}
The details of the transformers used in \Cref{relative:sec:cv-app} are summarized in \Cref{relative:tab:transformers-cv}.

\begin{table}[ht]
   \centering
   \caption{Timm transformers used in \Cref{relative:sec:cv-app}.}
   \label{relative:tab:transformers-cv}
   \begin{tabular}{llll}
      \toprule
      \multicolumn{1}{c}{Version} & \multicolumn{1}{c}{Timm model name} & \multicolumn{1}{c}{Encoding Dim} & \multicolumn{1}{c}{Training data} \\
      \midrule
      ViT                         & \glsxtrlong{vitbp16224}             & 768                              & JFT-300M, ImageNet                \\
      ViT                         & \glsxtrlong{vitsp16224}             & 384                              & ImageNet                          \\
      ViT                         & \glsxtrlong{vitbr50384}             & 768                              & ImageNet                          \\
      RexNet                      & \glsxtrlong{rexnet100}              & 1280                             & ImageNet                          \\
      \bottomrule
   \end{tabular}
\end{table}

\section{Additional results}
\begin{figure}[ht]
   \centering
   \begin{overpic}[trim=-0.27cm -0.15cm 0cm 0cm,clip,width=1\linewidth]{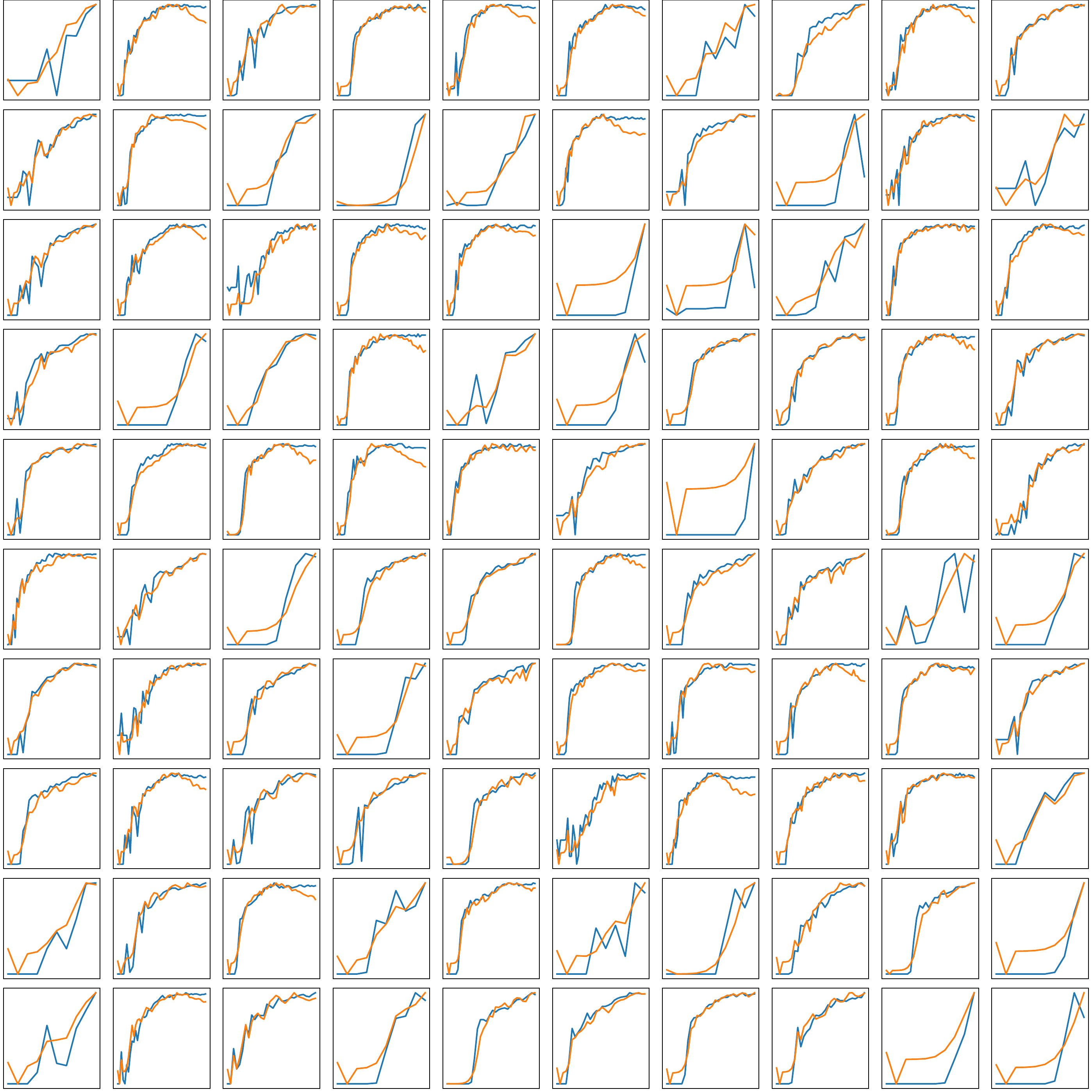}
   \end{overpic}
   \caption[Correlations between performance and latent similarity]{Correlations between performance and latent similarity with the reference model for multiple different models, over time.}
   \label{relative:fig:correlation-grid}
\end{figure}

In this Section we report additional results on the correlation between latent similarity and performance in \Cref{relative:fig:correlation-grid}, results on the multilingual stitching both with Amazon coarse-grained in \Cref{relative:tab:multilingual-full-coarse-grained} and fine-grained in \Cref{relative:tab:multilingual-full-fine-grained}, results on the image classification stitching on \gls{cifarh} fine-grained in \Cref{relative:tab:cifar100-fine}. Moreover, we evaluate the stitching performance of a multilingual transformer in \Cref{relative:tab:xmlr-multilingual-full-fine-grained}.

\begin{table}[ht]
   \scriptsize
   \centering
   \caption[Further results on Zero-Shot Stitching on \gls{amazon} coarse-grained]{Stitching performance comparison with different encodings techniques. The table reports the mean weighted F1 (± std) and \glsxtrshort{mae} classification performance on \gls{amazon} coarse-grained, across 5 different seeds. All the language pairs are shown.}
   \label{relative:tab:multilingual-full-coarse-grained}
   \begin{tabular}{clcccccc}
      \toprule
                          &                  & \multicolumn{2}{c}{Absolute} & \multicolumn{4}{c}{Relative}                                                                                              \\
      \cmidrule(lr){3-4}
      \cmidrule(l){5-8}
                          &                  & \multicolumn{2}{c}{}         & \multicolumn{2}{c}{Translated} & \multicolumn{2}{c}{Wikipedia}                                                            \\
      \cmidrule(lr){5-6}
      \cmidrule(l){7-8}
      \textbf{Decoder}    & \textbf{Encoder} & FScore                       & \glsxtrshort{mae}              & FScore                        & \glsxtrshort{mae} & FScore           & \glsxtrshort{mae} \\
      \midrule
      \multirow{4}{*}{en} & en               & $91.54 \pm 0.58$             & $0.08 \pm 0.01$                & $90.06 \pm 0.60$              & $0.10 \pm 0.01$   & $90.45 \pm 0.52$ & $0.10 \pm 0.01$   \\
                          & es               & $43.67 \pm 1.09$             & $0.56 \pm 0.01$                & $82.78 \pm 0.81$              & $0.17 \pm 0.01$   & $78.53 \pm 0.30$ & $0.21 \pm 0.00$   \\
                          & fr               & $54.41 \pm 1.61$             & $0.45 \pm 0.02$                & $78.49 \pm 0.66$              & $0.21 \pm 0.01$   & $70.41 \pm 0.57$ & $0.29 \pm 0.01$   \\
                          & ja               & $48.72 \pm 0.90$             & $0.51 \pm 0.01$                & $65.72 \pm 0.55$              & $0.34 \pm 0.01$   & $66.31 \pm 0.80$ & $0.34 \pm 0.01$
      \\[2ex]
      \multirow{4}{*}{es} & en               & $33.23 \pm 1.00$             & $0.66 \pm 0.01$                & $78.68 \pm 2.74$              & $0.21 \pm 0.03$   & $76.65 \pm 3.23$ & $0.23 \pm 0.03$   \\
                          & es               & $91.64 \pm 1.02$             & $0.08 \pm 0.01$                & $89.96 \pm 0.77$              & $0.10 \pm 0.01$   & $89.62 \pm 0.94$ & $0.10 \pm 0.01$   \\
                          & fr               & $47.66 \pm 0.70$             & $0.52 \pm 0.01$                & $78.57 \pm 1.80$              & $0.21 \pm 0.02$   & $75.25 \pm 0.76$ & $0.25 \pm 0.01$   \\
                          & ja               & $53.10 \pm 2.27$             & $0.46 \pm 0.02$                & $67.69 \pm 0.24$              & $0.32 \pm 0.00$   & $61.84 \pm 0.61$ & $0.38 \pm 0.01$
      \\[2ex]
      \multirow{4}{*}{fr} & en               & $51.00 \pm 2.63$             & $0.49 \pm 0.03$                & $83.32 \pm 1.80$              & $0.17 \pm 0.02$   & $75.55 \pm 0.37$ & $0.24 \pm 0.00$   \\
                          & es               & $51.96 \pm 2.81$             & $0.48 \pm 0.03$                & $82.50 \pm 0.83$              & $0.17 \pm 0.01$   & $77.12 \pm 0.88$ & $0.23 \pm 0.01$   \\
                          & fr               & $88.22 \pm 0.75$             & $0.12 \pm 0.01$                & $85.68 \pm 1.37$              & $0.14 \pm 0.01$   & $86.45 \pm 0.96$ & $0.13 \pm 0.01$   \\
                          & ja               & $50.32 \pm 4.16$             & $0.50 \pm 0.04$                & $69.38 \pm 0.73$              & $0.31 \pm 0.01$   & $62.79 \pm 0.27$ & $0.37 \pm 0.00$
      \\[2ex]
      \multirow{4}{*}{ja} & en               & $53.82 \pm 2.62$             & $0.46 \pm 0.03$                & $68.66 \pm 3.62$              & $0.31 \pm 0.04$   & $70.26 \pm 3.16$ & $0.29 \pm 0.03$   \\
                          & es               & $44.91 \pm 2.21$             & $0.55 \pm 0.02$                & $70.37 \pm 6.94$              & $0.29 \pm 0.06$   & $58.54 \pm 1.21$ & $0.41 \pm 0.01$   \\
                          & fr               & $66.46 \pm 1.30$             & $0.34 \pm 0.01$                & $76.49 \pm 1.13$              & $0.23 \pm 0.01$   & $63.94 \pm 2.70$ & $0.36 \pm 0.02$   \\
                          & ja               & $83.30 \pm 0.67$             & $0.17 \pm 0.01$                & $81.04 \pm 0.82$              & $0.19 \pm 0.01$   & $80.80 \pm 1.25$ & $0.19 \pm 0.01$   \\
      \bottomrule
   \end{tabular}
\end{table}

\begin{table}[ht]
   \scriptsize
   \centering
   \caption[Further results on Zero-Shot Stitching on \gls{amazon} fine-grained]{Stitching performance comparison with different encodings techniques. The table reports the mean weighted F1 (± std) and \glsxtrshort{mae} classification performance on \gls{amazon} fine-grained, across 5 different seeds. All the language pairs are shown.  }
   \label{relative:tab:multilingual-full-fine-grained}
   \begin{tabular}{clcccccc}
      \toprule
                          &                  & \multicolumn{2}{c}{Absolute} & \multicolumn{4}{c}{Relative}                                                                                              \\
      \cmidrule(lr){3-4}
      \cmidrule(l){5-8}
                          &                  & \multicolumn{2}{c}{}         & \multicolumn{2}{c}{Translated} & \multicolumn{2}{c}{Wikipedia}                                                            \\
      \cmidrule(lr){5-6}
      \cmidrule(l){7-8}
      \textbf{Decoder}    & \textbf{Encoder} & FScore                       & \glsxtrshort{mae}              & FScore                        & \glsxtrshort{mae} & FScore           & \glsxtrshort{mae} \\
      \midrule
      \multirow{4}{*}{en} & en               & $65.46 \pm 2.89$             & $0.38 \pm 0.02$                & $61.18 \pm 1.92$              & $0.44 \pm 0.02$   & $62.36 \pm 2.23$ & $0.43 \pm 0.02$   \\
                          & es               & $22.70 \pm 0.41$             & $1.39 \pm 0.03$                & $51.67 \pm 1.20$              & $0.62 \pm 0.01$   & $45.40 \pm 0.68$ & $0.76 \pm 0.01$   \\
                          & fr               & $30.75 \pm 0.67$             & $1.19 \pm 0.02$                & $49.18 \pm 0.83$              & $0.69 \pm 0.02$   & $40.29 \pm 0.90$ & $0.91 \pm 0.02$   \\
                          & ja               & $24.85 \pm 0.91$             & $1.37 \pm 0.07$                & $37.34 \pm 1.49$              & $0.99 \pm 0.02$   & $37.73 \pm 0.70$ & $1.01 \pm 0.02$
      \\[2ex]
      \multirow{4}{*}{es} & en               & $21.24 \pm 0.81$             & $1.43 \pm 0.07$                & $51.02 \pm 2.54$              & $0.68 \pm 0.05$   & $47.70 \pm 5.08$ & $0.73 \pm 0.10$   \\
                          & es               & $61.29 \pm 3.04$             & $0.43 \pm 0.02$                & $57.89 \pm 3.80$              & $0.48 \pm 0.03$   & $57.96 \pm 4.40$ & $0.48 \pm 0.03$   \\
                          & fr               & $29.02 \pm 0.85$             & $1.26 \pm 0.05$                & $48.40 \pm 1.02$              & $0.71 \pm 0.02$   & $44.92 \pm 1.83$ & $0.77 \pm 0.01$   \\
                          & ja               & $29.23 \pm 1.32$             & $1.22 \pm 0.02$                & $37.22 \pm 1.56$              & $1.03 \pm 0.04$   & $34.56 \pm 0.87$ & $1.08 \pm 0.04$
      \\[2ex]
      \multirow{4}{*}{fr} & en               & $27.39 \pm 1.22$             & $1.23 \pm 0.06$                & $45.55 \pm 3.55$              & $0.76 \pm 0.09$   & $39.01 \pm 1.25$ & $0.88 \pm 0.06$   \\
                          & es               & $29.47 \pm 3.68$             & $1.18 \pm 0.07$                & $40.29 \pm 1.72$              & $0.90 \pm 0.04$   & $41.29 \pm 2.01$ & $0.83 \pm 0.04$   \\
                          & fr               & $56.40 \pm 1.89$             & $0.51 \pm 0.01$                & $53.58 \pm 0.70$              & $0.57 \pm 0.01$   & $54.23 \pm 0.95$ & $0.56 \pm 0.01$   \\
                          & ja               & $25.92 \pm 1.31$             & $1.25 \pm 0.05$                & $38.60 \pm 1.03$              & $0.96 \pm 0.02$   & $35.22 \pm 0.56$ & $1.08 \pm 0.02$
      \\[2ex]
      \multirow{4}{*}{ja} & en               & $29.36 \pm 0.59$             & $1.17 \pm 0.04$                & $38.19 \pm 2.28$              & $0.88 \pm 0.03$   & $36.57 \pm 1.72$ & $0.98 \pm 0.02$   \\
                          & es               & $25.64 \pm 1.77$             & $1.28 \pm 0.04$                & $34.23 \pm 2.62$              & $1.00 \pm 0.05$   & $33.16 \pm 2.28$ & $1.06 \pm 0.03$   \\
                          & fr               & $31.79 \pm 1.91$             & $1.06 \pm 0.02$                & $38.50 \pm 2.46$              & $0.89 \pm 0.02$   & $36.68 \pm 3.14$ & $1.00 \pm 0.05$   \\
                          & ja               & $54.09 \pm 1.35$             & $0.60 \pm 0.02$                & $50.89 \pm 1.70$              & $0.65 \pm 0.02$   & $51.64 \pm 1.47$ & $0.65 \pm 0.02$   \\
      \bottomrule
   \end{tabular}
\end{table}

\begin{table}[ht]
   \small
   \centering
   \caption[Zero-shot stitching performance comparison with XLM-R multilingual]{
      Stitching performance comparison on XLM-R, a multilingual model by design.
      The table reports the mean weighted F1 ($\pm$ std) and \glsxtrshort{mae} classification performance on \gls{amazon} fine-grained, across 5 different seeds.}
   \label{relative:tab:xmlr-multilingual-full-fine-grained}
   \begin{tabular}{llcccc}
      \toprule
                          &                  & \multicolumn{2}{c}{Absolute} & \multicolumn{2}{c}{Relative}                                        \\
      \textbf{Decoder}    & \textbf{Encoder} & FScore                       & \glsxtrshort{mae}            & FScore           & \glsxtrshort{mae} \\
      \midrule
      \multirow{4}{*}{en} & en               & $65.27 \pm 0.94$             & $0.41 \pm 0.01$              & $58.24 \pm 1.92$ & $0.51 \pm 0.03$   \\
                          & es               & $59.55 \pm 0.76$             & $0.48 \pm 0.01$              & $52.81 \pm 1.57$ & $0.62 \pm 0.02$   \\
                          & fr               & $58.58 \pm 1.04$             & $0.49 \pm 0.01$              & $54.01 \pm 1.34$ & $0.59 \pm 0.02$   \\
                          & ja               & $57.98 \pm 0.77$             & $0.52 \pm 0.01$              & $48.47 \pm 2.67$ & $0.71 \pm 0.04$   \\
      \cmidrule{1-6}
      \multirow{4}{*}{es} & en               & $60.32 \pm 1.50$             & $0.47 \pm 0.01$              & $45.69 \pm 2.19$ & $0.87 \pm 0.07$   \\
                          & es               & $61.25 \pm 1.74$             & $0.44 \pm 0.01$              & $57.61 \pm 0.73$ & $0.51 \pm 0.01$   \\
                          & fr               & $59.50 \pm 1.41$             & $0.47 \pm 0.01$              & $45.16 \pm 3.30$ & $0.83 \pm 0.09$   \\
                          & ja               & $58.24 \pm 1.31$             & $0.51 \pm 0.02$              & $41.14 \pm 1.76$ & $0.99 \pm 0.05$   \\
      \cmidrule{1-6}
      \multirow{4}{*}{fr} & en               & $58.00 \pm 4.21$             & $0.49 \pm 0.03$              & $52.37 \pm 1.66$ & $0.66 \pm 0.03$   \\
                          & es               & $56.87 \pm 3.79$             & $0.49 \pm 0.03$              & $54.99 \pm 0.46$ & $0.57 \pm 0.01$   \\
                          & fr               & $57.99 \pm 3.88$             & $0.47 \pm 0.02$              & $57.00 \pm 0.90$ & $0.52 \pm 0.01$   \\
                          & ja               & $55.83 \pm 3.32$             & $0.53 \pm 0.03$              & $39.15 \pm 1.21$ & $1.02 \pm 0.03$   \\
      \cmidrule{1-6}
      \multirow{4}{*}{ja} & en               & $59.53 \pm 1.73$             & $0.48 \pm 0.01$              & $39.46 \pm 2.34$ & $1.04 \pm 0.07$   \\
                          & es               & $57.02 \pm 1.36$             & $0.51 \pm 0.00$              & $40.74 \pm 2.75$ & $0.97 \pm 0.09$   \\
                          & fr               & $57.48 \pm 1.06$             & $0.51 \pm 0.01$              & $43.36 \pm 3.70$ & $0.89 \pm 0.11$   \\
                          & ja               & $61.43 \pm 0.97$             & $0.45 \pm 0.01$              & $57.67 \pm 1.17$ & $0.51 \pm 0.01$   \\
      \bottomrule
   \end{tabular}
\end{table}

\begin{table}[ht]
   \small
   \centering
   \caption[Further results on Zero-Shot Stitching on \gls{cifarh} fine-grained]{
      Stitching performance comparison with different encodings techniques.
      The table reports the mean weighted F1 ($\pm$ std) classification performance on \gls{cifarh} fine-grained, across 5 different seeds.}
   \label{relative:tab:cifar100-fine}
   \begin{tabular}{llll}
      \toprule
      \textbf{Decoder}                            & \textbf{Encoder}           & \multicolumn{1}{c}{Absolute} & \multicolumn{1}{c}{Relative} \\
      \midrule
      \multirow{4}{*}{\glsxtrshort{rexnet100}}    & \glsxtrshort{rexnet100}    & $72.77 \pm 0.19$             & $71.39 \pm 0.18$             \\
                                                  & {\glsxtrshort{vitbp16224}} & \multicolumn{1}{c}{-}        & $40.68 \pm 0.50$             \\
                                                  & {\glsxtrshort{vitbr50384}} & \multicolumn{1}{c}{-}        & $38.18 \pm 0.24$             \\
                                                  & {\glsxtrshort{vitsp16224}} & \multicolumn{1}{c}{-}        & $44.11 \pm 0.84$             \\
      \cmidrule{1-4}
      \multirow{4}{*}{{\glsxtrshort{vitbp16224}}} & \glsxtrshort{rexnet100}    & \multicolumn{1}{c}{-}        & $57.81 \pm 0.39$             \\
                                                  & {\glsxtrshort{vitbp16224}} & $88.69 \pm 0.14$             & $87.05 \pm 0.34$             \\
                                                  & {\glsxtrshort{vitbr50384}} & $1.08 \pm 0.19$              & $66.65 \pm 1.79$             \\
                                                  & {\glsxtrshort{vitsp16224}} & \multicolumn{1}{c}{-}        & $73.73 \pm 0.60$             \\
      \cmidrule{1-4}
      \multirow{4}{*}{{\glsxtrshort{vitbr50384}}} & \glsxtrshort{rexnet100}    & \multicolumn{1}{c}{-}        & $66.91 \pm 0.79$             \\
                                                  & {\glsxtrshort{vitbp16224}} & $1.10 \pm 0.09$              & $75.70 \pm 0.68$             \\
                                                  & {\glsxtrshort{vitbr50384}} & $85.85 \pm 0.18$             & $85.04 \pm 0.38$             \\
                                                  & {\glsxtrshort{vitsp16224}} & \multicolumn{1}{c}{-}        & $75.52 \pm 0.36$             \\
      \cmidrule{1-4}
      \multirow{4}{*}{{\glsxtrshort{vitsp16224}}} & \glsxtrshort{rexnet100}    & \multicolumn{1}{c}{-}        & $56.60 \pm 0.39$             \\
                                                  & {\glsxtrshort{vitbp16224}} & \multicolumn{1}{c}{-}        & $70.14 \pm 0.46$             \\
                                                  & {\glsxtrshort{vitbr50384}} & \multicolumn{1}{c}{-}        & $62.85 \pm 1.22$             \\
                                                  & {\glsxtrshort{vitsp16224}} & $84.11 \pm 0.14$             & $83.24 \pm 0.13$             \\
      \bottomrule
   \end{tabular}
\end{table}

\begin{figure}[ht]
   \centering
   \begin{overpic}[trim=-1cm 1cm -0.5cm 0cm,width=.45\linewidth]{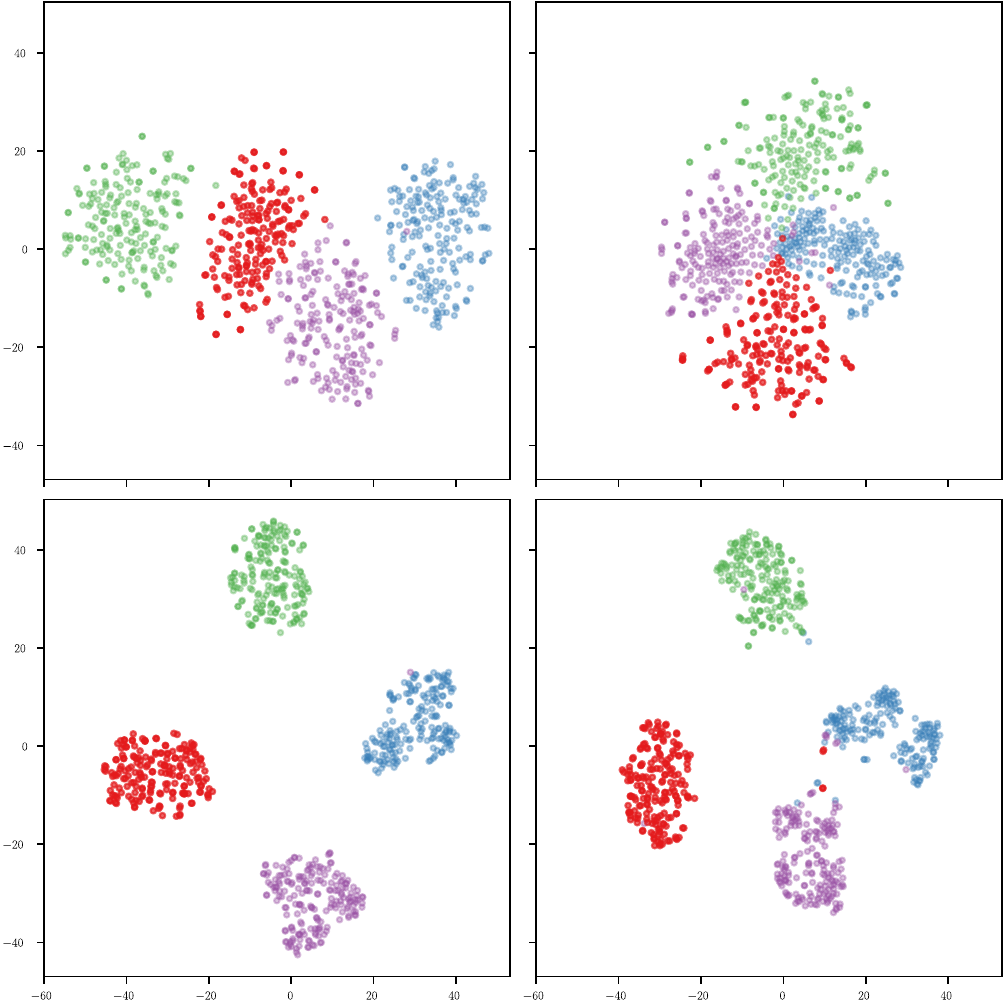}
      \put(18, 90){\gls{fasttext}}
      \put(63, 90){\gls{word2vec}}
      \put(-1, 54){\rotatebox{90}{\small Absolute}}
      \put(-1, 12){\rotatebox{90}{\small Relative}}
   \end{overpic}
   \hfill
   \begin{overpic}[trim=-1cm 1cm -0.5cm 0cm,width=.45\linewidth]{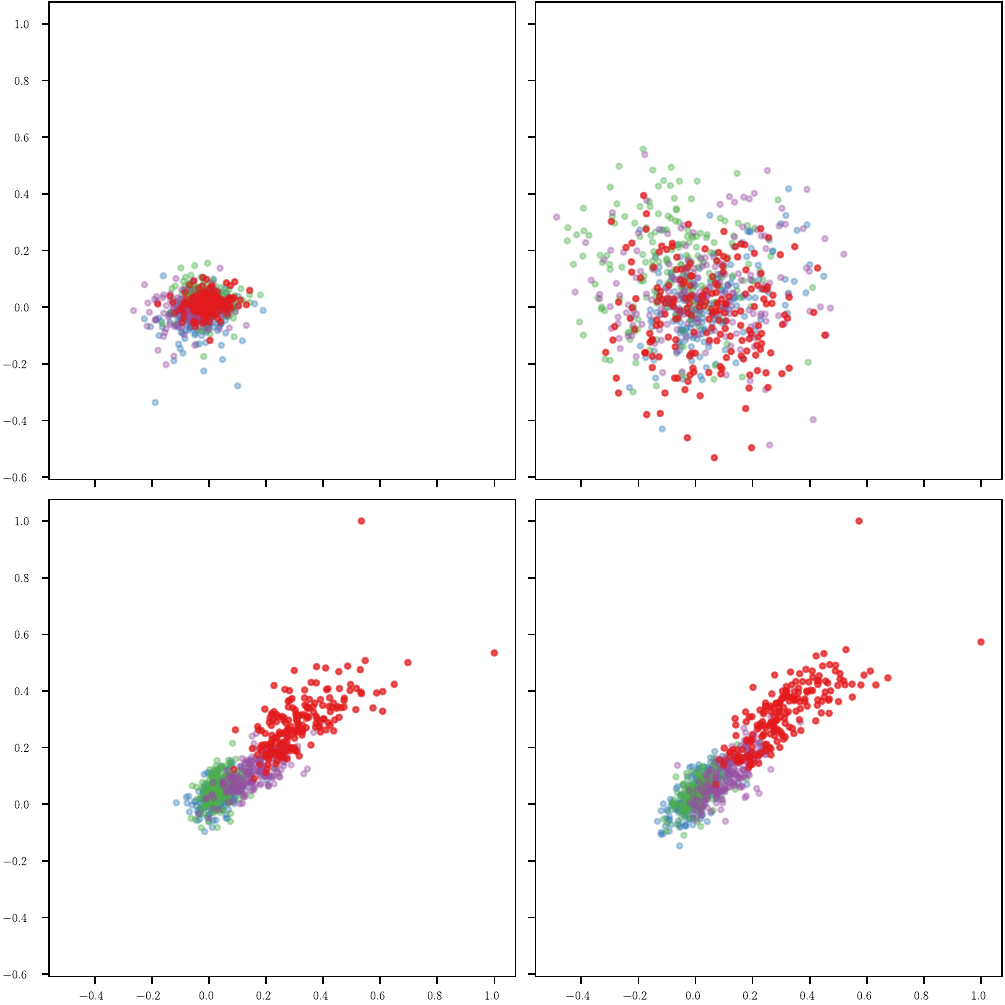}
      \put(18, 90){\gls{fasttext}}
      \put(63, 90){\gls{word2vec}}
   \end{overpic}
   \caption[Alternative visualization of \Cref{relative:fig:latent-rotation-comparison} with t-SNE]{{Same encodings as in \Cref{relative:fig:latent-rotation-comparison} (left) but with tSNE \textit{(left)} dimensionality reduction or visualizing only their first two dimensions \textit{(right)}.}}
   \label{relative:fig:word-embedding-proj}
\end{figure}

\clearpage

\chapter{Direct Translation}

\section{Additional results}

\begin{figure}[!ht]
    \centering
    \begin{minipage}{.60\columnwidth}
        \begin{overpic}[trim=-1cm -1cm 0 0,width=1\linewidth]{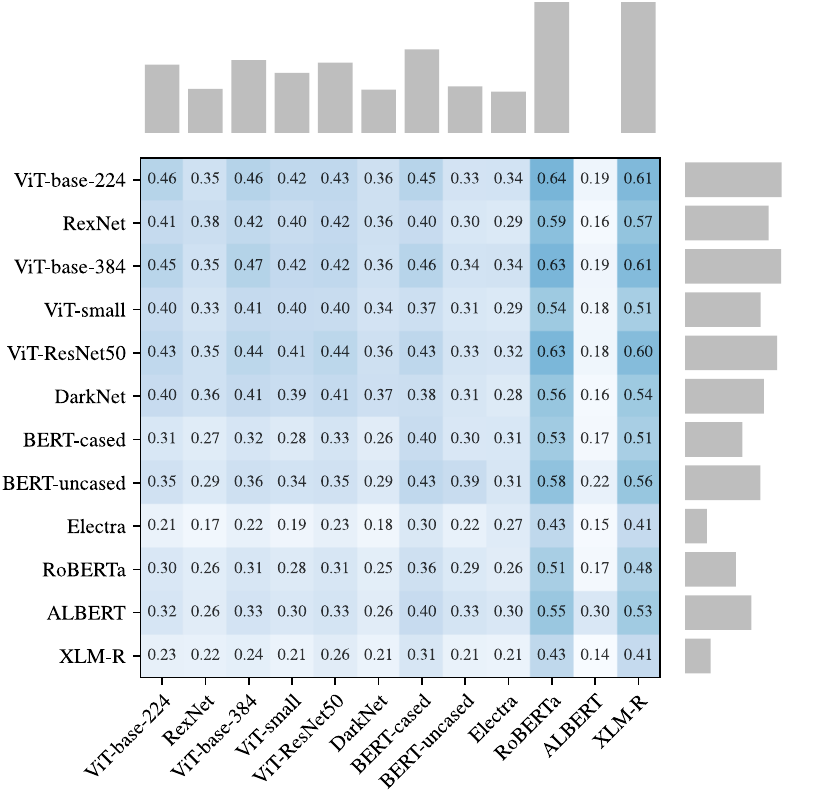}
            \put(0.5, 44){\rotatebox{90}{ Decoder}}
            \put(45, 2){ Encoder}
        \end{overpic}
    \end{minipage}
    \begin{minipage}{.375\columnwidth}
        \scriptsize
        \begin{tabular}{lllr}
            \toprule
            {}                                               & \textbf{Encoder}           & \textbf{Score} & \textbf{Scale} \\
            \midrule
            \multirow{5}{*}{\rotatebox{90}{\textbf{Vision}}} & \glsxtrshort{vitbp16224}   & 0.46           & 90.45          \\
                                                             & \glsxtrshort{rexnet100}    & 0.38           & 13.46          \\
                                                             & \glsxtrshort{vitbp16384}   & 0.47           & 89.66          \\
                                                             & \glsxtrshort{vitsp16224}   & 0.40           & 50.17          \\
                                                             & \glsxtrshort{vitbr50384}   & 0.44           & 32.10          \\
                                                             & \glsxtrshort{cspdarknet53} & 0.37           & 11.62          \\
            \midrule
            \multirow{6}{*}{\rotatebox{90}{\textbf{Text}}}   & \glsxtrshort{bertbc}       & 0.40           & 15.43          \\
                                                             & \glsxtrshort{bertbu}       & 0.39           & 14.54          \\
                                                             & \glsxtrshort{electrabd}    & 0.27           & 11.94          \\
                                                             & \glsxtrshort{robertab}     & 0.51           & 11.06          \\
                                                             & \glsxtrshort{albertbv2}    & 0.30           & 32.27          \\
                                                             & \glsxtrshort{xlmrobertab}  & 0.41           & 18.75          \\
            \bottomrule
        \end{tabular}
    \end{minipage}

    \caption[Performance comparison between different encoders and data modalities]{Performance comparison between different encoders and data modalities on  the \gls{n24news} multimodal dataset. On the right, the accuracy of models trained end-to-end on a single data modality (Score) and their average norm (Scale). On the left the stitching performance between pairs of encoders and decoder. This shows the importance of translating from good encoders, that can even improve unimodal decoder performances. Results obtained with $2000$ anchors and \texttt{SVD}, with a \glsxtrshort{mlp} as classification head.}
    \label{translation:sup:cross-modality}
\end{figure}

In \Cref{translation:sup:cross-modality}, we present the outcomes of the multimodal experiment presented in \Cref{translation:sec:cross-modality} with an \glsxtrshort{mlp} employed as the classification head, instead of \glsxtrshortpl{svm}. The findings highlight the \glsxtrshort{mlp}'s capability to leverage cross-modal information, leading to improved performance. However, the underlying mechanisms responsible for this enhancement remain unclear and warrant further investigation.

\begin{figure}[ht]
    \centering
    \begin{overpic}[trim={-0.5 0 0 0},clip,width=0.9\linewidth]{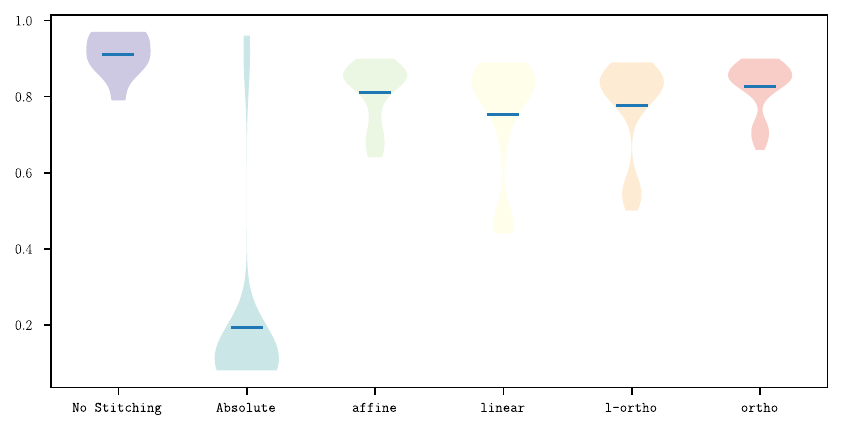}
        \put(-3, 23){\rotatebox{90}{\small Accuracy}}
        \put(50, -1){\small Method}

    \end{overpic}
    \caption[Cross-domain stitching on \gls{cifart} and grayscale \gls{cifart}]{Cross-domain stitching on \gls{cifart} and grayscale \gls{cifart}. 84 stitched pairs (pre-trained encoder - \glsxtrshort{svm} classifier) for 5 different seeds.}
    \label{translation:sup:fig:cross-domain}
\end{figure}

\begin{figure}[ht]
    \centering
    \begin{overpic}[trim=0cm -0cm 0 -0cm, clip,width=0.9\linewidth]{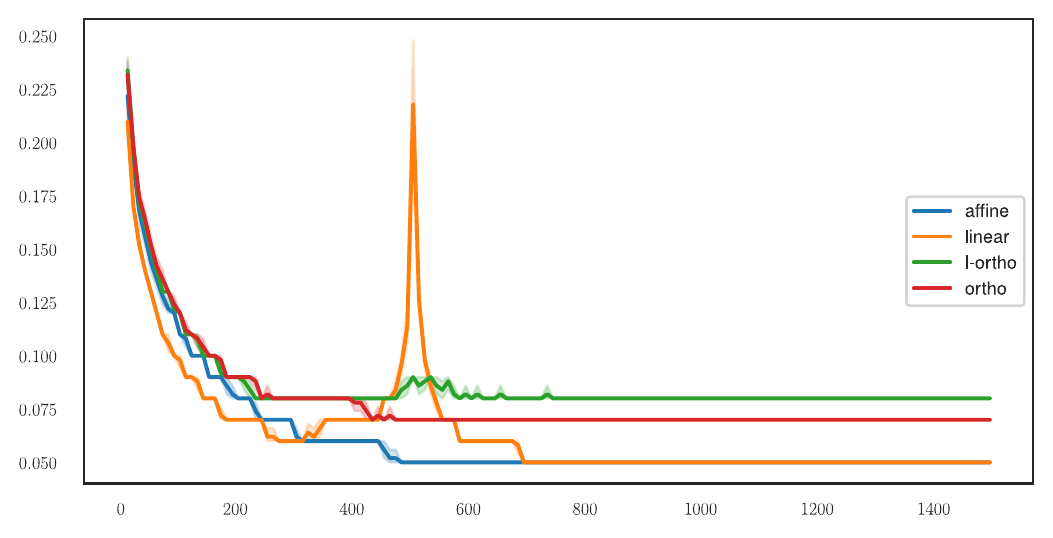}
        \put(-2, 15){\rotatebox{90}{Reconstruction MSE}}
        \put(40, -1.5){Number of anchors}
    \end{overpic}
    \caption[Performance comparison of \texttt{affine}, \texttt{linear}, \texttt{l-ortho} and \texttt{ortho}]{Performance comparison (reconstruction error) of \texttt{affine}, \texttt{linear}, \texttt{l-ortho} and \texttt{ortho} at varying anchor number on reconstruction task. Results on stitching 2 different \gls{cifarh}-trained AEs with 5 samplings for each anchor quantity. The naive absolute baseline is flat on 0.38 as mean.}
    \label{translation:sup:translation:fig:anchors-num}
\end{figure}

In \Cref{translation:sup:stitching:mlp:std,translation:sup:stitching:mlp:l2} quantitative results for stitching of \glsxtrshort{mlp} classifiers (again, differently from \Cref{translation:tab:stitching:svm,translation:tab:stitching:svm:l2} where \glsxtrshortpl{svm} are used) trained on top of pre-trained feature extractors, with and without additional L2 normalization, respectively.

\begin{figure}[!ht]
    \hfill \begin{overpic}[trim=-0.27cm 0cm -.3cm 0cm, clip,width=.96\linewidth,]{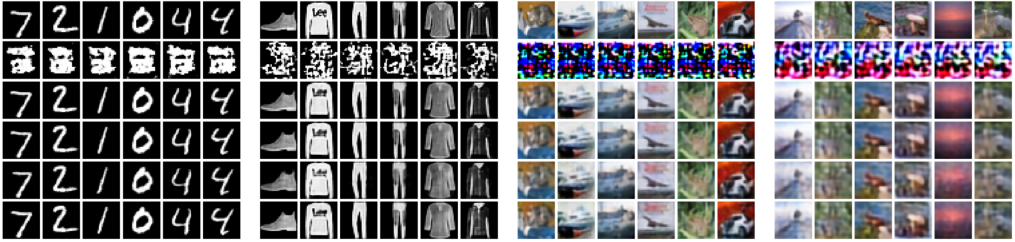}
        \put(.1, 20.7){\rotatebox{0}{\tiny \texttt{S}}}
        \put(-1.5, 16.7){\rotatebox{0}{\tiny \texttt{Abs.}}}
        \put(-3.4, 13){\rotatebox{0}{\tiny \texttt{affine}}}
        \put(-3.4, 9){\rotatebox{0}{\tiny \texttt{linear}}}
        \put(-4.25, 5.2){\rotatebox{0}{\tiny \texttt{l-ortho}}}
        \put(-2.75, 1.5){\rotatebox{0}{\tiny \texttt{ortho}}}
    \end{overpic}

    \caption[Translation reconstruction examples grouped by dataset.]{Reconstruction examples grouped by dataset. Each column is a different image, from top to bottom: original image, absolute stitching, \texttt{LSS} stitching, \texttt{OLSS} stitching, and \texttt{SVD} stitching. An L2 normalization is applied to the decoder input.}
    \label{translation:sup:fig:aes-decoders-with-l2-norm}
\end{figure}

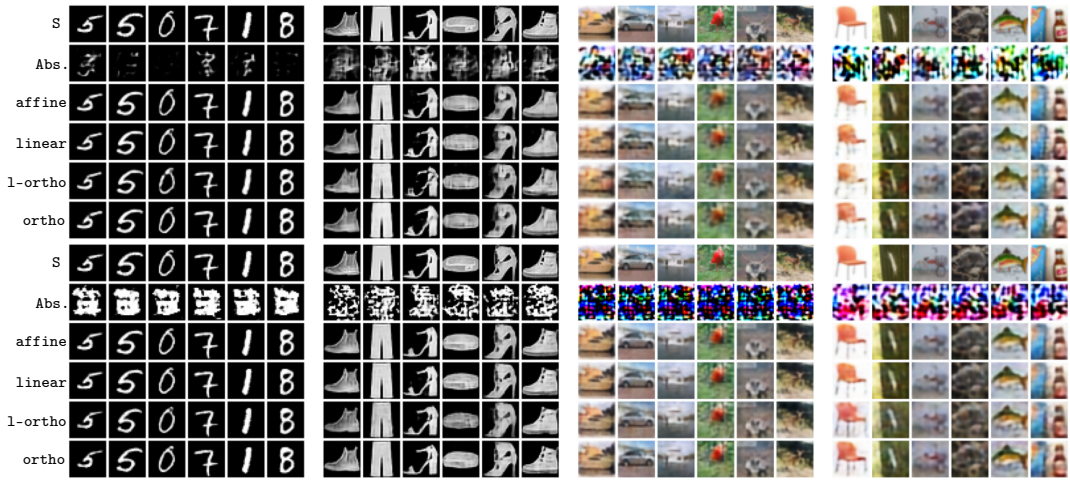
\begin{figure}[!ht]
    \hfill \begin{overpic}[trim=-0.27cm 0cm -.3cm 0cm, clip,width=.96\linewidth,]{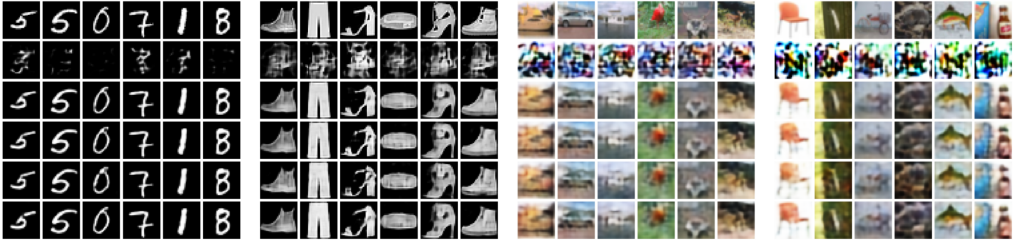}
        \put(.1, 20.7){\rotatebox{0}{\tiny \texttt{S}}}
        \put(-1.5, 16.7){\rotatebox{0}{\tiny \texttt{Abs.}}}
        \put(-3.4, 13){\rotatebox{0}{\tiny \texttt{affine}}}
        \put(-3.4, 9){\rotatebox{0}{\tiny \texttt{linear}}}
        \put(-4.25, 5.2){\rotatebox{0}{\tiny \texttt{l-ortho}}}
        \put(-2.75, 1.5){\rotatebox{0}{\tiny \texttt{ortho}}}
    \end{overpic}
    \\\vspace{0.1cm}
    \hfill \begin{overpic}[trim=-0.27cm 0cm -.3cm 0cm, clip,width=.96\linewidth,]{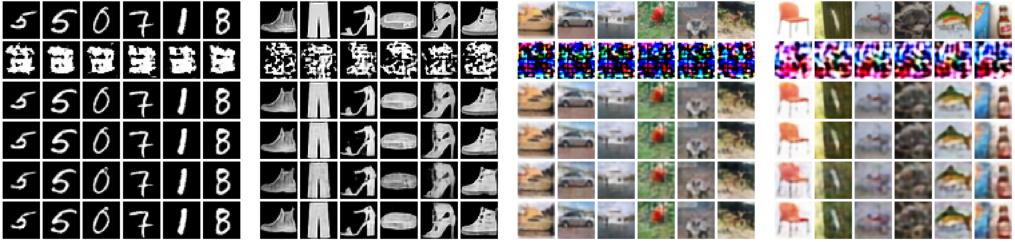}
        \put(.1, 20.7){\rotatebox{0}{\tiny \texttt{S}}}
        \put(-1.5, 16.7){\rotatebox{0}{\tiny \texttt{Abs.}}}
        \put(-3.4, 13){\rotatebox{0}{\tiny \texttt{affine}}}
        \put(-3.4, 9){\rotatebox{0}{\tiny \texttt{linear}}}
        \put(-4.25, 5.2){\rotatebox{0}{\tiny \texttt{l-ortho}}}
        \put(-2.75, 1.5){\rotatebox{0}{\tiny \texttt{ortho}}}
    \end{overpic}

    \caption[Additional reconstruction examples grouped by dataset]{Additional reconstruction examples grouped by dataset. Each column is a different image, from top to bottom: original image, absolute stitching, \texttt{LSS} stitching, \texttt{OLSS} stitching, and \texttt{SVD} stitching. In the first row, no additional normalization is applied on the decoder input; in the second row, an L2 normalization is applied instead.}
    \label{translation:sup:fig:more-aes-decoders-with-l2-norm}
\end{figure}

In \Cref{translation:sup:fig:aes-decoders-with-l2-norm,translation:sup:fig:more-aes-decoders-with-l2-norm}, there are additional reconstruction examples with the same autoencoding setting as in \Cref{translation:fig:aes-decoders-without-l2-norm}, and with additional L2 normalization, respectively.

In \Cref{translation:sup:table:aes-decoder-with-norm-1000-anchors} there are more quantitative results for stitching of autoencoders, with added L2 normalization (at training time) to the decoders of the reconstruction setting of \Cref{translation:table:aes-decoder-without-norm-1000-anchors}.

\begin{table}[!ht]
    \caption[Zero-shot stitching for generation.]{Zero-shot stitching for generation. With \texttt{SVD} for estimating ${\gls{transformationapprox}}$ and standard scaling as pre-processing. An L2 normalization is applied to the decoder input. We report the latent cosine similarity (\textit{lcos}) and MSE (\textit{lmse}) between the target encoding and the translated one, but also the reconstruction MSE (\textit{rmse}) between the input and the output.}
    \label{translation:sup:table:aes-decoder-with-norm-1000-anchors}
    \scriptsize
    \centering
    \begin{tabular}{lrrrrrrrrrrrr}
        \toprule
                         & \multicolumn{3}{c}{    \gls{mnist} } & \multicolumn{3}{c}{    \gls{fmnist} } & \multicolumn{3}{c}{   \gls{cifart} } & \multicolumn{3}{c}{    \gls{cifarh}}                                                                                                                 \\
        \cmidrule(lr){2-4} \cmidrule(lr){5-7} \cmidrule(lr){8-10} \cmidrule(lr){11-13}
                         & \emph{lcos}                          & \emph{lmse}                           & \emph{rmse}                          & \emph{lcos}                          & \emph{lmse} & \emph{rmse} & \emph{lcos} & \emph{lmse} & \emph{rmse} & \emph{lcos} & \emph{lmse} & \emph{rmse} \\
        \midrule
        \texttt{Abs.}    & 0.39                                 & 0.98                                  & 0.28                                 & 0.53                                 & 0.97        & 0.33        & 0.62        & 1.23        & 0.46        & 0.59        & 1.17        & 0.38        \\
        \texttt{affine}  & 0.99                                 & 0.15                                  & 0.01                                 & 0.99                                 & 0.16        & 0.03        & 0.99        & 0.16        & 0.04        & 0.99        & 0.12        & 0.05        \\
        \texttt{linear}  & 0.98                                 & 0.17                                  & 0.01                                 & 0.98                                 & 0.18        & 0.03        & 0.99        & 0.16        & 0.04        & 0.99        & 0.13        & 0.05        \\
        \texttt{l-ortho} & 0.89                                 & 0.41                                  & 0.02                                 & 0.91                                 & 0.41        & 0.04        & 0.96        & 0.39        & 0.05        & 0.93        & 0.30        & 0.08        \\
        \texttt{ortho}   & 0.97                                 & 0.21                                  & 0.02                                 & 0.97                                 & 0.23        & 0.03        & 0.99        & 0.21        & 0.05        & 0.96        & 0.22        & 0.07        \\
        \bottomrule
    \end{tabular}
\end{table}

\begin{table}[!ht]
    \tiny
    \centering
    \caption[Cross-architecture stitching with various ${\gls{transformationapprox}}$ and standard sacling]{Cross-architecture stitching with various methods for estimating ${\gls{transformationapprox}}$ and employing standard scaling. The stitched decoders are simple \glsxtrshortpl{mlp}. 5 runs for each encoder-decoder pair. (C) and (F) next to \gls{cifarh} indicate, respectively, coarse-grained and fine-grained.}
    \label{translation:sup:stitching:mlp:std}
    \begin{tabular}{lllllllll}
        \toprule
                                                                & \multicolumn{1}{l}{\texttt{Dataset}} & \multicolumn{1}{c}{\texttt{no-stitch}} & \multicolumn{1}{c}{\texttt{absolute}} & \multicolumn{1}{c}{\texttt{relative}} & \multicolumn{1}{c}{\texttt{affine}} & \multicolumn{1}{c}{\texttt{linear}} & \multicolumn{1}{c}{\texttt{l-ortho}} & \multicolumn{1}{c}{\texttt{ortho}} \\
        \midrule
        \multirow{5}{*}{\rotatebox{90}{\tiny \textbf{Vision}}}  & \gls{cifart}                         & $0.95 \pm 0.03$                        & $0.16 \pm 0.22$                       & $0.73 \pm 0.21$                       & $0.93 \pm 0.05$                     & $0.89 \pm 0.11$                     & $0.90 \pm 0.09$                      & $0.93 \pm 0.04$                    \\
                                                                & \gls{cifarh}-\texttt{C}              & $0.82 \pm 0.07$                        & $0.11 \pm 0.21$                       & $0.39 \pm 0.17$                       & $0.76 \pm 0.08$                     & $0.71 \pm 0.15$                     & $0.74 \pm 0.11$                      & $0.78 \pm 0.07$                    \\
                                                                & \gls{cifarh}-\texttt{F}              & $0.68 \pm 0.14$                        & $0.06 \pm 0.20$                       & $0.13 \pm 0.09$                       & $0.59 \pm 0.13$                     & $0.55 \pm 0.18$                     & $0.56 \pm 0.17$                      & $0.62 \pm 0.12$                    \\
                                                                & \glsxtrshort{fmnist}                 & $0.87 \pm 0.02$                        & $0.14 \pm 0.20$                       & $0.64 \pm 0.12$                       & $0.85 \pm 0.02$                     & $0.83 \pm 0.05$                     & $0.80 \pm 0.06$                      & $0.84 \pm 0.02$                    \\
                                                                & \gls{mnist}                          & $0.92 \pm 0.03$                        & $0.15 \pm 0.20$                       & $0.36 \pm 0.14$                       & $0.92 \pm 0.03$                     & $0.87 \pm 0.08$                     & $0.74 \pm 0.12$                      & $0.88 \pm 0.03$                    \\
        \midrule \multirow{4}{*}{\rotatebox{90}{\textbf{Text}}} & \gls{trec}                           & $0.41 \pm 0.07$                        & $0.15 \pm 0.04$                       & $0.27 \pm 0.09$                       & $0.40 \pm 0.08$                     & $0.37 \pm 0.11$                     & $0.23 \pm 0.08$                      & $0.41 \pm 0.09$                    \\
                                                                & \gls{agnews}                         & $0.76 \pm 0.08$                        & $0.24 \pm 0.02$                       & $0.36 \pm 0.10$                       & $0.68 \pm 0.08$                     & $0.65 \pm 0.08$                     & $0.64 \pm 0.10$                      & $0.68 \pm 0.10$                    \\
                                                                & \gls{dbpedia}                        & $0.64 \pm 0.19$                        & $0.07 \pm 0.02$                       & $0.15 \pm 0.08$                       & $0.57 \pm 0.19$                     & $0.53 \pm 0.19$                     & $0.44 \pm 0.21$                      & $0.56 \pm 0.17$                    \\
                                                                & \gls{imdb}                           & $0.62 \pm 0.04$                        & $0.50 \pm 0.01$                       & $0.50 \pm 0.01$                       & $0.59 \pm 0.04$                     & $0.58 \pm 0.04$                     & $0.57 \pm 0.04$                      & $0.60 \pm 0.04$                    \\
        \bottomrule
    \end{tabular}
\end{table}

\begin{table}[!ht]
    \tiny
    \centering
    \caption[Cross-architecture stitching with various ${\gls{transformationapprox}}$ and L2 normalization]{Cross-architecture stitching with various methods for estimating ${\gls{transformationapprox}}$ and applying L2 as normalization. The stitched decoders are simple \glsxtrshortpl{mlp}. 5 runs for each encoder-decoder pair. (C) and (F) next to \gls{cifarh} indicate, respectively, coarse-grained and fine-grained.}
    \label{translation:sup:stitching:mlp:l2}
    \begin{tabular}{lllllllll}
        \toprule
                                                                & \multicolumn{1}{l}{\texttt{Dataset}} & \multicolumn{1}{c}{\texttt{no-stitch}} & \multicolumn{1}{c}{\texttt{absolute}} & \multicolumn{1}{c}{\texttt{relative}} & \multicolumn{1}{c}{\texttt{affine}} & \multicolumn{1}{c}{\texttt{linear}} & \multicolumn{1}{c}{\texttt{l-ortho}} & \multicolumn{1}{c}{\texttt{ortho}} \\
        \midrule
        \multirow{5}{*}{\rotatebox{90}{\tiny \textbf{Vision}}}  & \gls{cifart}                         & $0.95 \pm 0.03$                        & $0.16 \pm 0.22$                       & $0.73 \pm 0.21$                       & $0.93 \pm 0.04$                     & $0.89 \pm 0.11$                     & $0.89 \pm 0.11$                      & $0.93 \pm 0.04$                    \\
                                                                & \gls{cifarh}-\texttt{C}              & $0.82 \pm 0.07$                        & $0.11 \pm 0.21$                       & $0.39 \pm 0.17$                       & $0.77 \pm 0.07$                     & $0.75 \pm 0.13$                     & $0.71 \pm 0.15$                      & $0.78 \pm 0.06$                    \\
                                                                & \gls{cifarh}-\texttt{F}              & $0.68 \pm 0.14$                        & $0.06 \pm 0.20$                       & $0.13 \pm 0.09$                       & $0.60 \pm 0.12$                     & $0.57 \pm 0.18$                     & $0.54 \pm 0.18$                      & $0.61 \pm 0.12$                    \\
                                                                & \glsxtrshort{fmnist}                 & $0.87 \pm 0.02$                        & $0.14 \pm 0.20$                       & $0.64 \pm 0.12$                       & $0.86 \pm 0.02$                     & $0.79 \pm 0.09$                     & $0.83 \pm 0.05$                      & $0.84 \pm 0.02$                    \\
                                                                & \gls{mnist}                          & $0.92 \pm 0.03$                        & $0.15 \pm 0.20$                       & $0.36 \pm 0.14$                       & $0.91 \pm 0.03$                     & $0.80 \pm 0.17$                     & $0.86 \pm 0.08$                      & $0.86 \pm 0.04$                    \\
        \midrule \multirow{4}{*}{\rotatebox{90}{\textbf{Text}}} & \gls{trec}                           & $0.41 \pm 0.07$                        & $0.15 \pm 0.04$                       & $0.27 \pm 0.09$                       & $0.51 \pm 0.06$                     & $0.27 \pm 0.10$                     & $0.47 \pm 0.13$                      & $0.49 \pm 0.06$                    \\
                                                                & \gls{agnews}                         & $0.76 \pm 0.08$                        & $0.24 \pm 0.02$                       & $0.36 \pm 0.10$                       & $0.68 \pm 0.08$                     & $0.64 \pm 0.10$                     & $0.65 \pm 0.08$                      & $0.66 \pm 0.10$                    \\
                                                                & \gls{dbpedia}                        & $0.64 \pm 0.19$                        & $0.07 \pm 0.02$                       & $0.15 \pm 0.08$                       & $0.55 \pm 0.19$                     & $0.53 \pm 0.21$                     & $0.51 \pm 0.18$                      & $0.49 \pm 0.15$                    \\
                                                                & \gls{imdb}                           & $0.62 \pm 0.04$                        & $0.50 \pm 0.01$                       & $0.50 \pm 0.01$                       & $0.60 \pm 0.04$                     & $0.58 \pm 0.04$                     & $0.59 \pm 0.04$                      & $0.59 \pm 0.04$                    \\
        \bottomrule
    \end{tabular}
\end{table}

\newpage

\subsection{Scale invariance}
\label{translation:sup:scale-invariance}

In this Section, we delve into the concept of scale invariance in \glspl{nn} and its implications for model compositionality. We start by focusing on the effect of rescaling operations on the latent input encodings and demonstrate that, by construction, certain classifiers exhibit scale-invariance properties without the need for additional priors. Then, by examining the behavior of networks when subjected to a specific type of input manipulation, \textit{rescaling injection}, we aim to demonstrate the robustness and versatility of \glspl{nn} in handling different scales of input data. As illustrated in \Cref{chap:translation}, this is a key advantage in improving the adaptability of our method.

The softmax function, commonly used in neural classifiers, is known to be a temperature-controlled variant of the maximum function:
\begin{equation}
    \operatorname{softmax}(x)_i=\frac{e^{\frac{y_i}{T}}}{\sum_j^N e^{\frac{y_j}{T}}}\,.
\end{equation}
This means that the softmax temperature can be used to control the level of confidence of the classifier's predictions. In this study, we show that a similar effect can also be achieved by rescaling the latent encodings given as input to a trained (and frozen) classifier.

In order to demonstrate this, we first note that the rescaling factor, $\alpha$, can be factored out of the matrix multiplication in the Linear layers of the classifier. This can be represented mathematically as:
$\mathbf{y} = \alpha \mathbf{W}\mathbf{x} + b$, where $\mathbf{x}$ is the input latent encoding, $\mathbf{W}$ is the weight matrix, $b$ is the bias vector, $\alpha$ is the rescaling factor, and $\mathbf{y}$ is the output of the linear layer. This implies that the rescaling operation can be ``pushed through'' the classifier without affecting its final prediction as it becomes equivalent to some temperature value applied at the softmax level.

Furthermore, we investigate the effect of rescaling when non-linear activation functions are involved and posit that as long as the function has a monotonic interval, if we rescale all the dimensions by an amount similar to the mean scale of the encodings on which the classifier was trained, we end up in the monotonic interval, without losing the scale-invariance property.

\begin{figure}[!ht]
    \centering
    \begin{overpic}[trim=1.25cm 2cm 1 3cm,clip,width=0.55\linewidth]{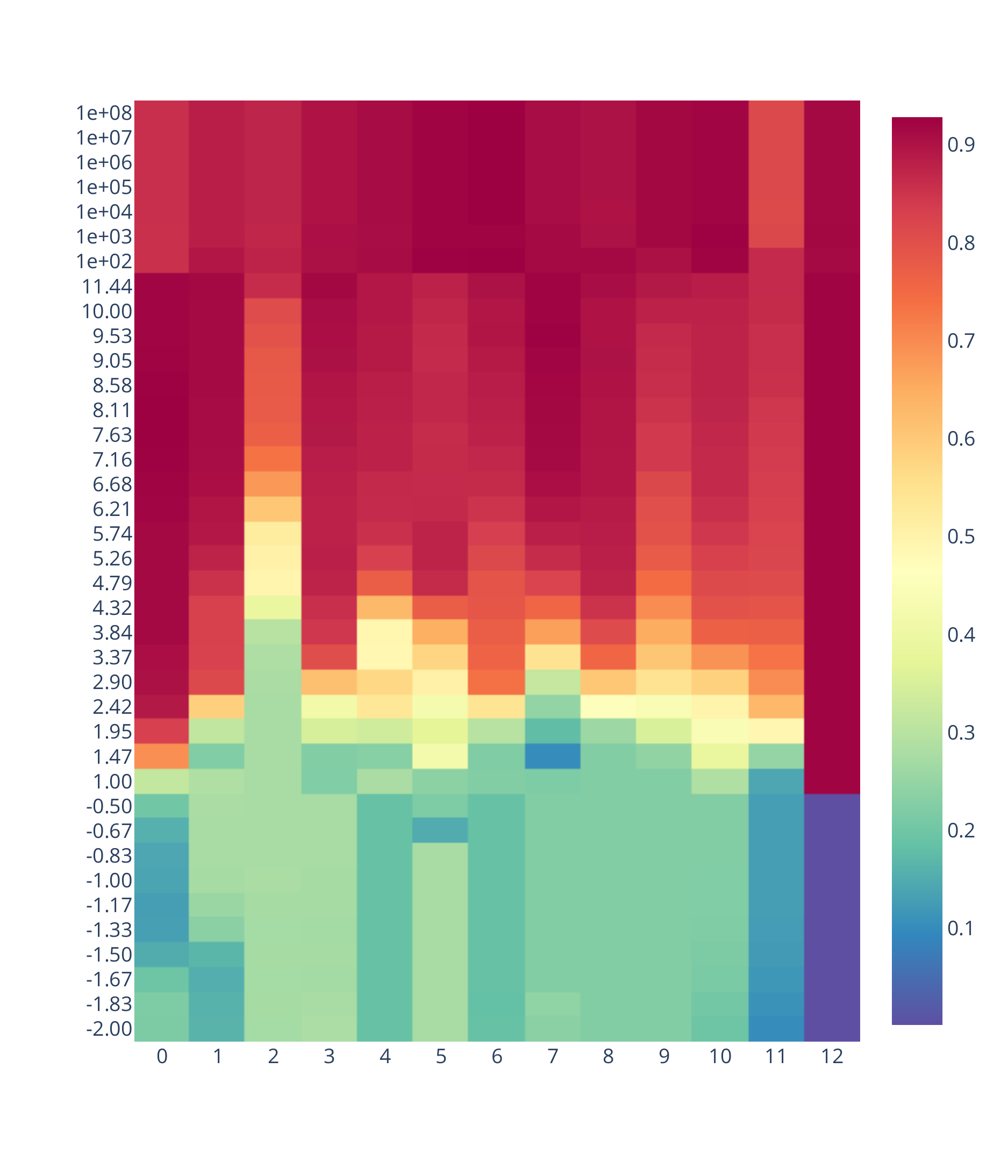}
        \put(33, 0){Rescaled layer}
        \put(0, 35){\rotatebox{90}{Rescaling factor $\alpha$}}
        \put(93, 59){\rotatebox{-90}{Accuracy}}
    \end{overpic}
    \caption[Scale invariance of {\glsxtrshort{robertab}}]{Scale invariance of {\glsxtrshort{robertab}} according to the performance of a downstream classifier trained on the encodings of the last attention layer. At each layer (with 0 being the embedding layer and 12 the output one), one for each run, we rescale the encodings by the specified $\alpha$ and measure its effect on the final accuracy. The performance without any rescaling is $0.92$. }
    \label{translation:fig:rescaled-layer-acc}
\end{figure}

In summary, our study provides empirical evidence that neural classifiers that utilize the softmax activation function can, in practice, maintain their scale-invariance properties when the input latent encodings are rescaled. This property is essential to our method, as it allows us to ignore the exact scale when decoding toward an L2-normalized absolute space.
\paragraph{Pre-trained models and scale-invariance.}
We observed that large pre-trained models, such as transformers and resnets, are robust to internal rescaling of the encodings. Although we do not have a strong theoretical explanation for this phenomenon, we hypothesize that normalization layers and the linear separability of the information encoded in the angles instead of the norms may play a significant role.
In \Cref{translation:fig:rescaled-layer-acc}, we demonstrate the invariance a large transformer exhibits when the rescaling injection is applied at different layers: surprisingly, when the rescaling surpasses a certain threshold, the performance difference becomes negligible.
These results further emphasize the robustness of these pre-trained models to the rescaling injection and suggest that the scale of the embedding is not a critical factor in their performance.

\paragraph{Rescale Injection.}

\begin{figure}
    \centering
    \begin{overpic}[trim=0 -0.5cm 0 0, clip,width=.9\linewidth]{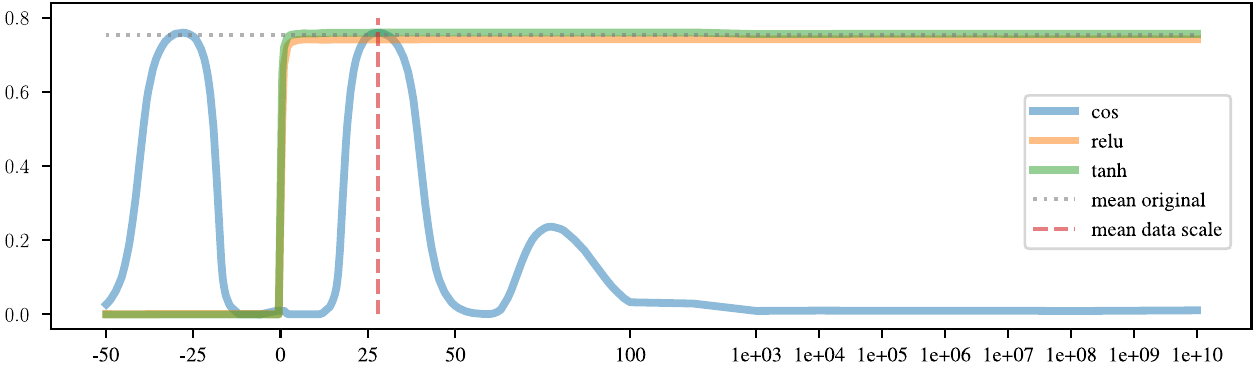}
        \put(-3.5, 13){\rotatebox{90}{\small Accuracy}}
        \put(43, -1){\small Rescale Factor}
    \end{overpic}

    \caption[Performance comparison of three \glsxtrshortpl{mlp}]{Performance comparison of three \glsxtrshortpl{mlp} with different activation functions, namely cosine (blue), ReLU (orange), and tanh (green) at different rescaling factors $\alpha$. The ReLU and tanh \glsxtrshortpl{mlp} exhibit scale invariance, while the cosine activation function is only invariant on the mean data scale and its periodic cycles.
    }
    \label{translation:fig:scale-inv-mono}
\end{figure}

We define the \textit{rescaling injection} as the operation of artificially altering the scale of the features produced at a specific layer of the network. This is achieved by normalizing the embeddings to unit norm and then rescaling them by a factor of $\alpha$. By varying the value of $\alpha$, we can observe how the network's performance is affected at different scales. Through this empirical analysis, we aim to provide insight into the scale invariance properties of \glspl{nn} and their potential for use in model compositionality.

In \Cref{translation:fig:scale-inv-mono}, we present experimental results investigating the scale invariance properties of \glspl{nn}. We trained simple multi-layer perceptrons (\glsxtrshortpl{mlp}) composed of two hidden layers, with no normalization layers, using encodings produced by the Clip Vision transformer (\texttt{clip-vit-base-patch32}) on the \gls{cifarh} (fine) dataset. The \glsxtrshortpl{mlp} were evaluated using different activation functions: cosine (blue), tanh (orange), and ReLU (green). The rescaling injection technique was applied directly to the input embeddings, rescaling them by $\alpha$.

We can observe that the scale of the embeddings does not significantly impact the \glsxtrshortpl{mlp}' performance when using monotone activation functions that do not flip signs. This is a non-trivial result, as the nonlinearity of the activation function, the presence of bias terms $b$, and the absence of normalization layers make it difficult to predict the effect of an input rescaling on the performance of the network.
It is particularly interesting to see that the cosine activation function shows an oscillatory performance, comparable to the original embeddings when rescaled by the mean embeddings scale (vertical red line) or its opposite since it is symmetric.

Our findings indicate that, surprisingly, even the internal layers of large deep learning models exhibit a \textit{positive scale invariance}, as illustrated in \Cref{translation:fig:rescaled-layer-acc}. The underlying mechanism for this behavior is not straightforward, but we hypothesize that it may result from the interplay between various factors, such as the choice of activation function, the use of normalization layers, the optimization objective and regularization techniques employed during the training phase. Further research is needed to understand and explain this phenomenon fully.

\subsection{Implementation Details}
All the experiments were conducted using a machine equipped with an Intel Core i7-9700k CPU, 64 GB of RAM, and an NVIDIA 2080TI GPU.

\paragraph{Decoder structure.}
The full implementation details can be found in the attached code,the various experiments can be run by their corresponding notebook.

\begin{itemize}
    \item \textit{Autoencoding}. Since the autoencoders were used only on image data, the architecture was a simple sequence of convolutions (in the encoder part) and deconvolutions (in the decoder part). Each interleaved with nonlinear activations.
    \item \textit{Classification}. \Cref{chap:translation} refers to "\glsxtrshort{svm}" as the standard \glsxtrshort{svm} implementation in scikit-learn \citep{scikit}, with default parameters. The experiments with "\glsxtrshort{mlp}" as a classifier refer to a simple stack of 3 linear layers, interleaved by nonlinear activations.
\end{itemize}

\paragraph{Software and Technologies.}
The research of this study was facilitated by the use of various technologies and tools, which include:
\begin{itemize}
    \item \textit{NN-Template} \citep{nn-template}, was used to kick-start the project while also ensuring best practices were adhered to.
    \item \textit{DVC} \citep{dvc}, was implemented for data versioning.
    \item \textit{PyTorch Lightning} \citep{lightning}, contributed to maintaining the integrity of the results and promoting clean, modular code.
    \item \textit{Weights and Biases} \citep{wandb}, were employed for logging experiments, running comparisons over extensive sweeps, and sharing models.
    \item \textit{Transformers by HuggingFace} \citep{wolf-etal-2020-transformers}, provided pre-configured transformers for processing both image and text data.
    \item \textit{Datasets by HuggingFace} \citep{lhoest-etal-2021-datasets}, facilitated access to a majority of NLP datasets and ImageNet for computer vision purposes.
\end{itemize}

\paragraph{Pre-trained encoders.} All the pre-trained encoders used come from HuggingFace and are listed in \Cref{translation:sup:hf:models}. They are various both in terms of architecture and encoding size.
\begin{table}[ht]
    \centering
    \caption[HuggingFace models used as encoders (feature extractors)]{HuggingFace models used as encoders (feature extractors) in the various experiments, with their encoding dimensionality.}
    \label{translation:sup:hf:models}
    \begin{tabular}{clc}
        \toprule
        Modality                                           & HuggingFace model name    & Encoding Dim \\
        \midrule
        \multirow{7}{*}{\rotatebox{90}{\textbf{Language}}} & \glsxtrlong{bertbc}       & 768          \\
                                                           & \glsxtrlong{bertbu}       & 768          \\
                                                           & \glsxtrlong{electrabd}    & 768          \\
                                                           & \glsxtrlong{robertab}     & 768          \\
                                                           & \glsxtrlong{albertbv2}    & 768          \\
                                                           & \glsxtrlong{xlmrobertab}  & 768          \\
                                                           & \glsxtrlong{clip}         & 768          \\

        \midrule
        \multirow{7}{*}{\rotatebox{90}{\textbf{Vision}}}   & \glsxtrlong{rexnet100}    & 1280         \\
                                                           & \glsxtrlong{cspdarknet53} & 768          \\
                                                           & \glsxtrlong{vitsp16224}   & 384          \\
                                                           & \glsxtrlong{vitbp16224}   & 768          \\
                                                           & \glsxtrlong{vitbp16384}   & 768          \\
                                                           & \glsxtrlong{vitbr50384}   & 768          \\
                                                           & \glsxtrlong{clip}         & 768          \\

        \bottomrule
    \end{tabular}
\end{table}

\begin{table}
    \centering
    \caption[Cross-architecture stitching for reconstruction tasks.]{Cross-architecture stitching for reconstruction tasks. 5 different seeds, 2 different bottleneck sizes (250, 500) for the same architecture. Average over all combinations. 500 anchors used and standard scaling as normalization. The naive absolute baseline is impossible to compute due to the dimensionality mismatch.}
    \label{translation:sup:cross-architecture:generation}
    \scriptsize
    \centering
    \begin{tabular}{lrrrrrrrrrrrr}
        \toprule
                         & \multicolumn{3}{c}{    \gls{mnist} } & \multicolumn{3}{c}{    \gls{fmnist} } & \multicolumn{3}{c}{   \gls{cifart} } & \multicolumn{3}{c}{    \gls{cifarh}}                                                                                                                 \\
        \cmidrule(lr){2-4} \cmidrule(lr){5-7} \cmidrule(lr){8-10} \cmidrule(lr){11-13}
                         & \emph{lcos}                          & \emph{lmse}                           & \emph{rmse}                          & \emph{lcos}                          & \emph{lmse} & \emph{rmse} & \emph{lcos} & \emph{lmse} & \emph{rmse} & \emph{lcos} & \emph{lmse} & \emph{rmse} \\
        \midrule
        \texttt{affine}  & 0.95                                 & 0.09                                  & 0.02                                 & 0.95                                 & 0.09        & 0.04        & 0.98        & 0.06        & 0.05        & 0.98        & 0.07        & 0.06        \\
        \texttt{linear}  & 0.64                                 & 1.00                                  & 0.11                                 & 0.66                                 & 1.10        & 0.16        & 0.77        & 0.60        & 0.16        & 0.78        & 0.52        & 0.16        \\
        \texttt{l-ortho} & 0.87                                 & 0.16                                  & 0.03                                 & 0.89                                 & 0.14        & 0.06        & 0.95        & 0.12        & 0.08        & 0.95        & 0.13        & 0.08        \\
        \texttt{ortho}   & 0.91                                 & 0.14                                  & 0.03                                 & 0.92                                 & 0.13        & 0.06        & 0.96        & 0.12        & 0.09        & 0.96        & 0.12        & 0.09        \\
        \bottomrule
    \end{tabular}
\end{table}

\clearpage

{
\begingroup
\emergencystretch=1em

\printbibliography[heading=bibintoc]
\endgroup
}

\end{document}